\documentclass[oneside]{book}

\usepackage[utf8]{inputenc}
\usepackage{graphicx}
\usepackage{subfiles}
\usepackage{amsmath}
\usepackage{amssymb}
\usepackage[linesnumbered,ruled,vlined]{algorithm2e}
\usepackage{multirow}
\usepackage{booktabs}
\usepackage[table,xcdraw]{xcolor}
\usepackage{hyperref} 
\usepackage[final]{pdfpages}
\usepackage{caption}
\usepackage{subcaption}
\usepackage{float}

\DeclareMathOperator*{\argmin}{argmin}

\SetKwInput{KwInput}{Input}

\title{A Markov Reward Process-Based Approach to Spatial Interpolation}
\author{Laurens Arp, TODO: add supervisors in a neat way }
\date{September 2020}

\begin{document}

\includepdf[pages=2]{temp4.pdf}

\chapter*{Abstract}
The interpolation of spatial data can be of tremendous value in various applications, such as forecasting weather from only a few measurements of meteorological or remote sensing data. Existing  methods for spatial interpolation, such as variants of kriging and spatial autoregressive models, tend to suffer from at least one of the following limitations: (a) the assumption of stationarity, (b) the assumption of isotropy, and (c) the trade-off between modelling local or global spatial interaction. Addressing these issues in this work, we propose the use of Markov reward processes (MRPs) as a spatial interpolation method, and we introduce three variants thereof: (i) a basic static discount MRP (SD-MRP), (ii) an accurate but mostly theoretical optimised MRP (O-MRP), and (iii) a transferable weight prediction MRP (WP-MRP). All variants of MRP interpolation operate locally, while also implicitly accounting for global spatial relationships in the entire system through recursion. Additionally, O-MRP and WP-MRP no longer assume stationarity and are robust to anisotropy. We evaluated our proposed methods by comparing the mean absolute errors of their interpolated grid cells to those of 7 common baselines, selected from models based on spatial autocorrelation, (spatial) regression, and deep learning.

We performed detailed evaluations on two publicly available datasets (local GDP values, and COVID-19 patient trajectory data). The results from these experiments clearly show the competitive advantage of MRP interpolation, which achieved significantly lower errors than the existing methods in $23$ out of $40$ experimental conditions, or $35$ out of $40$ when including O-MRP.

\tableofcontents


\chapter{Introduction}
\label{chap:introduction}

Research or industry, data science or application, to many of us missing data remains a fact of life. The problem of ``filling in'' 
missing data between known observations is known as interpolation. This problem can become especially difficult to solve in spatial settings, as observations are made in two- or three-dimensional space, and one cannot necessarily assume \textit{stationarity} -- that is, the distribution of values may change depending on the location in question. Moreover, some spatial effects may have different properties depending on the direction of the interaction; this is referred to as \textit{anisotropy} (as opposed to isotropy).

Nonetheless, spatial interpolation is typically done by exploiting the property of \textit{spatial autocorrelation}: the values of a variable tend to be more similar the closer their spatial proximity is to one another. While this concept has been useful in designing 
spatial models, pure spatial autocorrelation would only consider either local or distance-based pair-wise (global) interactions, and not the geographical properties of any intermediate locations. This could be problematic; for example, two locations separated by a mountain range or a body of water may be less similar than two locations that are part of the same city, even if the distances between both pairs of locations were the same. 
The problem is further exacerbated by the two-dimensionality of the problem, as there are many different paths from one point on a 2D grid to another.

To date, a number of spatial interpolation methods have  
been proposed. Some of these, such as kriging 
\cite{matheron1963principles} and inverse distance weighting  \cite{shepard1968two}, rely solely on the spatial autocorrelation of the target variable to predict 
unknown values in a grid. Others, such as spatial autoregressive- \cite{anselin1988spatial} and moving average \cite{haining1978moving} models, optionally extend this with the use of explanatory variables. All these methods tend to suffer from at least one of the problems mentioned above, with individual methods offering different trade-offs in assumptions and limitations. 

\begin{figure}[t]
  \centering
  \includegraphics[width=\linewidth]{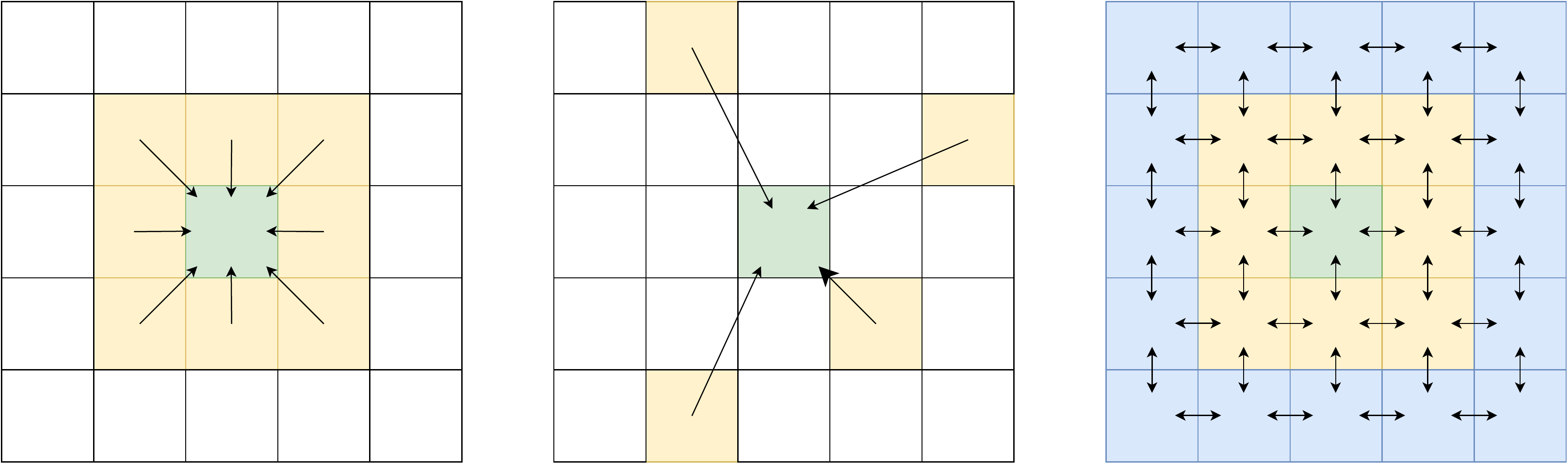}
  \caption{Illustration of local (left), distance-based (middle) and system-oriented (right) perspectives. In local and distance-based perspectives, the predicted value of the green cell is determined by the yellow cells (equal weights if local, unequal weights if distance-based). In the system-oriented perspective, which is the perspective used by our proposed method, the prediction for the green cell is made using the yellow neighbours, which were in turn affected by their own neighbours (blue, yellow and green cells).}
  \label{fig:local_global}
\end{figure}
In this thesis, 
we propose a Markov reward process (MRP)-based approach to spatial interpolation aimed at overcoming these limitations. This approach, inspired by the application of MRPs in reinforcement learning \cite{littman1994markov}, will treat the target variable prediction at a certain location as the ``expected reward'' of a state. While this interpolation method is local in principle, the use of a recursive definition using predicted neighbouring values allows MRPs to be implicitly global. This enables us to consider the entire region as a global system of locations, each with particular local properties, indirectly mutually interacting with all other locations via multiple paths, as illustrated in Figure \ref{fig:local_global}.\\ 
The main challenges in adapting MRPs to spatial interpolation are twofold. First, the necessary equivalence needs to be formalised between reinforcement learning concepts and their respective interpolation counterparts. 
Second, as MRPs are based on the concept of \textit{discounts} that can be exploited to represent spatial autocorrelation, 
determining the best discount setting poses a challenge. Simple approaches might use a single discount parameter, 
automatically tuned using a training set and applied to a test set, though this would assume both stationarity and isotropy. A more ambitious approach might be to construct a similarly data-driven manner of assigning weights, but specific to every individual pair of neighbours.

Our main contributions in this work are as follows: 
\begin{itemize}
    \item We propose a novel, 
    MRP-based spatial interpolation method, combining the strengths of local spatial autorcorrelation and implicit global spatial interactions.
    \item We introduce three MRP interpolation models, two of which are robust to anisotropy and non-stationarity.
    \item We evaluate our methods in terms of mean absolute error on spatial data from four cities from two different countries. As target variables, we used a global dataset on GDP by World Bank \cite{gdp_world_bank}, where we also tested for international transferability, and a South Korean dataset of COVID-19 patient trajectories \cite{covid_dacon}.  
    \item We compare our methods to various existing models for spatial interpolation, including kriging, spatial autoregressive models and convolutional neural networks. In this comparison, a practically applicable MRP outperformed all baselines in $23$ out of $40$ conditions. All methods were configured using automated algorithm configuration for (close to) optimal performance and to ensure a fair comparison.
\end{itemize}


In the next chapter (Chapter \ref{chap:problem_statement}), we will formalise the problem of spatial interpolation, and introduce the notation we will use throughout the thesis. Chapter \ref{chap:related_work} will cover existing work and methods for spatial interpolation, and Chapter \ref{chap:methods} will explain our own proposed methods in detail. As for the evaluation of those methods, Chapter \ref{chap:experiments} will cover the details of our experimental setup, and Chapter \ref{chap:results} will analyse the results of those experiments. We conclude the thesis with the conclusion in Chapter \ref{chap:conclusion}, and also add an ethical statement to give some insight into the ethical implications of our proposed methods, as well as the responsible use of our experimental datasets.

\chapter{Problem statement}
\label{chap:problem_statement}

We formalise the problem of spatial interpolation as follows. First, we define a two-dimensional matrix $\mathbf{Y}$ of size $H \times W$ 
of target variable values $y_{h,w}$, 
where $h$ represents the row index and $w$ represents the column index of a cell in $\mathbf{Y}$. Moreover, for convenient indexing we define $\mathbf{c} = [c_1,c_2,...,c_{H \times W}]$, 
where every $c \in \mathbf{c}$ represents an index pair $(h,w)$. 
For a target variable $y$ of interest, every cell in the matrix indexed by $c$ 
is associated with a true value $y^{*}_{c}$, which may be either known in $\mathbf{Y}$ or not. If it is known, we set the target value $y_{c} = y^{*}_{c}$, and if it is not known, we set $y_{c} = \varnothing$ (null value, not to be confused with an empty set). Second, we define the spatial feature vectors $\textbf{x}_{c} = [x_{c}^{1},x_{c}^{2},...,x_{c}^{D}]$ 
(where $D$ is the number of spatial features) for all $c \in \mathbf{c}$. These feature vectors contain explanatory variables derived from the geographical and spatial properties of the location at $c$, such as the number of houses, hospitals and bus stops. \\
The problem of spatial interpolation can now be formulated as creating a model $\mathcal{M}(\mathbf{Y},\mathbf{x})$ using known true values from $\mathbf{Y}$ and/or explanatory variable vectors $\mathbf{x}_{c}$ to predict an estimated target value $\hat{y}_{c} = \mathcal{M}(\mathbf{Y},\mathbf{x}_{c})$ for all $c$ where $y_{c} = \varnothing$. Our objective is to minimise the mean difference between $\hat{y}_{c}$ and the unknown true values $y^{*}_{c}$. 
Concretely, if we define $\mathbf{Y}_{\varnothing}$ as $\{c : y_{c} = \varnothing\}$, we wish to find an optimal model $\mathcal{M}^{*}$:
\begin{equation}
    \mathcal{M}^{*} \in \argmin_{\mathcal{M}} \frac{1}{|\mathbf{Y}_{\varnothing}|} \sum_{c \in \mathbf{Y}_{\varnothing}} (\mathcal{M}(\mathbf{Y},\mathbf{x}_{c}) - y^{*}_{c})
    \label{eq:problem_statement3}
\end{equation}

\chapter{Related work}
\label{chap:related_work}
To date, various spatial interpolation methods have been proposed. For convenience, we will categorise methods into (i) \textit{purely spatial methods}, and (ii) \textit{regression with spatial and explanatory variables}. Purely spatial methods consist of those models relying solely on known true values and their spatial locations to fill in the grid: $\hat{y}_{c} = \mathcal{M}(\mathbf{Y})$. Conversely, regression with spatial and explanatory variables predict $\hat{y}_{c}$ using spatial autocorrelation, while also supporting the use of explanatory variables: $\hat{y}_{c} = \mathcal{M}(\mathbf{Y},\mathbf{x}_{c})$.

\section{Purely spatial methods}

Given the widespread use of kriging-based interpolation, we will sub-categorise the purely spatial methods further into kriging and non-kriging methods.
\subsection{Kriging}

Kriging \cite{matheron1963principles,krige1951statistical}, 
more generally known as Gaussian processes outside of geostatistics, is a set of methods centred on the concept of the covariance of values and distance. That is, the relationship between the value of the target variable in two locations and the distance between these locations (spatial autocorrelation) is being modeled. This model, called the \textit{variogram} (also known as the \textit{kernel} in Gaussian processes) can take various forms, such as linear, exponential or Gaussian functions. Contemporary contributions to the kriging field include a scalable gradient-based surrogate function method \cite{bouhlel2019gradient} and a neural network-based method to overcome kriging's limitation of disregarding the characteristics of intermediate locations in paths between pairs of locations \cite{sato2019performance}.

Let us define a cell of interest as $c_i$, and a neighbouring cell of $c_i$ as $c_j$. The basic model for kriging interpolation can now be formulated as follows: 
\begin{equation}
    y_{c_i} = \sum_{c_j \in N(c_i)} (\gamma_{c_j,c_i} \cdot y_{c_j}) + \epsilon_{c_i}
    \label{eq:kriging}
\end{equation}
Here $N(c_i)$ is a sample of locations with known measurements neighbouring $c_i$, $\gamma_{c_j,c_i}$ is a weight scalar associated with neighbour $c_j$, and $\epsilon_{c_i}$ is the error or residual of the model. Essentially, Equation \ref{eq:kriging} is taking a weighted sum of a sample of neighbouring values. The challenge in kriging is to determine the various weights $\gamma_{c_j,c_i}$, which is what the variogram is used for. The variability of measurements $var(d)$ is modeled as a function of the distance $d$. In its basic form, this results in the following formula:
\begin{equation}
    var(d) = \frac{1}{2 \cdot c(d)} \cdot \sum_{(c_i,c_j) \in D_d} (y_{c_i} - y_{c_j})^{2}
    \label{eq:variogram}
\end{equation}
Here $c(d)$ is the number of times a distance $d$ appeared in the dataset and $\frac{1}{2 \cdot c(d)}$ is a normalisation term. Furthermore, we define $D_d$ as the set of location pairs $(c_i,c_j)$ where $d(c_i,c_j) = d$; that is, the set of all location pairs sharing the same distance between the elements of the pairs. From this, a vector $\mathbf{\gamma}$ of weights $\gamma_{c_j,c_i}$ can be computed by solving the matrix equation 
\begin{equation}
    \gamma = A^{-1} \cdot b
    \label{eq:kriging_weight_solving}
\end{equation}
Here A is matrix of size $(|N(c_i)| \times |N(c_i)|)$ containing $var(d)$ for the distances $d(c_j,c_{j^{*}})$ for all combinations of $c_j$ and $c_{j^{*}}$ in $N(c_i)$, and $b$ is a vector containing $var(d)$ for the distances $d(c_i,c_j)$ for all $c_j$ in $N(c_i)$. Solving Equation \ref{eq:kriging_weight_solving} gives us the weight vector $\mathbf{\gamma}$ needed to perform interpolation using Equation \ref{eq:kriging}.\\
It is worth mentioning that the kriging model suffers from a number of assumptions, most notably stationarity and anisotropy. Moreover, the covariance of the target variable over distance (variogram) is assumed to be constant for the entire dataset. Specific variants of kriging offer different trade-offs in assumptions. We 
we will now describe a selection of these specific kriging variants, although the list will by no means be exhaustive.

First, let us consider \textbf{ordinary kriging} (OK) \cite{cressie1988spatial}. In this variant, the assumption is one of strict stationarity: the (unknown) mean of the target variable is assumed to be constant throughout the entire dataset. To alleviate the problems of strict stationarity, \textbf{universal kriging} (UK) \cite{matheron1969universal} allows for a trend in the mean to exist within the dataset, though it still assumes weak stationarity, as it only allows for a trend in the mean modelled on the coordinates of the location in question. Moreover, Journel and Rossi found that modelling trends in kriging only matters in extrapolation, rather than interpolation \cite{journel1989we}. In \textbf{simple kriging} \cite{webster1987mapping} the mean is known and assumed to be constant, which is then used in its predictions. A final variant worth special attention is \textbf{regression kriging}  \cite{matheron1969krigeage}, also known as kriging after detrending, which bears some resemblance to our later approach as it combines elements of kriging and regression. In this variant, the mean is assumed to have a drift as in universal kriging, 
but unlike universal kriging, this drift is predicted using external explanatory variables rather than coordinates in a separate model. Strongly related would be \textbf{kriging with external drift} \cite{marechal1984kriging}, which combines the drift prediction and kriging into a single form. 

Regardless of the degree to which stationarity and isotropy is assumed by any variant, all kriging-based methods are limited by their reliance on pair-wise distance-based covariance models. Intermediate locations are not considered, and explanatory features are typically only used to compensate for trends in the data. However, depending on the region and dataset in question, kriging can perform very well, and can be applied fairly easily to any grid.

\subsection{Non-kriging methods}
Although the various forms of kriging tend to be the more popular option for spatial interpolation not using external explanatory variables, other methods do exist and are used in popular geographic information system (GIS) software such as ArcGIS \cite{arcgis}. This section will briefly cover a selection of the more popular methods available. Unlike kriging or the later regression-based methods, this set of methods typically does not require model parameters to be trained by data. Instead, neighbouring observations are either used directly (in the case of methods like nearest neighbour), or estimated using a manually defined distance parameter (in inverse distance weighting).

\textbf{Inverse distance weighting} \cite{shepard1968two}.  Much like kriging, Inverse Distance Weighting (IDW) relies on predicting $\hat{y}_{c_i}$ using a weighted sum of a sample of neighbours. Canonically all weights sum to 1. The IDW interpolation formula is thus fairly straightforward:
\begin{equation}
    \hat{y}_{c_i} = \sum_{y^{*}_{c_j} \in \mathbf{Y} \cap \mathbf{Y}_{\varnothing} } \gamma_{c_j} \cdot y_{c_j}
    \label{eq:idw}
\end{equation}
As in the case of kriging in Equation \ref{eq:kriging}, the core IDW formula is based on a weighted sum, with the main challenge being to find the appropriate weights. Unlike the variogram used in kriging, IDW assumes a linear discount over distance (which would also be a model option for the variogram), and thus defines its weights using inverse weighting $\frac{1}{d(c_j,c_i)}$. The final weight $\gamma_{c_j}$ assigned to a location is given by dividing the inverse distance by the sum of all weights, thus guaranteeing that weights will sum to $1$ and bypassing the problem of distance scaling. This leads to the following formula:
\begin{equation}
    \gamma_{c_j} = \frac{\frac{1}{d(c_j,c_i)}}{\sum_{k \in \mathbf{Y} \cap \mathbf{Y}_{\varnothing}} \frac{1}{d(c_k,c_i)}}
\end{equation} 
Beside this basic form, more sophisticated versions of IDW do exist, such as allowing the user greater control through the use of a \textit{power} parameter determining how much relative emphasis is placed on nearby measurements. Recent contributions include the new probabilistic weighting scheme proposed by Łukaszyk \cite{lukaszyk2004new} and the approach proposed by Lu and Wong allowing the weighting parameter to vary, resulting in an adaptive version of IDW \cite{lu2008adaptive}.

\textbf{Nearest neighbour}. A staple of many data-driven applications, nearest neighbour (NN) methods can also be used for spatial interpolation \cite{olivier2012nearest,song2017efficient}. In its simplest form, $\hat{y}_{c_i}$ would simply be equal to $y_{c_j}$ of the nearest $j$ where $y_{c_j} \neq \varnothing$. In the case of ties, one can take the average value of neighbouring locations $\frac{\sum_{c_j \in N(c_i)} y_{c_j}}{|N(c_i)|}$ as an estimate for $\hat{y}_{c_i}$, or select one $y_{c_j}$ at random to represent $\hat{y}_{c_i}$. 
The method could be extended to $k$-NN allowing for the selection of $k$ nearest neighbours, of which the average or most frequent option could be used. 

\textbf{Natural neighbour}. Also known as Sibson interpolation (after its author) \cite{sibson1981brief} or area-stealing interpolation, natural neighbour interpolation is yet another method based on weighted sums; in fact, its basic interpolation formula is identical to Equation \ref{eq:kriging}. As with most weighted sum-based methods, natural neighbour interpolation is distinguished from other methods by how it computes the weights $\gamma_{c_j}$. The main idea of its approach is the use of Voronoi polygons \cite{voronoi1908nouvelles}, also known as Dirichlet regions or Thiessen polytopes \cite{burrough2015principles}. Although the exact construction of these polygons is outside the scope of this project, the basic concept is fairly straightforward. For every known observation $c_k \in \mathbf{Y} : y_{c_k} \neq \varnothing$ a polygon $P_{c_k}$ is created consisting of all points for which $c_k$ is the nearest known observation. Next, when predicting the value of an unknown cell $c_i$, another Voronoi polygon $P_{c_i}$ is created for this point, overlapping with the polygons of its neighbours $c_{j_1},...,c_{j_n}$. The weight $\gamma_{c_j}$ of a neighbouring value $y_{c_j}$ is then computed as the proportion of overlap between $P_{c_i}$ and $P_{c_j}$. This can be conveniently formulated using set notation:
\begin{equation}
    \gamma_{c_j} = \frac{|\{P_{c_i} \cup P_{c_j}\}|}{\sum_{c_k \in N(c_i)} |\{P_{c_i} \cup P_{c_k}\}|}
    \label{eq:natural_neighbours}
\end{equation}
Recent work on natural neighbour interpolation has often focused on improvements to the efficiency of the algorithm, as the computation of Voronoi polygons can be an expensive operation. Examples of this type of work 
include the algorithm proposed by Ledoux and Gold \cite{ledoux2005efficient} and the algorithm proposed by Park et al \cite{park2006discrete}.
\\

\textbf{Splines}. In spline interpolation, the goal is to interpolate the grid in such a manner as to make the resulting landscape (if plotted) as smooth as possible \cite{harder1972interpolation}. The analogy often used to explain the concept of this method is to imagine a sheet of stretchy rubber, anchored at known observations at the height of their values. The sheet would then form a smooth surface between the known observation anchor points, and the heights of the sheets at locations with unknown values would form the estimated value. Moving beyond the analogy, the basic interpolation formula for splines is:
\begin{equation}
    \hat{y}_{c_i} = T(c_i) + \sum_{c_j \in N(c_i)} \gamma_{c_j} \cdot R(c_i,c_j)
    \label{eq:splines}
\end{equation}
While the exact mathematical details of $T(c_i)$ and $R(c_i,c_j)$ are beyond the scope of this work, we refer to the original publication of splines applied to interpolation by Harder and Desmarais \cite{harder1972interpolation} for an in-depth overview. The purpose of $T(c_i)$ is to define a "trend" for $c_i$, which allows the function to have some momentum carried over beyond a point. Similarly, $R(c_i,c_j)$ controls the smoothness of the interpolated surface; in the analogy, one can imagine the contrast between a large rubber sheet with some room for smooth curvature of the surface, and a small rubber sheet pulled taut on all the anchor points, resulting in a very straight surface with sharp edges at anchor points. Finally, $\gamma_{c_j}$ is found by solving a system of linear equations, similar to approaches like kriging in Equation \ref{eq:kriging_weight_solving}. \\
Much like natural neighbour interpolation, much of the more recent work focuses on finding efficient solutions for large-scale datasets, such as the work by Hancock and Hutchinson \cite{hancock2006spatial}. However, as illustrated by the work of Sharifi et al \cite{sharifi2019downscaling}, splines interpolation is also seeing applications in downscaling remote sensing data.

As in the case of kriging, methods like IDW and spline interpolation rely heavily on distance-based interactions between points, and do not account for the characteristics of intermediate locations. Conversely, nearest neighbour and natural neighbour interpolation only consider locations within a pre-defined neighbourhood, and thus do not allow for spatial interaction beyond a certain threshold.

\section{Regression with spatial and explanatory variables}

Let us now turn to common approaches for spatial regression. These methods, unlike purely spatial methods, can deploy external features or explanatory variables in order to predict a target variable, and tend to originate in the field of (geo-)statistics. Note that the set of methods covered in this section is by no means exhaustive, nor intended to be. For a more comprehensive overview of (geo-)statistical methods for spatial prediction tasks, including various regression and interpolation methods, we refer to the 2017 survey by Ziang \cite{jiang2018survey}.\\

\textbf{Basic regression}. When we mention basic or simple regression, we refer to a class of methods used for typical, non-spatial regression. In its simplest from, this can consist of linear regression; that is, 
\begin{equation}
    y_{c_i} = \mathbf{x}_{c_i} \cdot \mathbf{\theta} + \epsilon_{c_i}
    \label{eq:linear_regression}
\end{equation}
, where $\mathbf{\theta}$ is a weight vector set by a training algorithm, $\mathbf{x}_{c_i}$ is the vector of features for location $c_i$, and $\epsilon_{c_i}$ represents the residuals or error of the model. For convenient notation, we add a bias feature of $1$ to $\mathbf{x}_{c_i}$, allowing the corresponding weight from $\mathbf{\theta}$ to represent the intercept of the linear model. Generalising to an entire dataset, the model can be compactly formulated using a vectorised notation:
\begin{equation}
    \mathbf{y} = \mathbf{X} \cdot \mathbf{\theta} + \mathbf{\epsilon}
    \label{eq:linear_regression_vectorised}
\end{equation}
Basic regression is not limited to linear regression, however; any model or algorithm mapping a feature vector to a predicted target variable would fall under this category. A large number of these models, and their training algorithms, are implemented in the popular Python library scikit-learn \cite{sklearn}. A selection of popular regression models implemented in scikit-learn include linear regression using Ordinary Least Squares (OLS) training \cite{montgomery2012introduction}, linear regression using Stochastic Gradient Descent (SGD) training \cite{bottou2010large} and Support Vector Machines (SVMs) \cite{cortes1995support}.\\

\textbf{Spatial Autoregressive Models} \cite{anselin1988spatial}. Unlike simple regression, Spatial Autoregressive (SAR) models explicitly model spatial relationships. They do so by adding an extra term to a linear regression model representing the spatial  autoregression, or autocorrelation over distance, of the target variable. Thus, the general (vectorised) form of SAR models is:
\begin{equation}
    \mathbf{y} = \phi \cdot \mathbf{M} \cdot \mathbf{y} + \mathbf{X} \cdot \mathbf{\theta} + \mathbf{\epsilon}
    \label{eq:sar}
\end{equation}
In Equation \ref{eq:sar}, note that the right-hand side of the summation is simply equal to simple linear regression, multiplying features with their associated weights. The left-hand side, however, introduces new concepts. The first is $\phi$, a scalar weight parameter similar to $\mathbf{\theta}$'s elements scaling the vector produced by $\mathbf{M} \cdot \mathbf{y}$. Here $\mathbf{M}$ is a $H \times W$ matrix, modelling the spatial relationship between instances, which is an important design choice in SAR. In its simplest form, $\mathbf{M}$ would be a sparse matrix with a weight of $1$ for instances that are direct neighbours, and $0$ for instances that are not. In this case, common neighbourhood definitions include Rook (left, right, up, down) and Queen (left, right, up, down, diagonals) \cite{kissling2008spatial}. Another approach would be to assign a distance-based weight between all instances, rather than using a binary neighbourhood definition. In either case, $\mathbf{M}$ is multiplied by the vector of known values $\mathbf{y}$ to get a vector of weighted sums of neighbouring values. Of course, in practice not all true values will be known, in which case one might turn to inputation techniques such as mean substitution or hot-deck imputation \cite{andridge2010review}. Adding the left-hand (spatial) vector to the right-hand (regression) vector results in a final prediction vector. It may be of interest to note that the SAR model is also commonly referred to as the simultaneous autoregressive model (still SAR) or the spatial lag model. 

The basic SAR model has seen various extensions in recent work. For example, Yang et al. 
proposed an extension of the basic SAR model with lagged explanatory variables and cross-variable lags \cite{yang2017identification}. Another, very recent extension to SAR would be the work by Fix et al aimed at being resilient to extremes in areal data \cite{fix2020simultaneous}.

\textbf{Moving average models}.
Moving average (MA) models, more often used in time-series contexts \cite{durbin1959efficient}, can also be used for spatial regression problems \cite{haining1978moving}. The key idea of this approach is to, rather than using the lagged target variable $y_{c_j}$ from a neighbour $c_j$ like in SAR models, use the lagged error $\epsilon_{c_j}$ instead. The intuition behind this approach is that the linear regression term models the general relationship between explanatory variables and the target variable, while the errors shift the location of predictions based on the geographical location. For example, one can consider that the errors of predicting house prices based on the size of the house would differ per neighbourhood, even if the sizes are identical. If the errors of neighbours are high (predicted lower than the true value), we would expect the value at $c_i$ to also be higher than the model would indicate.

We can now define a spatial variant of an MA model as:
\begin{equation}
    \mathbf{y} = \phi \cdot \mathbf{M} \cdot \mathbf{\epsilon}^{(1)} + \mathbf{X} \cdot \mathbf{\theta} + \mathbf{\epsilon}^{(2)}
    \label{eq:ma}
\end{equation}
, where $\mathbf{\epsilon}^{(1)}$ consists of the errors used by the MA model to compute a new prediction, and $\mathbf{\epsilon}^{(2)}$ represents the residuals of the MA model itself. As Equation \ref{eq:ma} shows, the general form of SAR and MA models are very similar; in fact, the only difference is that, rather than using the vector $Y$ of labels for all instances $c_i$, we multiply $\mathbf{M}$ by the vector of prediction errors $\mathbf{\epsilon}^{(1)}$ as our spatial term. A drawback of this approach is that, much like SAR required imputation methods for handling unavailable $y_{c_j}$ for neighbours $c_j$, so too do MA models suffer from the same limitation (as $y_{c_j}$ is needed to compute $\epsilon_{c_j}$). Furthermore, the definition of the spatial term in Equation \ref{eq:ma} is recursive in nature: in order to ascertain the error of a neighbour $c_j$, we'd need to know the errors of $c_j$'s neighbours in turn, which could propagate indefinitely. A possible strategy to deal with this in time-series applications would be to use the average value of the entire time-series as a prediction for the starting observation, thus enabling an initial error measurement. In spatial contexts, which are typically two-dimensional and multi-directional, a common approach is to use the "MA by AR" approach, as introduced by Durbin \cite{durbin1959efficient} and expanded to 2-D contexts by Francos and Friedlander \cite{francos1998parameter}. In this approach the errors are computed by using a regression model for initial predictions, the errors of which are then used to train the "real" MA model. In the original work, the authors used a spatial autoregressive regression model for 
this purpose.

\textbf{Autoregressive moving average models}. SAR and MA models can be used in conjunction, which results in autoregressive moving average (ARMA) models \cite{whittle1951hypothesis} \cite{whittle1963prediction}. These models have been used in various time-series applications, but spatial variants of ARMA models have also been created \cite{haining1978moving}. In a spatial context, this would result in the following equation:
\begin{equation}
    \mathbf{y} = \phi_1 \cdot \mathbf{M} \cdot \mathbf{y} + \phi_2 \cdot \mathbf{M} \cdot \mathbf{\epsilon}^{(1)} + \mathbf{X} \cdot \mathbf{\theta}  + \mathbf{\epsilon}^{(2)}
    \label{eq:sar_and_ma}
\end{equation}
As shown in Equation \ref{eq:sar_and_ma}, a combined SAR and MA model would simply add the spatial terms of both constituent models to the linear regression term. An extra advantage of combining these two models would be that, if accounting for missing initial errors by using those generated by the linear regression term, it allows the MA term to use a more sophisticated (spatial) method to determine $\epsilon^{(1)}_{c_i}$, as the prediction for any instance can be computed using Equation \ref{eq:sar}.\\
A recent early-access paper of particular relevance to the current COVID-19 pandemic is the paper by Qiu et al using ARMA to model a transmission network of infectious disease, specifically influenza \cite{qiu2020spatial}.

\textbf{Geographically weighted regression}. Geographically weighted regression (GWR) is founded on the assumption of spatial non-stationarity \cite{fotheringham1998geographically}. As such, the relationship between features and the target variable can be different for different subsets of the dataset. GWR is based on simple linear regression, but instead of learning one weight vector $\mathbf{\theta}$ for the entire dataset, the weights instead form a surface of potential values. The values of the weights in $\mathbf{\theta}$ are determined by the geographical location of the instance in question. Thus, the equation for this model is:
\begin{equation}
    y_{c_i} = \theta_i^0 + \sum_{k=1}^{|\mathbf{\theta|}} \theta_{c_i}^k \cdot \mathbf{x}_{c_i}^k  + \mathbf{\epsilon}_{c_i}
    \label{eq:gwr}
\end{equation}

Equation \ref{eq:gwr} requires some explanation. The first term, $\theta_{c_i}^0$, is simply a bias term; here $c_i$ represents the coordinates of the instance, and $0$ is the index of the weight in $\mathbf{\theta}$. In the second term, $\sum_{k=1}^{|\theta|} \theta_{c_i}^k \cdot \mathbf{x}_{c_i}^k$, we take the weighted sum of all elements of feature vector $\mathbf{x}$, where the corresponding weight from $\mathbf{\theta}$ is specific to location $c_i$. If we add a bias feature of $1$ to $\mathbf{x}$, we can vectorise Equation \ref{eq:gwr} to:
\begin{equation}
    \mathbf{y} = \mathbf{\theta}_{c_i} \cdot \mathbf{x}_{c_i}  + \mathbf{\epsilon}
    \label{eq:gwr2}
\end{equation}
In Equation \ref{eq:gwr2}, we can see that the only difference with Equation \ref{eq:linear_regression_vectorised} is that it uses a location-specific weight vector $\theta_{c_i}$. The actual values of $\mathbf{\theta}$ are computed using a modified version of OLS, in which a location-specific weight matrix $M_{c_i}$ is used to multiply the features with. Thus, the location-specific weight vector $\theta_{c_i}$ is computed using:
\begin{equation}
    \theta_{c_i} = (\mathbf{X}^{T} \cdot \mathbf{M}_{c_i} \cdot \mathbf{X})^{-1} \cdot \mathbf{X}^{T} \cdot {\mathbf{M}_{c_i}} \cdot \mathbf{y}
    \label{eq:gwr3}
\end{equation}
Equation \ref{eq:gwr3} involves many matrix multiplications, and is in fact simply OLS weight fitting ($(\mathbf{X}^{T} \cdot \mathbf{X})^{-1} \cdot \mathbf{X}^{T} \cdot \mathbf{y}$) with multiplications by $\mathbf{M}_{c_i}$ added in. The manner of determining the weight matrix $\mathbf{M}_{c_i}$ is referred to as a \textit{kernel}, and can take various forms such as uniform, Gaussian, or exponential functions.\\

Finally, deep learning and \textbf{convolutional neural networks} (CNN) in particular have been used to great effect in many computer vision applications, such as interpolation-based image and video resolution upscaling  \cite{dong2015image,shi2016real}. Such computer vision-based interpolation CNNs could also be applied to general spatial interpolation by replacing the vector of $3$ RGB values of an image by the spatial feature vector $\mathbf{x}$. Moreover, in their 2020 publication, Hashimoto and Suto formulated a CNN architecture for the specific purpose of spatial interpolation \cite{hashimoto2020sicnn}. 

As in the case of nearest neighbour and natural neighbour interpolation, this category of methods is limited by their use of a pre-defined local neighbourhood, dismissing information outside of the neighbourhood radius. 

\section{Summary and evaluation of related work}


Though not without merits, every method listed in this chapter suffers from its own set of limitations. We will start with the purely spatial methods.

In the case of kriging, the pair-wise distance-based weight computation is limited, as the spatial characteristics of intermediate locations could be highly relevant. Kriging also typically assumes isotropy and stationarity -- although some variants of kriging allow for the existence of trends in the data, weak stationarity is still assumed. Inverse distance weighting, though useful for a quick solution to spatial interpolation, suffers from all shortcomings of kriging, while being less flexible due to its use of purely distance-based weights rather than a variogram. Nearest neighbour interpolation is likewise simple to use, and therefore useful when quick rough solutions are needed, but suffers from its static, predefined neighbourhood definition, and will be very sensitive to extreme values in these neighbourhoods. Natural neighbour, while elegant in its approach of assigning weights to neighbours, is limited to purely local spatial effects of direct neighbours. Spline interpolation can be somewhat complex to apply and requires many parameters to be tuned effectively; moreover, its interpolation is rather distance-based and does not consider intermediate locations.

Moving on to regression methods capable of incorporating explanatory variables and spatial effects, we will start with basic regression. Given that basic regression uses purely local explanatory features to predict target variables directly, the lack of any spatial effects and the resulting waste of information from nearby known measurements would be its greatest weakness. SAR, MA and ARMA, while all subtly different, rely on rigidly defined neighbourhoods through the weight matrix $\mathbf{M}$. Although the weight matrix does allow for neighbourhoods to be flexibly assigned, they are rather rigid once set. In the case of MA and ARMA, determining the prediction errors $\epsilon^{(1)}$ is not trivial since it requires the prediction model to have already made predictions (in order to compute errors) prior to training the model. Though there are strategies to address this problem, all of these may introduce additional uncertainty to the model. 
Much like basic regression, geographically weighted regression disregards potentially valuable information from the known values of neighbouring cells. Although it does account for spatial effects by changing the regression weights based on the geographical location in question, its spatial effects are limited. Finally, convolutional neural networks will typically contain a large amount of trainable parameters, especially deep CNNs, which means training the models may require more time, as well as more data, than other types of models; this latter requirement may be particularly problematic in our case, where we typically only have a few hundred training examples available in a training set. Moreover, since the performance of neural networks can be heavily dependent on the architecture of the network, CNNs may require further resource-intense neural architecture search (NAS) to perform well. 
Moreover, all CNNs will rely on strictly defined neighbourhoods in their convolutional kernels, thus dismissing potentially relevant information from outside these neighbourhoods.

In the next chapter, we will introduce our proposed methods aimed at overcoming the limitations of the existing work.

\chapter{Proposed methods}
\label{chap:methods}

In this section, we will first explain the reinforcement learning background of our proposed method, after which we will explain the general idea behind the use of these techniques for spatial interpolation. Next, we propose three variants of MRP interpolation, and 
explain the assumptions and properties of each variant. The creation of every variant was motivated by the need for variety in terms of applicability, as well as stationarity and isotropy assumptions. 

\section{Markov reward processeses}
Markov reward processes (MRPs) are important to the field of reinforcement learning (RL), as they form the basis for the widely used Markov \textit{decision} processes \cite{bellman1957markovian} (MDPs). The main idea of an MRP is that it models a \textit{state space} $S$ an \textit{agent} (such as a robot or a video game character) can find itself in. States $s \in S$ could be described using various different characteristics, but a simple case conveniently close to our problem would be states described solely by a location. In every state $s$, the agent has a number of actions $a(s)$ available to it, and every action $a \in a(s)$ would transition the agent from state $s$ to a successor state $s'$, resulting in an immediate reward $R_a$ for the agent for taking action $a$. Using this information, a \textit{state value} $v(s)$ for state $s$ could be computed for every state, by assigning the average reward of all actions $a \in a(s)$ as a state value $v(s)$. However, because it can be useful to look further into the future than only considering the immediate reward $R_a$, the estimated state value of the successor state $s'$ should also be taken into account: an action $a$ with a high $R_a$ might take the agent to a successor state $s'$ from which no further rewards can be acquired, which may not be desirable. Thus $v(s)$ is computed using the average \textit{action value} $v(a) = R_a + v(s')$ instead, which adds the estimated value of $s'$ as an expected future reward to $R_a$. At the same time, in many cases the relative certainty of receiving a reward early on may be 
more important than a less certain reward in the future. To model this, MRPs incorporate a \textit{discount} parameter $\gamma$, which controls the degree to which an agent is near-sighted (low $\gamma$, future rewards have a low weight) or far-sighted (high $\gamma$, future rewards have a high weight). Thus the action value $v(a)$ can be computed as $v(a) = R_a + \gamma \cdot v(s')$, leading to the following recursive definition of state value $v(s)$:
\begin{equation}
    v(s) = \frac{\sum_{a \in a(s)} R_a + \gamma \cdot v(s')} {|a(s)|}
    \label{eq:mrp_rl}
\end{equation}
Of course, since \textit{estimated} successor state values are used in this recursive formula rather than known values, Equation \ref{eq:mrp_rl} must be iterated until an equilibrium of stable values is reached. Once the state values are known, an MDP would extend this by defining an optimal \textit{policy} (actions to take given a state) based on the computed state values; however, we are only interested in the state values themselves, which is why we only use MRPs and not MDPs in our method.

\section{Markov reward processes for spatial interpolation}
\begin{figure}[]
  \centering
  \includegraphics[width=0.65\linewidth]{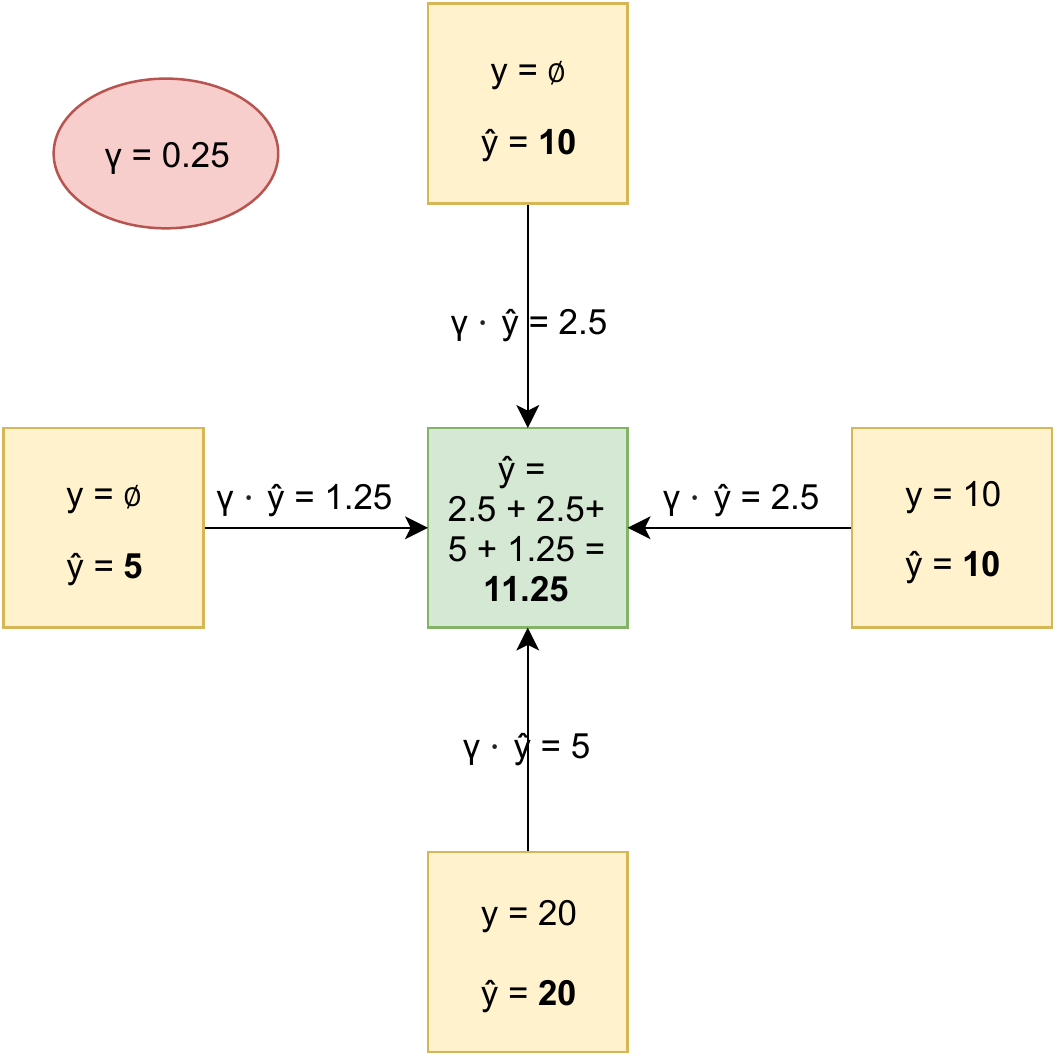}
  \caption{Simple example of an $\hat{y}$ update for a single cell (green). The estimated values of its neighbours (yellow) are discounted by $\gamma = 0.25$ and summed (optionally averaged) to determine a new $\hat{y}$. This process is repeated for every cell in the MRP in one iteration.}
  \label{fig:mrp_idea}
\end{figure}

In our problem definition, we are no longer dealing with an agent, but interpret a state $s$ 
as a location $c$, an action $a$ as a spatial interaction between locations, and the state value $v(s)$ as a predicted target value $\hat{y}$, which is the value we are ultimately interested in.
In MRPs, the main value being computed is the state value $v(s)$, here $\hat{y}_c$, based on the expected future reward of a state (cell) $c$. As before, let us define a cell of interest $c_i = (h,w)$ and an arbitrary neighbouring cell $c_j = (h^{*},w^{*})$. For all $c_i \in \mathbf{Y}$, the estimated state value is computed using the summed (or average in canonical RL) estimated neighbouring values $\hat{y}_{c_j}$, weighted by a \textit{discount} parameter $\gamma$, for all $c_j \in N(c_i)$ (note that, in the RL analogy, we are only using $v(s')$ and not $R_a$). 
Here $N(c_i)$ is the set of direct neighbours 
to $c_1$, where neighbourhoods can be defined flexibly, but a simple strategy would be to use shared borders (rook neighbourhoods). This leads to a recursive definition of $\hat{y}_{c_i}$:
\begin{equation}
\label{eq:mrp_interp_sv}
    \hat{y}_{c_i} = \sum_{c_j \in N(c_i)} \gamma \cdot \hat{y}_{c_j}
\end{equation}
As a result of this recursive definition, all estimations depend on other estimations in turn, which means a single computation of Equation \ref{eq:mrp_interp_sv} per cell will not be sufficient to estimate $y_{c_i}$. Therefore, Equation \ref{eq:mrp_interp_sv} is iterated until an equilibrium is reached.\\
Intuitively, in spatial interpolation the weighted sum of neighbouring predictions represents a local ``spatial lag'', similar to SAR and many other spatial autocorrelation-based methods. However, the iterative use of recursive computations of $\hat{y}_{c_j}$, rather than known values $y_{c_j}$, sets MRPs apart from other methods by modelling a local computation of a recursive spatial lag. This recursivity allows every location $c_i$ to implicitly influence every other location $c_j$, through all possible paths between $c_i$ and $c_j$.\\ 
Figure \ref{fig:mrp_idea} shows the basic idea of MRP interpolation. In the next sections, we will explain the three MRP variants in detail. The differences between these three variants are determined by how they compute the discount weights between $c_j$ and $c_i$. 
\par

\subsection{Static discount MRP.}
The static discount MRP (SD-MRP) is the MRP interpolation variant closest to a canonical MRP as used in RL. If the true target value $y_{c_i}$ at location $c_i$ is already known, we use this as a static prediction for $\hat{y}_{c_i}$; if it is not known, we compute an estimated $\hat{y}_{c_i}$ using Equation \ref{eq:mrp_interp_sv}. By iterating Equation \ref{eq:mrp_interp_sv} until equilibrium, we obtain a matrix with interpolated values of all cells, anchored by information from the cells $c$ where the true values $y_{c}$ are known. \\
An advantage of using this basic version of MRP interpolation would be that it requires no explanatory variables, which makes it very straightforward to apply in any scenario. Moreover, its non-reliance on additional data means it does not need a training phase, although it is advisable to tune $\gamma$ either through the use of a separate training set or subsampled true target values of the dataset itself. The main disadvantage would be that it uses a static, global $\gamma$ discount parameter, which means it assumes stationarity and isotropy.

\subsection{Optimised MRP.}

Given the assumptions of stationarity and isotropy of SD-MRP, we were interested in seeing whether we could create MRP interpolation methods robust to non-stationarity and anisotropy. To this end, we would like to use location-specific weights $\gamma_{c_i,c_j}$ instead of a static weight $\gamma$. The challenge, then, would be to find a method to reliably find a weight vector $\Gamma$, consisting of individual weights $\gamma_{c_i,c_j}$ 
for all neighbouring cells in $\mathbf{Y}$, such that the total interpolation error is minimal. If $y^{*}_{c}$ is known for all $c \in \mathbf{Y}$, we are free to use black-box optimisation to directly find the set of optimal weights $\Gamma^{*} = \{\gamma^{*}_{c_j,c_i} : (c_i, c_j \in \mathbf{Y}) \wedge (c_j \in N(c_i)) \}$ using interpolation loss on a selection of artificially hidden nodes as a loss function $\mathcal{L}$:
\begin{equation}
\label{eq:mrp_optimisation}
    \Gamma^{*} \in \argmin_{\Gamma \in \mathbb{R}^{|\Gamma|}} \mathcal{L}(\mathbf{Y},\Gamma) \\
\end{equation}
However, if not all $y^{*}_{c}$ are known, the weights to and from these cells cannot be optimised with any certainty. Moreover, these highly customised optimal weights will not be transferable to other regions, as they reflect the landscape of the region they were optimised for. As a result, optimised MRP (O-MRP) is not practically applicable in most situations: fully observable regions need no interpolation, and the weights cannot be re-used except in very specific circumstances (e.g., recurrent measurements that may be not be complete immediately, or the inference of the true values of noisy measurements). 

\begin{algorithm}[t]
    \small 
    \SetAlgoLined
    \KwInput{Training set matrices $\mathbf{X}^{train}$ and $\mathbf{Y}^{train}$, test set matrix $\mathbf{X}^{test}$ and $\mathbf{Y}^{test}$, maximum MRP iterations $max\_iter$}
    \KwResult{Interpolated matrix $\mathbf{\hat{Y}}$} 
    $\Gamma^{*} = optimise\_weights(\mathbf{Y}^{train})$\\
    $\mathcal{M}_{w}$ := $fit\_model(X^{train},\Gamma^{*})$ \\
    $iter := 0$\\
    \While{$iter < max\_iter$}{
        \ForAll{$c_i \in \mathbf{Y}^{test}$} {
            \uIf{$y_{c_i} \neq \varnothing $} {
                $\hat{y}_{c_i} := 0$\\
                \ForAll{$c_j \in N(c_i)$} {
                    $\hat{y}_{c_i} := \hat{y}_{c_i} + \mathcal{M}_{w}(\mathbf{x}^{test}_{c_j,c_i}) \cdot \hat{y}_{c_j}$ 
                }
            } 
            \Else {
                $\hat{y}_{c_i} := y_{c_i}$
            }
        }
        $iter := iter+1$
    }
    $\mathbf{\hat{Y}} := \hat{y}_{c}$ for $c \in \mathbf{Y}^{test}$ \\ 
    \textbf{return} $\mathbf{\hat{Y}}$
    
    \caption{Full pipeline for WP-MRP}
    \label{alg:wp_mrp}
\end{algorithm}

\subsection{Weight prediction MRP.}
In an ideal case, we would use a method allowing us to exploit the benefits of optimal location-specific weights, regardless of where the MRP is used. This has led to the final version of MRP interpolation, in which we use the spatial feature vectors $\mathbf{x}_{c_i}$ and $\mathbf{x}_{c_j}$ as inputs to a weight prediction model to predict the optimal weight $\gamma^{*}_{c_i,c_j}$ from spatial data describing the locations (such as houses, shops and land use), rather than optimisation. 
If successful, this weight prediction MRP (WP-MRP) would combine the greater accuracy and resillience to non-stationarity and anisotropy from O-MRPs, and the broad applicability from SD-MRPs (provided explanatory variable data is available).

To train the weight prediction model, we run O-MRP on a training region to acquire an optimal weight configuration. Next, we create a feature vector $\mathbf{x}_{c_i,c_j}$ by combining $\mathbf{x}_{c_i}$ and $\mathbf{x}_{c_j}$ for every $(c_i,c_j)$ pair in $\mathbf{Y}$. For every location pair, $\mathbf{x}_{c_i,c_j}$ is matched with the optimised weight $\gamma^{*}_{c_i,c_j}$, resulting in a tabular training set $\mathbf{X}$ with optimal weights $\Gamma^{*}$ as a ground truth. We can thus define a regression model $\mathcal{M}_{w}(\mathbf{x})$, such that, if $\mathbf{Y}_{N} = \{(c_i,c_j) : (c_i,c_j \in \mathbf{Y}) \wedge (c_j \in N(c_i) )\}$:
\begin{equation}
\label{eq:mrp_weight_prediction}
    \mathcal{M}_{w}^{*} \in \argmin_{\mathcal{M}_{w}} \frac{1}{|\Gamma^{*}|} \sum_{(c_i,c_j) \in \mathbf{Y}_{N}} (\mathcal{M}_{w}(\mathbf{x}_{c_i,c_j}) - \gamma^{*}_{(c_i,c_j)})
\end{equation} 

Here we train $\mathcal{M}_{w}(\mathbf{x})$ on $\Gamma^{*}$ using any regression (machine learning) algorithm. The full pipeline of WP-MRP is outlined in Algorithm \ref{alg:wp_mrp} (which assumes available functions for black-box optimisation and model fitting; any algorithm of choice for either task can be used). Line 1 runs O-MRP on a training set, and line 2 fits a weight prediction model to the optimal weights found by O-MRP. Lines 3-16 show the iterative updates of cells in $\mathbf{Y}$, and lines 17-18 create and return the predictions in the form of a grid $\mathbf{\hat{Y}}$. 

\chapter{Experiments}
\label{chap:experiments}
We were interested in answering two main questions with our experiments:
\begin{itemize}
    \item Can MRP interpolation achieve lower mean absolute prediction errors for various proportions of missing values compared to baseline methods?
    \item To what extent could these models be generalised and transferred across national and international regions?
\end{itemize}
In the case of the first question, while we were primarily interested in finding out which method would have the best performance, we wished to assure that any performance differences were not caused by a susceptibility to the proportion of known $y^{*}$ values. This also led to the question of transferability: if data is scarce for one region, but plentiful for another, it could be useful to be able to take advantage of the large amounts of available data in the latter region to aid predictions in the former. 

\section{Baselines}
Our selection of baselines was aimed at including competitive interpolation and regression methods used for spatial and geo-spatial modelling in practice. The selection we made consisted of: 
\begin{itemize}
    \item \textbf{Ordinary kriging} (OK) 
    and \textbf{universal kriging} (UK), as implemented by the Python library PyKrige \cite{pykrige}. The general prediction model for kriging derived from Equation \ref{eq:kriging}, with weight parameters solved using Equations \ref{eq:variogram} and \ref{eq:kriging_weight_solving}, can be formulated as: 
    \begin{equation}
        \hat{y}_{c_i} = \sum_{c_j \in N(c_i)} (\gamma_{c_j,c_i} \cdot y_{c_j})
        \label{eq:kriging_prediction}
    \end{equation}
    \item \textbf{Non-spatial (basic) regression}, as described by Equations \ref{eq:linear_regression} and \ref{eq:linear_regression_vectorised}, but using auto-sklearn \cite{feurer2019auto} to automatically select the best performing conventional regression model. Since we are now explicitly estimating $\hat{y}$, our prediction model based on Equation \ref{eq:linear_regression_vectorised} becomes:
    \begin{equation}
        \mathbf{\hat{y}} = \mathbf{X} \cdot \mathbf{\theta}
        \label{eq:linear_regression_vectorised_pred}
    \end{equation}
    \item \textbf{Spatial autoregressive-} (SAR), \textbf{moving average-} (MA) and \textbf{autoregressive moving average} (ARMA) models, with prediction models based on Equations \ref{eq:sar}, \ref{eq:ma} and \ref{eq:sar_and_ma}, and using the ``MA by AR'' approach \cite{haining1978moving} for MA and ARMA. In the interest of fairness, we implemented these models manually, since this allowed us to use auto-sklearn as in the case of basic regression. The terms resulting from $\mathbf{m}_c \cdot \mathbf{y}$ and $\mathbf{m}_c \cdot \mathbf{\epsilon}$ thus became additional features supplementing $\mathbf{x}$, and $\phi_1$ and $\phi_2$ became additional weights to train. The prediction models for estimating $\hat{y}$ for SAR (Equation \ref{eq:sar_pred}), MA (Equation \ref{eq:ma_pred}) and ARMA (Equation \ref{eq:arma_pred} are thus formulated as:
    \begin{equation}
        \mathbf{\hat{y}} = \phi \cdot \mathbf{M} \cdot \mathbf{y} + \mathbf{X} \cdot \mathbf{\theta}
        \label{eq:sar_pred}
    \end{equation}
    \begin{equation}
        \mathbf{\hat{y}} = \phi \cdot \mathbf{M} \cdot \mathbf{\epsilon} + \mathbf{X} \cdot \mathbf{\theta}
        \label{eq:ma_pred}
    \end{equation}
    \begin{equation}
        \mathbf{\hat{y}} = \phi_1 \cdot \mathbf{M} \cdot \mathbf{y} + \phi_2 \cdot \mathbf{M} \cdot \mathbf{\epsilon} + \mathbf{X} \cdot \mathbf{\theta}
        \label{eq:arma_pred}
    \end{equation}
    \item \textbf{Convolutional neural networks} (CNN), using automated neural architecture search (NAS) implemented by auto-keras \cite{jin2019auto} for all training sets (50 trials, 1000 epochs). 
\end{itemize}

\section{Data}

Our interpolation setup requires data of two types: a matrix $\mathbf{Y}$ of ground truth target values, and (depending on the method used) spatial features $\mathbf{x}_{c}$ for all $c \in \mathbf{Y}$. Consequently, our data collection process for evaluating this approach was also split into these two categories.

\subsection{Ground truth values}

\begin{figure}[t]
     \centering
     \begin{subfigure}[b]{0.28\textwidth}
         \centering
         \includegraphics[width=\textwidth]{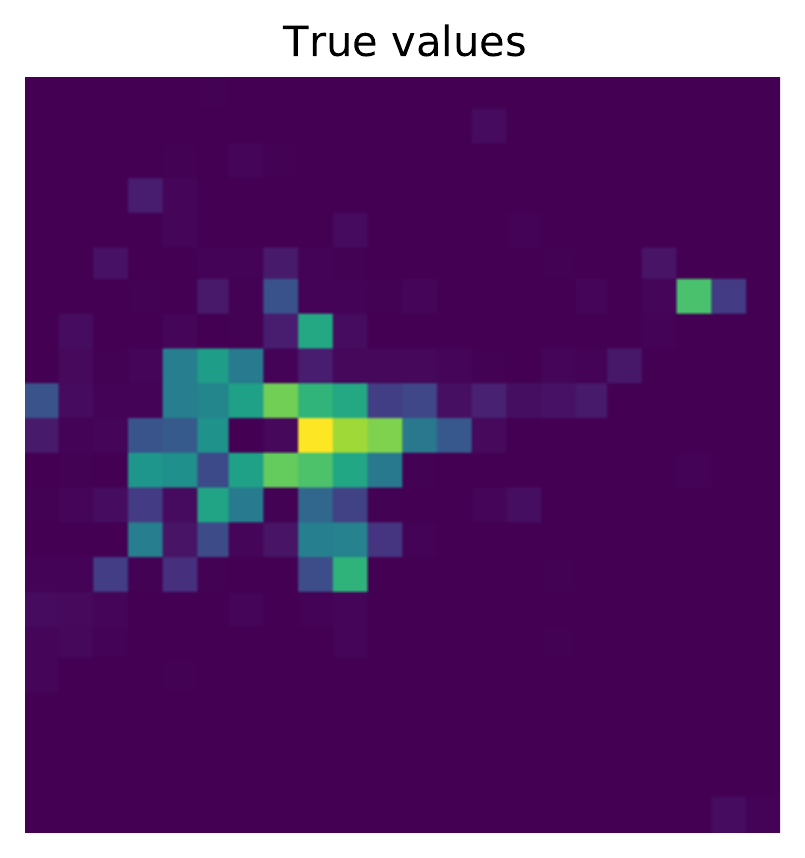}
         \caption{True values}
         \label{fig:true}
     \end{subfigure}
     \hfill
     \begin{subfigure}[b]{0.28\textwidth}
         \centering
         \includegraphics[width=\textwidth]{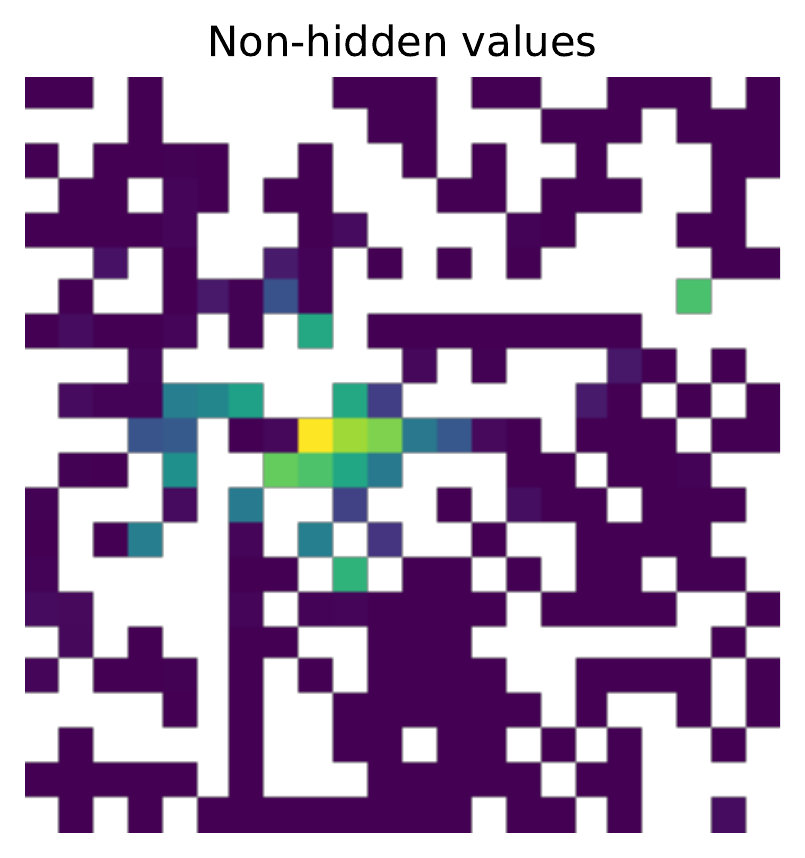}
         \caption{Non-hidden values}
         \label{fig:nonhidden}
     \end{subfigure}
     \hfill
     \begin{subfigure}[b]{0.345\textwidth}
         \centering
         \includegraphics[width=\textwidth]{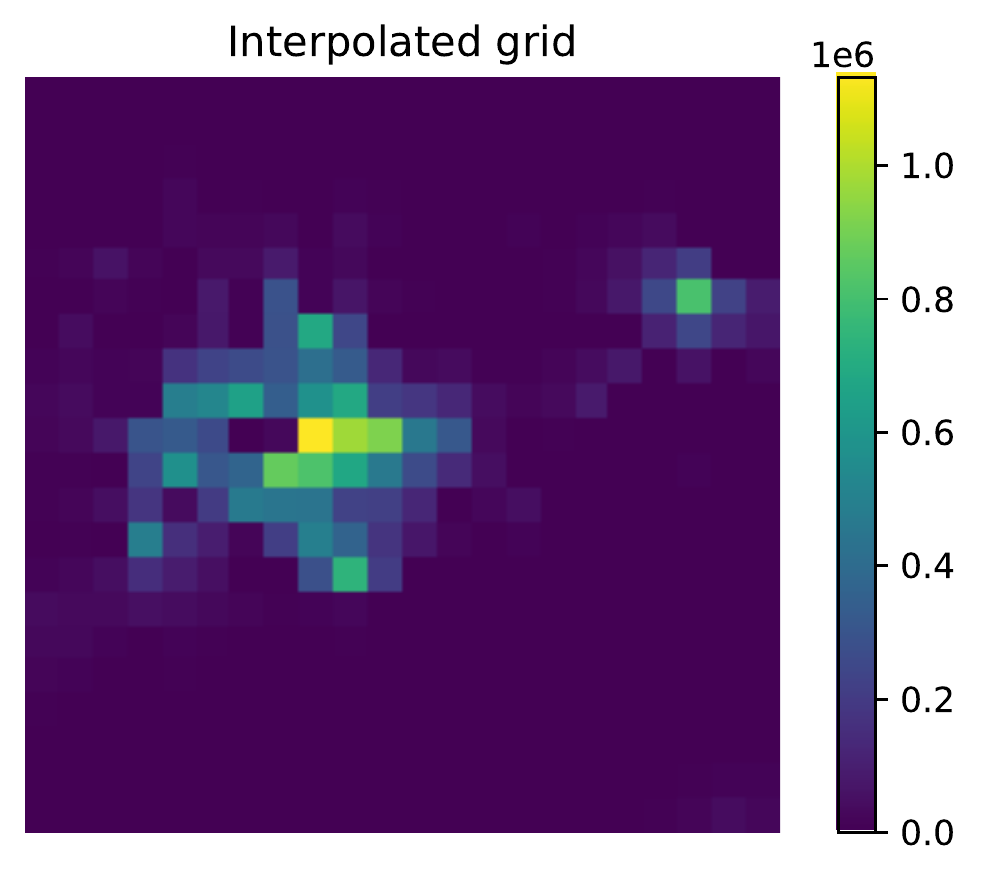}
         \caption{Interpolated grid}
         \label{fig:interpolated}
     \end{subfigure}
        \caption{Taipei GDP grid, consisting of (a) ground truth values, (b) ground truth values where half were artificially hidden, and (c) an interpolated grid combining ground truth values for non-hidden cells with SD-MRP-predicted hidden values.}
        \label{fig:ground_truth_preprocessing} 
\end{figure}

For our ground truth data, we chiefly used a gridded dataset containing world-wide GDP estimates sourced from World Bank \cite{gdp_world_bank}. Although this dataset was gridded already, for greater flexibility we defined our own matrix with a fixed cell width and height of $0.02^{\circ}$ (approximately 1 kilometer in the experimental regions we used), and averaged the values from the dataset to the appropriate cell in $\mathbf{Y}$. As the GDP dataset is fully available at every location in the world, it was very convenient for the evaluation of our method. However, as an indication of the generalisibility of our results, as well as an interesting test case for a potential application of our method, we also used a dataset on COVID-19 patient trajectories in South Korea \cite{covid_dacon}. We aggregated the locations in the trajectories as the total number of visits by infected patients per grid cell, a meaningful variable to be able to predict for virus containment policies. 
To evaluate our methods, we artificially hid a number of true values in $\mathbf{Y}$, allowing us to have access to both $y^{*}_{c}$ and any number of locations where $y_{c} = \varnothing$ (missing values). 
This process was controlled by the parameter $p$, representing the proportion of values in a dataset that is artificially hidden. An example of the ground truth preprocessing pipeline can be found in Figure \ref{fig:ground_truth_preprocessing}. In Figure \ref{fig:true}, the original ground truth data is displayed, whereas Figure \ref{fig:nonhidden} shows the remaining true values after artificially hiding values (at $p=0.5$). Finally, Figure \ref{fig:interpolated} shows the resulting reconstructed original values by running SD-MRP on the values from Figure \ref{fig:nonhidden}.

\subsection{Spatial features} 

\begin{figure}[t]
     \centering
     \begin{subfigure}[b]{0.45\textwidth}
         \centering
         \includegraphics[width=\textwidth,height=0.285\textheight]{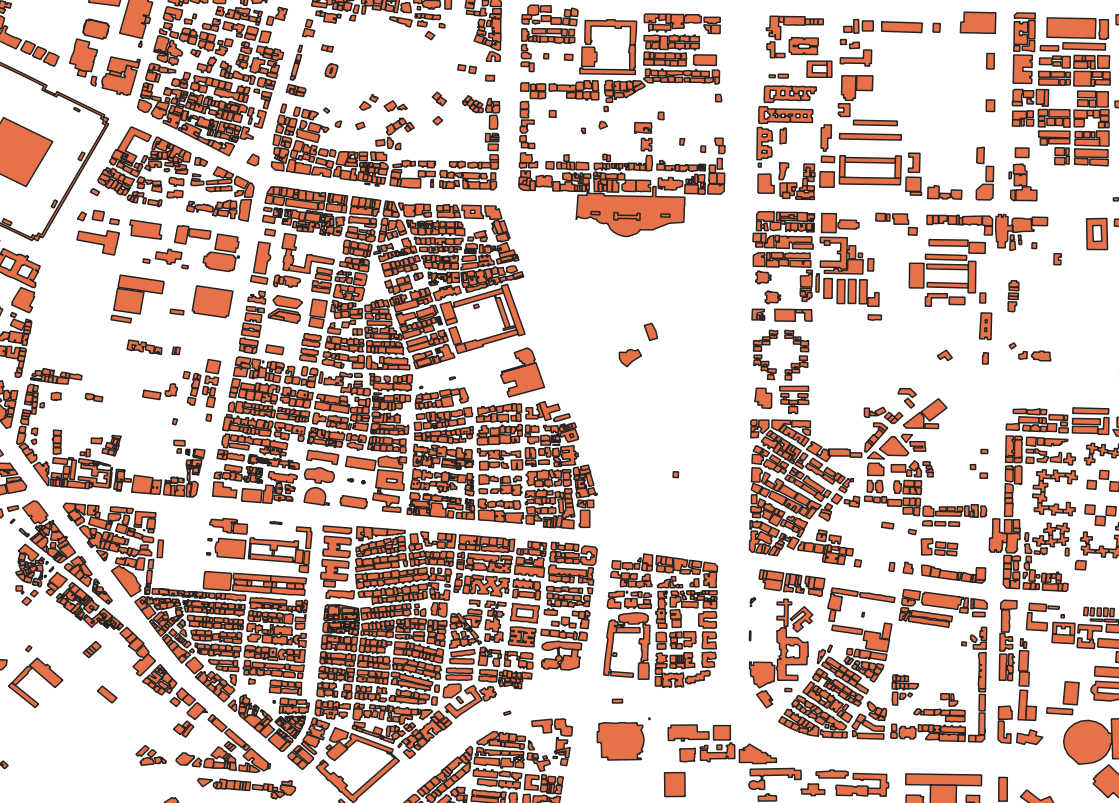}
         \caption{Example map data for part of the Taipei city centre \cite{osm}}
         \label{fig:osm}
     \end{subfigure}
     \hfill
     \begin{subfigure}[b]{0.45\textwidth}
         \centering
         \includegraphics[width=\textwidth]{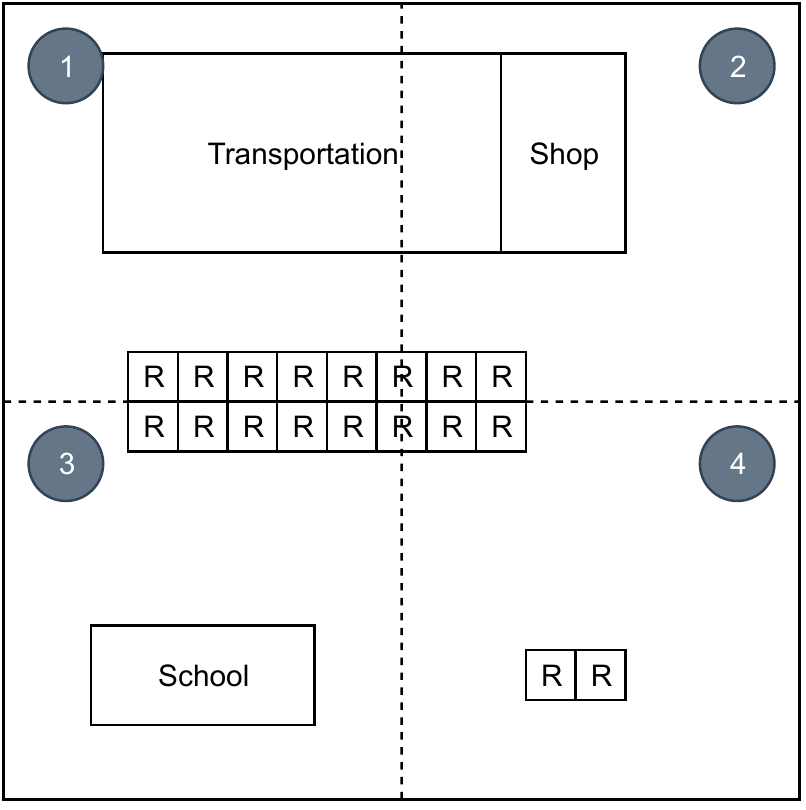}
         \caption{Illustration of splitting up map data into four locations}
         \label{fig:aggregation}
     \end{subfigure}
     \hfill
     \begin{subfigure}[b]{0.8\textwidth}
        \vspace{0.5cm}
        \centering
        \begin{tabular}{|l|llll|}
        \hline
        \textbf{Location} & \multicolumn{1}{l|}{\textbf{Transportation}} & \multicolumn{1}{l|}{\textbf{Residential}} & \multicolumn{1}{l|}{\textbf{Shop}} & \textbf{School} \\ \hline
        1                 & 0.8                                          & 5.5                                       & 0                                  & 0               \\ \hline
        2                 & 0.2                                          & 2.5                                       & 1                                  & 0               \\ \hline
        3                 & 0                                            & 5.5                                       & 0                                  & 1               \\ \hline
        4                 & 0                                            & 4.5                                       & 0                                  & 0               \\ \hline
        \end{tabular}%
         \caption{Features per location from Figure \ref{fig:aggregation} from type counts}
         \label{fig:features}
     \end{subfigure}
        \caption{Preprocessing pipeline for spatial features.}
        \label{fig:spatial_features}
\end{figure}

As spatial features for $\mathbf{x}$, we aggregated a selection of vector and point map data sourced from OpenStreetMap \cite{osm}. For all $c \in \mathbf{Y}$, every element in $\mathbf{x}_{c}$ represented the count of all objects in the map data corresponding to a certain \textit{type}, such as apartments, houses and shops. An example of the map data can be found in Figure \ref{fig:osm}, and the process of turning this map data into feature vectors for individual locations is illustrated in Figures \ref{fig:aggregation} and \ref{fig:features}. There are various design choices available for preprocessing and feature selection on this type of data; a summary of our preprocessing options can be found in Table \ref{tab:preprocessing}. 

\begin{table}[t]
\resizebox{\textwidth}{!}{%
\begin{tabular}{|l|l|l|}
\hline
\textbf{Problem}                                & \textbf{Design option}  & \textbf{Design option explanation}                                                                                                                                 \\ \hline
\multirow{2}{*}{\textbf{Missing values}}        & drop                    & \begin{tabular}[c]{@{}l@{}}Drop all objects from a spatial dataset if\\ they have no identifiable  object type.\end{tabular}                                               \\ \cline{2-3} 
                                                & replace                 & \begin{tabular}[c]{@{}l@{}}Add objects from a spatial dataset without an\\ identifiable type as objects of the type "generic".\end{tabular}                           \\ \hline
\multirow{5}{*}{\textbf{Feature selection}}     & top $n$                 & \begin{tabular}[c]{@{}l@{}}Only use the $n$ most frequently occurring\\ object types to construct $x$.\end{tabular}                                                       \\ \cline{2-3} 
                                                & top $n$ percent         & \begin{tabular}[c]{@{}l@{}}Only use the top $n$ percent of most frequent\\ object types to construct $x$.\end{tabular}                                                    \\ \cline{2-3} 
                                                & top $n$ variable        & \begin{tabular}[c]{@{}l@{}}Only use the top $n$ object types based on the\\ variance of their frequency (computed beforehand)\\ to construct $x$.\end{tabular}                                         \\ \cline{2-3} 
                                                & frequency threshold     & \begin{tabular}[c]{@{}l@{}}Only use object types of which their relative\\ frequency exceeds a threshold to construct $x$.\end{tabular}                                   \\ \cline{2-3} 
                                                & taxonomy                & \begin{tabular}[c]{@{}l@{}}Map low-level object types to manually specified\\ higher-level categories to construct $x$.\end{tabular}                                      \\ \hline
\multirow{3}{*}{\textbf{Feature normalisation}} & mean normalisation      & \begin{tabular}[c]{@{}l@{}}Scale feature $f$ using the mean, max and min\\ value of $f$ in the dataset: $f := \frac{f - mean(f)}{max(f) - min(f)}$\end{tabular} \\ \cline{2-3} 
                                                & unit length scaling     & Scale feature vector $x$ to unit length: $x := \frac{x}{|x|}$                                                                                                      \\ \cline{2-3} 
                                                & $z$-score normalisation & \begin{tabular}[c]{@{}l@{}}Scale feature $f$ using the mean and standard\\ deviation of $f$ in the dataset: $f := \frac{f - mean(f)}{\sigma}$\end{tabular}      \\ \hline
\end{tabular}%
}
\caption{Overview of design options for the preprocessing pipeline of spatial features. Objects are individual shapes seen in Figure \ref{fig:aggregation}(a), object types are the different functions objects may have, as illustrated in \ref{fig:aggregation}(b) and (c), and a feature $f$ is a single element of feature vector $x$. Missing values refers to the manner in which geographical objects in a spatial dataset with no identifiable type attribute are handled, feature selection refers to strategies to keep or discard features based on various selection criteria, and feature normalisation refers to various methods to normalise the locations and ranges of features.} 
\label{tab:preprocessing}
\end{table}

In accordance with the design philosophy of programming by optimisation (PbO) \cite{hoos2012programming}, we did not commit to any of these choices, and instead used automated algorithm configuration to select the best possible feature construction pipeline per method. To this end, we ran the Baysian optimisation-based algorithm configurator SMAC \cite{smac} 
for 36 hours for all methods on a machine with an Intel Xeon E5-2683 v4 CPU running at 2.10GHz. This ensured that we were evaluating different methods using (close to) optimal configurations of preprocessing options (as well as algorithm hyperparameters, if applicable) specific to those methods.   

\subsection{Regions}

As test regions we used the city Daegu in South Korea and Taipei in Taiwan. For both cities, we ran experiments for three \textit{transferability conditions}: the available data in the city itself (\textit{same-city}), transferred from another city in the same country (\textit{same-country}), or transferred from a city in the other city's country (\textit{different-country}). Seoul was used as an extra training city in South Korea, and Taichung was used for Taiwan. 
\section{Experimental setup}
The basic form of our experiment consisted of executing $30$ runs of every method, at proportion of hidden values $p$ 
settings of $0.1$, $0.3$, $0.5$, $0.7$ and $0.9$, for Daegu and Taipei. A single run of training and testing a model for a particular condition, 
when involving optimisation, took about a half hour to complete, whereas methods such as kriging and SD-MRP took less than a second or a few seconds to run, respectively. 
For relevant baselines as well as WP-MRP, we ran auto-sklearn \cite{feurer2019auto} with a budget of 150 seconds to automatically select the best performing machine learning model and hyperparameters. For O-MRP and WP-MRP training, we used the covariance matrix adaptation evolution strategy (CMA-ES) algorithm as implemented in the Python CMA package \cite{hansen2019pycma}, using an iteration budget of 100 (early stopping allowed), to perform black-box optimisation.

We first ran this experimental setup for the same-city condition. Next, to test for transferability, we re-ran this experiment with same-country and different-country transferability conditions. In the case of the South Korean COVID-19 dataset, we were only able to test for national transferability. In the case of kriging, training regions were only used to select the best performing variogram model, and in the case of SD-MRP, they were only used to tune the global discount paramter $\gamma$. Finally, we should note that O-MRP can only be applied directly to the test region and therefore has no need of a training set; moreover, as explained in Chapter \ref{chap:methods} 
O-MRPs are mainly used as a theoretical lower bound for errors that could be achieved with MRP methods. 

For our performance comparisons we performed normality tests on the $30$-run samples for all conditions. 
If normally distributed, we ran one-tailed t-tests to establish statistical significance for performance improvements between methods. If one of the samples was not normally distributed, we ran a Wilcoxon signed-rank test instead, as it can function as an alternative to the t-test when comparing the mean of two samples for which a normal distribution cannot be assumed \cite{wilcoxon1992individual}. 
For all tests, we set $\alpha=0.05$. Methods were ranked based on the number of other methods they significantly outperformed, with a special emphasis placed on the top-3 methods and the worst-performing method.

\chapter{Results}
\label{chap:results}

All results of our experiments can be found in the thesis Appendix. In this chapter, we will highlight a number of specific results in Tables \ref{tab:gdp_seoul_daegu}, \ref{tab:gdp_taichung_taipei} and \ref{tab:covid_daegu_daegu}. We will first directly compare the performance of all different methods, particularly which methods perform best under which conditions. Next, we will interpret the results on the transferability potential of different methods.

Our experimental results have generally been favourable to our proposed methods. In $23$ out of $40$ conditions SD-MRP outperformed (ties allowed) all baselines. This number was $15$ for WP-MRP and $35$ for O-MRP. In total, MRPs accounted for $80$ out of $136$ top-3 performances.
In a number of conditions, such as GDP interpolation for Daegu (same-city training), SD-MRP or WP-MRP achieved lower mean errors than a higher-ranked baseline. We 
believe this is due to either the higher standard deviation, or the lack of a normal distribution, rendering the lower mean error not statistically significant. However, this does imply that training an MRP multiple times and selecting the best performing model may give better results than the baselines.




In general, in conditions where the baselines outperformed MRP methods, the MRP methods typically still achieved fairly good results. In fact, apart from $3$ conditions for WP-MRP and $1$ condition for SD-MRP, MRP methods were never the worst performing method. This was a positive outcome for MRPs compared to the baselines, which were not consistently competitive. 
The two kriging methods (with negligible performance differences between them) were the worst-performing methods most often, at $15$ out of $40$ conditions (particularly in the COVID-19 dataset), but were also the best in $5$ conditions and among the top 3 in $19$ conditions. When tested on Taipei and trained on Taichung or Taipei itself, kriging performed particularly well, at times surpassing even O-MRP.
We speculate that the relatively good performance of kriging for Taichung-trained conditions may partially be caused by the non-representativeness of Taichung as a training region, as -- unlike the other cities which grew naturally over time -- Taichung was developed as a highly planned city. Since kriging was applied directly to the test region, with only the type of variogram model being tuned by the training set, it may not have suffered from this effect.

\begin{table}[H]
\centering
\resizebox{0.92\textwidth}{!}{%
\begin{tabular}{|l|llllllllll|}
\hline
\multicolumn{1}{|c|}{}                                  & \multicolumn{10}{c|}{}                                                                                                                                                                                                                                                                                                                                                                                              \\
\multicolumn{1}{|c|}{\multirow{-2}{*}{}}                & \multicolumn{10}{c|}{\multirow{-2}{*}{\textbf{Test data:   Daegu GDP}}}                                                                                                                                                                                                                                                                                                                                             \\ \hline
\multicolumn{1}{|c|}{}                                  & \multicolumn{10}{c|}{\textbf{Training region: Seoul}}                                                                                                                                                                                                                                                                                                                                                               \\ \cline{2-11} 
\multicolumn{1}{|c|}{}                                  & \multicolumn{2}{c|}{p=0.1}                                                      & \multicolumn{2}{c|}{p=0.3}                                                     & \multicolumn{2}{c|}{p=0.5}                                                     & \multicolumn{2}{c|}{p=0.7}                                                     & \multicolumn{2}{c|}{p=0.9}                                                     \\ \cline{2-11} 
\multicolumn{1}{|c|}{\multirow{-3}{*}{\textbf{Method}}} & \multicolumn{1}{c}{$\mu$}                       & \multicolumn{1}{c}{$\sigma$}  & \multicolumn{1}{c}{$\mu$}                      & \multicolumn{1}{c}{$\sigma$}  & \multicolumn{1}{c}{$\mu$}                      & \multicolumn{1}{c}{$\sigma$}  & \multicolumn{1}{c}{$\mu$}                      & \multicolumn{1}{c}{$\sigma$}  & \multicolumn{1}{c}{$\mu$}                      & \multicolumn{1}{c|}{$\sigma$} \\ \hline
OK                                                      & \cellcolor[HTML]{F8CECC}10.129$^{(\otimes)}$    & \cellcolor[HTML]{F8CECC}0.767 & \cellcolor[HTML]{F8CECC}10.03$^{(\otimes)}$    & \cellcolor[HTML]{F8CECC}0.664 & \cellcolor[HTML]{F8CECC}10.27$^{(\otimes)}$    & \cellcolor[HTML]{F8CECC}0.68  & \cellcolor[HTML]{F8CECC}9.808$^{(\otimes)}$    & \cellcolor[HTML]{F8CECC}0.595 & \cellcolor[HTML]{F8CECC}10.19$^{(\otimes)}$    & \cellcolor[HTML]{F8CECC}0.749 \\
UK                                                      & \cellcolor[HTML]{F8CECC}10.166$^{(\otimes)}$    & \cellcolor[HTML]{F8CECC}0.615 & \cellcolor[HTML]{F8CECC}10.3$^{(\otimes)}$     & \cellcolor[HTML]{F8CECC}0.618 & \cellcolor[HTML]{F8CECC}9.971$^{(\otimes)}$    & \cellcolor[HTML]{F8CECC}0.647 & \cellcolor[HTML]{F8CECC}10.05$^{(\otimes)}$    & \cellcolor[HTML]{F8CECC}0.701 & \cellcolor[HTML]{F8CECC}10.21$^{(\otimes)}$    & \cellcolor[HTML]{F8CECC}0.947 \\
Basic                                                   & \cellcolor[HTML]{FFFFFF}3.7723                  & \cellcolor[HTML]{FFFFFF}1.457 & \cellcolor[HTML]{FFFFFF}3.757                  & \cellcolor[HTML]{FFFFFF}0.553 & \cellcolor[HTML]{FFFFFF}3.833                  & \cellcolor[HTML]{FFFFFF}0.413 & \cellcolor[HTML]{FFFFFF}3.842                  & \cellcolor[HTML]{FFFFFF}0.247 & \cellcolor[HTML]{FFFFFF}3.867                  & \cellcolor[HTML]{FFFFFF}0.111 \\
SAR                                                     & \cellcolor[HTML]{FFFFFF}4.7495                  & \cellcolor[HTML]{FFFFFF}1.606 & \cellcolor[HTML]{FFFFFF}3.805                  & \cellcolor[HTML]{FFFFFF}1.068 & \cellcolor[HTML]{FFFFFF}4.203                  & \cellcolor[HTML]{FFFFFF}1.422 & \cellcolor[HTML]{FFFFFF}4.812                  & \cellcolor[HTML]{FFFFFF}1.577 & \cellcolor[HTML]{FFFFFF}4.331                  & \cellcolor[HTML]{FFFFFF}1.369 \\
MA                                                      & \cellcolor[HTML]{FFFFFF}5.1548                  & \cellcolor[HTML]{FFFFFF}2.078 & \cellcolor[HTML]{FFFFFF}6.567                  & \cellcolor[HTML]{FFFFFF}4.967 & \cellcolor[HTML]{FFFFFF}6.085                  & \cellcolor[HTML]{FFFFFF}2.404 & \cellcolor[HTML]{FFFFFF}4.847                  & \cellcolor[HTML]{FFFFFF}1.836 & \cellcolor[HTML]{FFFFFF}5.334                  & \cellcolor[HTML]{FFFFFF}3.829 \\
ARMA                                                    & \cellcolor[HTML]{FFFFFF}4.5039                  & \cellcolor[HTML]{FFFFFF}1.181 & \cellcolor[HTML]{FFFFFF}4.675                  & \cellcolor[HTML]{FFFFFF}1.319 & \cellcolor[HTML]{FFFFFF}5.062                  & \cellcolor[HTML]{FFFFFF}1.326 & \cellcolor[HTML]{FFFFFF}5.412                  & \cellcolor[HTML]{FFFFFF}1.763 & \cellcolor[HTML]{FFFFFF}5.164                  & \cellcolor[HTML]{FFFFFF}1.559 \\
CNN                                                     & \cellcolor[HTML]{FFFFFF}3.0891                  & \cellcolor[HTML]{FFFFFF}0.165 & \cellcolor[HTML]{D5E8D5}3.091$^{(3)}$          & \cellcolor[HTML]{D5E8D5}0.165 & \cellcolor[HTML]{D5E8D5}3.089$^{(3)}$          & \cellcolor[HTML]{D5E8D5}0.165 & \cellcolor[HTML]{FFFFFF}3.089                  & \cellcolor[HTML]{FFFFFF}0.165 & \cellcolor[HTML]{D5E8D5}3.086$^{(3)}$          & \cellcolor[HTML]{D5E8D5}0.164 \\ \hline
SD-MRP                                                  & \cellcolor[HTML]{D5E8D5}2.1037$^{(2)}$          & \cellcolor[HTML]{D5E8D5}1.202 & \cellcolor[HTML]{D5E8D5}2.177$^{(2)}$          & \cellcolor[HTML]{D5E8D5}0.577 & \cellcolor[HTML]{D5E8D5}2.085$^{(2)}$          & \cellcolor[HTML]{D5E8D5}0.347 & \cellcolor[HTML]{D5E8D5}2.616${(2)}$           & \cellcolor[HTML]{D5E8D5}0.307 & \cellcolor[HTML]{D5E8D5}2.911$^{(2)}$          & \cellcolor[HTML]{D5E8D5}0.338 \\
WP-MRP                                                  & \cellcolor[HTML]{D5E8D5}2.7458$^{(3)}$          & \cellcolor[HTML]{D5E8D5}1.534 & \cellcolor[HTML]{D5E8D5}3.02$^{(3)}$           & \cellcolor[HTML]{D5E8D5}0.855 & \cellcolor[HTML]{D5E8D5}3.162$^{(3)}$          & \cellcolor[HTML]{D5E8D5}0.651 & \cellcolor[HTML]{D5E8D5}2.901$^{(3)}$          & \cellcolor[HTML]{D5E8D5}0.376 & \cellcolor[HTML]{FFFFFF}3.017                  & \cellcolor[HTML]{FFFFFF}0.228 \\ \hline
O-MRP                                                   & \cellcolor[HTML]{D5E8D5}\textbf{0.5288$^{(1)}$} & \cellcolor[HTML]{D5E8D5}0.327 & \cellcolor[HTML]{D5E8D5}\textbf{0.982$^{(1)}$} & \cellcolor[HTML]{D5E8D5}0.227 & \cellcolor[HTML]{D5E8D5}\textbf{1.444$^{(1)}$} & \cellcolor[HTML]{D5E8D5}0.252 & \cellcolor[HTML]{D5E8D5}\textbf{2.055$^{(1)}$} & \cellcolor[HTML]{D5E8D5}0.296 & \cellcolor[HTML]{D5E8D5}\textbf{2.346$^{(1)}$} & \cellcolor[HTML]{D5E8D5}0.352 \\ \hline
\end{tabular}%
}
\caption{Results per setting of $p$ per method for GDP interpolation, trained on Seoul and tested on Daegu. Top-3 methods were marked using $\mu^{(1)}$, $\mu^{(2)}$ and $\mu^{(3)}$ (green), and the worst performing method was marked with $\mu^{(\otimes)}$ (red), determined using statistical significance testing at $\alpha=0.05$. The lowest error per condition, regardless of statistical significance, was marked \textbf{bold}.}
\label{tab:gdp_seoul_daegu}
\end{table}

\begin{table}[H]
\centering
\resizebox{0.92\textwidth}{!}{%
\begin{tabular}{|l|llllllllll|}
\hline
\multicolumn{1}{|c|}{}                                  & \multicolumn{10}{c|}{}                                                                                                                                                                                                                                                                                                                                                                                                    \\
\multicolumn{1}{|c|}{\multirow{-2}{*}{}}                & \multicolumn{10}{c|}{\multirow{-2}{*}{\textbf{Test data:   Taipei GDP}}}                                                                                                                                                                                                                                                                                                                                                  \\ \hline
\multicolumn{1}{|c|}{}                                  & \multicolumn{10}{c|}{\textbf{Training region: Taichung}}                                                                                                                                                                                                                                                                                                                                                                  \\ \cline{2-11} 
\multicolumn{1}{|c|}{}                                  & \multicolumn{2}{c|}{p=0.1}                                                     & \multicolumn{2}{c|}{p=0.3}                                                         & \multicolumn{2}{c|}{p=0.5}                                                     & \multicolumn{2}{c|}{p=0.7}                                                       & \multicolumn{2}{c|}{p=0.9}                                                      \\ \cline{2-11} 
\multicolumn{1}{|c|}{\multirow{-3}{*}{\textbf{Method}}} & \multicolumn{1}{c}{$\mu$}                      & \multicolumn{1}{c}{$\sigma$}  & \multicolumn{1}{c}{$\mu$}                          & \multicolumn{1}{c}{$\sigma$}  & \multicolumn{1}{c}{$\mu$}                      & \multicolumn{1}{c}{$\sigma$}  & \multicolumn{1}{c}{$\mu$}                        & \multicolumn{1}{c}{$\sigma$}  & \multicolumn{1}{c}{$\mu$}                       & \multicolumn{1}{c|}{$\sigma$} \\ \hline
OK                                                      & \cellcolor[HTML]{D5E8D5}2.6273$^{(2,3)}$       & \cellcolor[HTML]{D5E8D5}0.276 & \cellcolor[HTML]{D5E8D5}2.585$^{(1,2,3)}$          & \cellcolor[HTML]{D5E8D5}0.269 & \cellcolor[HTML]{D5E8D5}\textbf{2.521$^{(1)}$} & \cellcolor[HTML]{D5E8D5}0.233 & \cellcolor[HTML]{D5E8D5}\textbf{2.702$^{(1,2)}$} & \cellcolor[HTML]{D5E8D5}0.234 & \cellcolor[HTML]{D5E8D5}2.684$^{(1,2)}$         & \cellcolor[HTML]{D5E8D5}0.252 \\
UK                                                      & \cellcolor[HTML]{D5E8D5}2.6603$^{(2,3)}$       & \cellcolor[HTML]{D5E8D5}0.24  & \cellcolor[HTML]{D5E8D5}2.628$^{(1,2,3)}$          & \cellcolor[HTML]{D5E8D5}0.267 & \cellcolor[HTML]{D5E8D5}2.698$^{(2)}$          & \cellcolor[HTML]{D5E8D5}0.382 & \cellcolor[HTML]{D5E8D5}2.707$^{(1,2})$          & \cellcolor[HTML]{D5E8D5}0.284 & \cellcolor[HTML]{D5E8D5}\textbf{2.62$^{(1,2)}$} & \cellcolor[HTML]{D5E8D5}0.223 \\
Basic                                                   & \cellcolor[HTML]{FFFFFF}5.1524                 & \cellcolor[HTML]{FFFFFF}1.705 & \cellcolor[HTML]{FFFFFF}5.173                      & \cellcolor[HTML]{FFFFFF}0.895 & \cellcolor[HTML]{FFFFFF}5.245                  & \cellcolor[HTML]{FFFFFF}0.753 & \cellcolor[HTML]{FFFFFF}5.148                    & \cellcolor[HTML]{FFFFFF}0.381 & \cellcolor[HTML]{FFFFFF}5.08                    & \cellcolor[HTML]{FFFFFF}0.24  \\
SAR                                                     & \cellcolor[HTML]{FFFFFF}6.2848                 & \cellcolor[HTML]{FFFFFF}2.396 & \cellcolor[HTML]{FFFFFF}6.917                      & \cellcolor[HTML]{FFFFFF}4.324 & \cellcolor[HTML]{F8CECC}9.434$^{(\otimes})$    & \cellcolor[HTML]{F8CECC}13.95 & \cellcolor[HTML]{F8CECC}6.023$^{(\otimes)}$      & \cellcolor[HTML]{F8CECC}0.381 & \cellcolor[HTML]{F8CECC}6.065$^{(\otimes})$     & \cellcolor[HTML]{F8CECC}0.852 \\
MA                                                      & \cellcolor[HTML]{F8CECC}7.5805$^{(\otimes)}$   & \cellcolor[HTML]{F8CECC}5.457 & \cellcolor[HTML]{F8CECC}9.742$^{(\otimes)}$        & \cellcolor[HTML]{F8CECC}11.33 & \cellcolor[HTML]{F8CECC}7.996$^({\otimes})$    & \cellcolor[HTML]{F8CECC}4.825 & \cellcolor[HTML]{F8CECC}9.533$^{(\otimes)}$      & \cellcolor[HTML]{F8CECC}15.58 & \cellcolor[HTML]{FFFFFF}13.05                   & \cellcolor[HTML]{FFFFFF}17.24 \\
ARMA                                                    & \cellcolor[HTML]{FFFFFF}5.7533                 & \cellcolor[HTML]{FFFFFF}0.884 & \cellcolor[HTML]{FFFFFF}5.822                      & \cellcolor[HTML]{FFFFFF}0.898 & \cellcolor[HTML]{FFFFFF}5.852                  & \cellcolor[HTML]{FFFFFF}1.067 & \cellcolor[HTML]{F8CECC}6.041$^{(\otimes})$      & \cellcolor[HTML]{F8CECC}0.843 & \cellcolor[HTML]{FFFFFF}6.008                   & \cellcolor[HTML]{FFFFFF}1.072 \\
CNN                                                     & \cellcolor[HTML]{FFFFFF}6.1006                 & \cellcolor[HTML]{FFFFFF}2.251 & \cellcolor[HTML]{FFFFFF}5.914                      & \cellcolor[HTML]{FFFFFF}1.26  & \cellcolor[HTML]{FFFFFF}5.824                  & \cellcolor[HTML]{FFFFFF}1.454 & \cellcolor[HTML]{FFFFFF}5.683                    & \cellcolor[HTML]{FFFFFF}1.359 & \cellcolor[HTML]{FFFFFF}6.137                   & \cellcolor[HTML]{FFFFFF}2.114 \\ \hline
SD-MRP                                                  & \cellcolor[HTML]{FFFFFF}3.9926                 & \cellcolor[HTML]{FFFFFF}1.451 & \cellcolor[HTML]{FFFFFF}4.441                      & \cellcolor[HTML]{FFFFFF}0.784 & \cellcolor[HTML]{FFFFFF}4.375                  & \cellcolor[HTML]{FFFFFF}0.409 & \cellcolor[HTML]{FFFFFF}4.516                    & \cellcolor[HTML]{FFFFFF}0.471 & \cellcolor[HTML]{D5E8D5}4.913$^{(3)}$           & \cellcolor[HTML]{D5E8D5}0.442 \\
WP-MRP                                                  & \cellcolor[HTML]{FFFFFF}5.4675                 & \cellcolor[HTML]{FFFFFF}2.538 & \cellcolor[HTML]{FFFFFF}5.934                      & \cellcolor[HTML]{FFFFFF}1.924 & \cellcolor[HTML]{FFFFFF}5.284                  & \cellcolor[HTML]{FFFFFF}0.962 & \cellcolor[HTML]{FFFFFF}5.343                    & \cellcolor[HTML]{FFFFFF}0.718 & \cellcolor[HTML]{FFFFFF}5.183                   & \cellcolor[HTML]{FFFFFF}0.319 \\ \hline
O-MRP                                                   & \cellcolor[HTML]{D5E8D5}\textbf{1.356$^{(1)}$} & \cellcolor[HTML]{D5E8D5}0.49  & \cellcolor[HTML]{D5E8D5}\textbf{2.472$^{(1,2,3)}$} & \cellcolor[HTML]{D5E8D5}0.573 & \cellcolor[HTML]{D5E8D5}3.513$^{(3)}$          & \cellcolor[HTML]{D5E8D5}0.531 & \cellcolor[HTML]{D5E8D5}4.264$^{(3)}$            & \cellcolor[HTML]{D5E8D5}0.477 & \cellcolor[HTML]{D5E8D5}4.909$^{(3)}$           & \cellcolor[HTML]{D5E8D5}0.485 \\ \hline
\end{tabular}%
}
\caption{Results per setting of $p$ per method for GDP interpolation, trained on Taichung and tested on Taipei, using the same notation as in Table \ref{tab:gdp_seoul_daegu}.} 
\label{tab:gdp_taichung_taipei}
\end{table}

\begin{table}[H]
\centering
\resizebox{0.92\textwidth}{!}{%
\begin{tabular}{|l|llllllllll|}
\hline
\multicolumn{1}{|c|}{}                                  & \multicolumn{10}{c|}{}                                                                                                                                                                                                                                                                                                                                                                                              \\
\multicolumn{1}{|c|}{\multirow{-2}{*}{}}                & \multicolumn{10}{c|}{\multirow{-2}{*}{\textbf{Test data:   Daegu COVID-19}}}                                                                                                                                                                                                                                                                                                                                        \\ \hline
\multicolumn{1}{|c|}{}                                  & \multicolumn{10}{c|}{\textbf{Training region: Daegu}}                                                                                                                                                                                                                                                                                                                                                               \\ \cline{2-11} 
\multicolumn{1}{|c|}{}                                  & \multicolumn{2}{c|}{p=0.1}                                                      & \multicolumn{2}{c|}{p=0.3}                                                     & \multicolumn{2}{c|}{p=0.5}                                                     & \multicolumn{2}{c|}{p=0.7}                                                     & \multicolumn{2}{c|}{p=0.9}                                                     \\ \cline{2-11} 
\multicolumn{1}{|c|}{\multirow{-3}{*}{\textbf{Method}}} & \multicolumn{1}{c}{$\mu$}                       & \multicolumn{1}{c}{$\sigma$}  & \multicolumn{1}{c}{$\mu$}                      & \multicolumn{1}{c}{$\sigma$}  & \multicolumn{1}{c}{$\mu$}                      & \multicolumn{1}{c}{$\sigma$}  & \multicolumn{1}{c}{$\mu$}                      & \multicolumn{1}{c}{$\sigma$}  & \multicolumn{1}{c}{$\mu$}                      & \multicolumn{1}{c|}{$\sigma$} \\ \hline
OK                                                      & \cellcolor[HTML]{F8CECC}12.365$^{(\otimes)}$    & \cellcolor[HTML]{F8CECC}1.342 & \cellcolor[HTML]{F8CECC}13.64$^{(\otimes)}$    & \cellcolor[HTML]{F8CECC}1.475 & \cellcolor[HTML]{F8CECC}13.04$^{(\otimes)}$    & \cellcolor[HTML]{F8CECC}1.513 & \cellcolor[HTML]{F8CECC}13.17$^{(\otimes)}$    & \cellcolor[HTML]{F8CECC}1.341 & \cellcolor[HTML]{F8CECC}13.4$^{(\otimes)}$     & \cellcolor[HTML]{F8CECC}1.202 \\
UK                                                      & \cellcolor[HTML]{F8CECC}12.546$^{(\otimes)}$    & \cellcolor[HTML]{F8CECC}1.511 & \cellcolor[HTML]{F8CECC}13.09$^{(\otimes)}$    & \cellcolor[HTML]{F8CECC}1.415 & \cellcolor[HTML]{F8CECC}13.18$^{(\otimes)}$    & \cellcolor[HTML]{F8CECC}1.313 & \cellcolor[HTML]{F8CECC}13.13$^{(\otimes)}$    & \cellcolor[HTML]{F8CECC}1.55  & \cellcolor[HTML]{F8CECC}13.48$^{(\otimes)}$    & \cellcolor[HTML]{F8CECC}1.444 \\
Basic                                                   & \cellcolor[HTML]{FFFFFF}2.5102                  & \cellcolor[HTML]{FFFFFF}0.234 & \cellcolor[HTML]{FFFFFF}2.56                   & \cellcolor[HTML]{FFFFFF}0.461 & \cellcolor[HTML]{FFFFFF}2.435                  & \cellcolor[HTML]{FFFFFF}0.606 & \cellcolor[HTML]{FFFFFF}2.492                  & \cellcolor[HTML]{FFFFFF}1.006 & \cellcolor[HTML]{FFFFFF}2.279                  & \cellcolor[HTML]{FFFFFF}1.625 \\
SAR                                                     & \cellcolor[HTML]{D5E8D5}1.1801$^{(2)}$          & \cellcolor[HTML]{D5E8D5}0.836 & \cellcolor[HTML]{D5E8D5}1.638$^{(2,3)}$        & \cellcolor[HTML]{D5E8D5}0.679 & \cellcolor[HTML]{D5E8D5}1.63$^{(3)}$           & \cellcolor[HTML]{D5E8D5}0.629 & \cellcolor[HTML]{FFFFFF}2.437                  & \cellcolor[HTML]{FFFFFF}2.029 & \cellcolor[HTML]{FFFFFF}2.402                  & \cellcolor[HTML]{FFFFFF}1.632 \\
MA                                                      & \cellcolor[HTML]{FFFFFF}1.8745                  & \cellcolor[HTML]{FFFFFF}0.395 & \cellcolor[HTML]{FFFFFF}2.239                  & \cellcolor[HTML]{FFFFFF}1.085 & \cellcolor[HTML]{FFFFFF}3.912                  & \cellcolor[HTML]{FFFFFF}7.824 & \cellcolor[HTML]{FFFFFF}3.807                  & \cellcolor[HTML]{FFFFFF}3.683 & \cellcolor[HTML]{F8CECC}26.69$^{(\otimes)}$    & \cellcolor[HTML]{F8CECC}83    \\
ARMA                                                    & \cellcolor[HTML]{FFFFFF}2.3224                  & \cellcolor[HTML]{FFFFFF}1.26  & \cellcolor[HTML]{FFFFFF}2.355                  & \cellcolor[HTML]{FFFFFF}0.901 & \cellcolor[HTML]{FFFFFF}2.31                   & \cellcolor[HTML]{FFFFFF}0.675 & \cellcolor[HTML]{FFFFFF}2.682                  & \cellcolor[HTML]{FFFFFF}1.478 & \cellcolor[HTML]{FFFFFF}3.453                  & \cellcolor[HTML]{FFFFFF}2.227 \\
CNN                                                     & \cellcolor[HTML]{D5E8D5}1.6711$^{(3)}$          & \cellcolor[HTML]{D5E8D5}0.247 & \cellcolor[HTML]{D5E8D5}1.794$^{(2,3)}$        & \cellcolor[HTML]{D5E8D5}0.376 & \cellcolor[HTML]{FFFFFF}1.9                    & \cellcolor[HTML]{FFFFFF}0.662 & \cellcolor[HTML]{FFFFFF}1.879                  & \cellcolor[HTML]{FFFFFF}1.083 & \cellcolor[HTML]{D5E8D5}1.819$^{(2)}$          & \cellcolor[HTML]{D5E8D5}1.97  \\ \hline
SD-MRP                                                  & \cellcolor[HTML]{FFFFFF}2.1936                  & \cellcolor[HTML]{FFFFFF}2.399 & \cellcolor[HTML]{FFFFFF}1.925                  & \cellcolor[HTML]{FFFFFF}1.014 & \cellcolor[HTML]{D5E8D5}1.61$^{(2)}$           & \cellcolor[HTML]{D5E8D5}0.699 & \cellcolor[HTML]{D5E8D5}1.708$^{(3)}$          & \cellcolor[HTML]{D5E8D5}0.397 & \cellcolor[HTML]{D5E8D5}1.652$^{(3)}$          & \cellcolor[HTML]{D5E8D5}0.248 \\
WP-MRP                                                  & \cellcolor[HTML]{FFFFFF}2.0035                  & \cellcolor[HTML]{FFFFFF}2.427 & \cellcolor[HTML]{D5E8D5}1.948$^{(2,3)}$        & \cellcolor[HTML]{D5E8D5}2.024 & \cellcolor[HTML]{D5E8D5}1.632$^{(3)}$          & \cellcolor[HTML]{D5E8D5}0.679 & \cellcolor[HTML]{D5E8D5}1.585$^{(2)}$          & \cellcolor[HTML]{D5E8D5}0.452 & \cellcolor[HTML]{FFFFFF}1.67                   & \cellcolor[HTML]{FFFFFF}0.245 \\ \hline
O-MRP                                                   & \cellcolor[HTML]{D5E8D5}\textbf{0.2647$^{(1)}$} & \cellcolor[HTML]{D5E8D5}0.311 & \cellcolor[HTML]{D5E8D5}\textbf{0.611$^{(1)}$} & \cellcolor[HTML]{D5E8D5}0.311 & \cellcolor[HTML]{D5E8D5}\textbf{0.757$^{(1)}$} & \cellcolor[HTML]{D5E8D5}0.366 & \cellcolor[HTML]{D5E8D5}\textbf{1.008$^{(1)}$} & \cellcolor[HTML]{D5E8D5}0.313 & \cellcolor[HTML]{D5E8D5}\textbf{1.197$^{(1)}$} & \cellcolor[HTML]{D5E8D5}0.253 \\ \hline
\end{tabular}%
}
\caption{Results per setting of $p$ per method for COVID-19 interpolation, trained on Daegu and tested on Daegu, using the same notation as in Table \ref{tab:gdp_seoul_daegu}.} 
\label{tab:covid_daegu_daegu}
\end{table}

Although WP-MRP did not compare as well to the baselines as SD-MRP, its performance was generally comparable to, if not slightly better than CNN's performance, despite being a much lighter model. 
It does appear, however, that WP-MRP rarely suffers from invalid results (outliers on an order between $10^{10}$ and even $10^{100}$) in its results; closer inspection showed that in these cases, the optimisation algorithm failed to converge for training on Taichung, and we ran our analyses excluding these runs. 
Given the results, though it outperformed the best-performing baseline in $15$ out of $40$ runs, we believe that further work is needed to realise the full potential of WP-MRP.

\begin{figure}[t]
  \centering
  \includegraphics[width=0.9\linewidth]{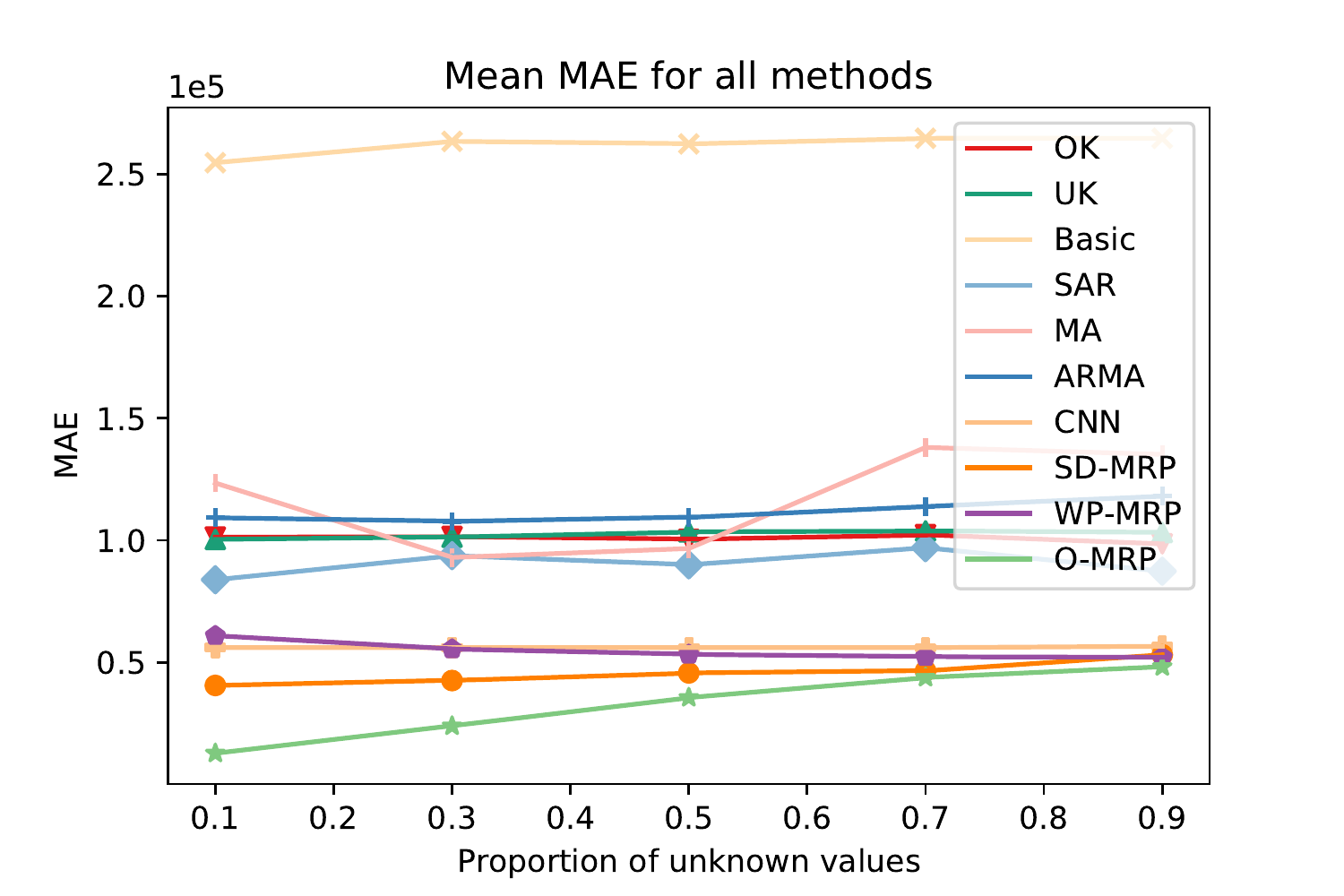}
  \caption{Performance of all methods as a function of $p$ for GDP interpolation trained on Seoul and tested on Daegu. }
  \label{fig:lines_all}
\end{figure}

\begin{figure}[t]
  \centering
  \includegraphics[width=0.9\linewidth]{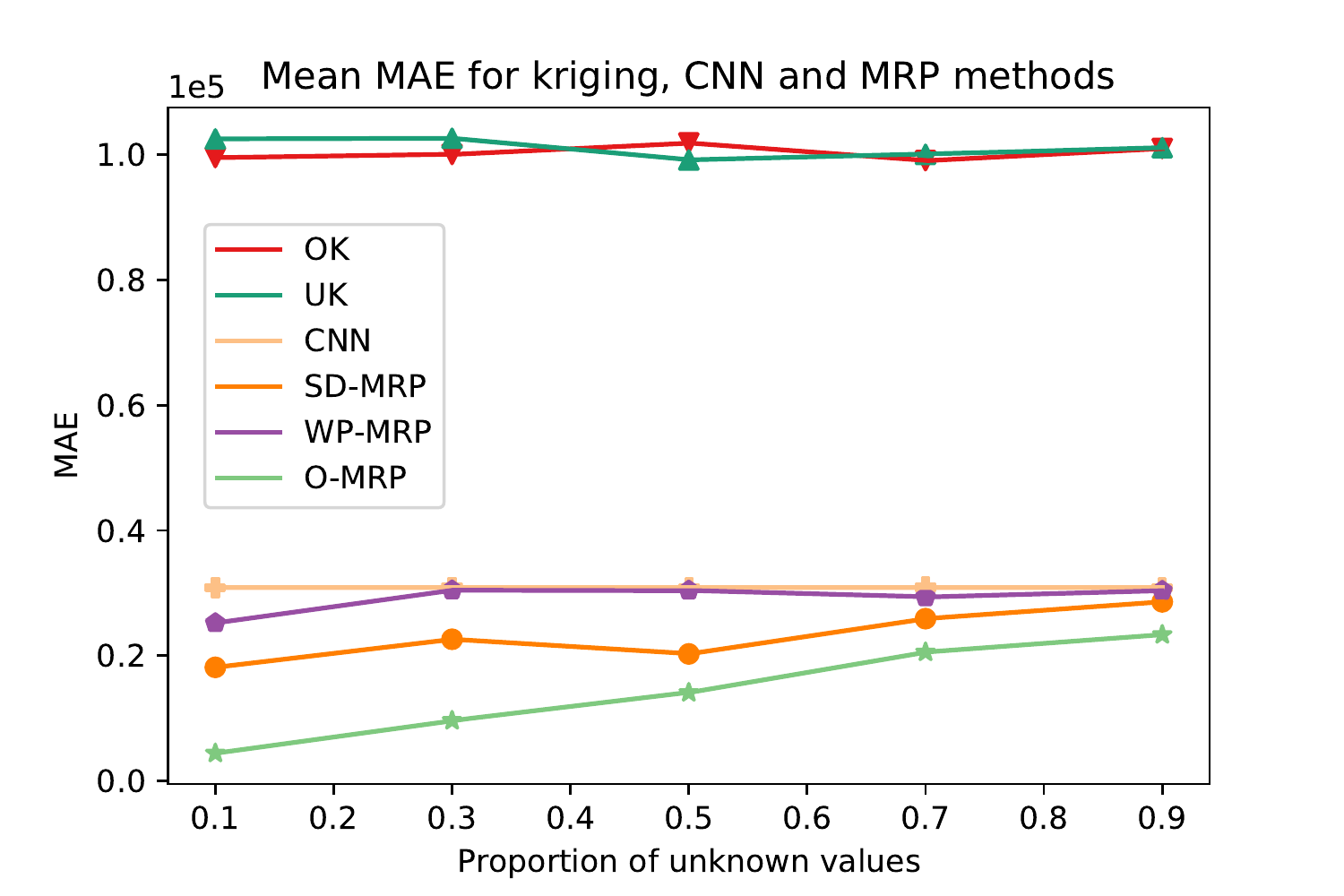}
  \caption{Performance MRP methods compared to the most competitive baselines (kriging and CNN) for GDP interpolation trained on Seoul and tested on Daegu. }
  \label{fig:lines_comp}
\end{figure}

When it comes to transferability, 
only a very mild reduction of accuracy can be observed for more distant training regions for WP-MRP. However, MRP interpolation generally appears not to be 
heavily affected by the distance between training and testing regions, which we consider a positive result. MRPs do appear to be more sensitive to the setting of $p$ than most baselines, as can be seen in Figures \ref{fig:lines_all} and \ref{fig:lines_comp}.


MRP methods were more prominently effective for the COVID-19 dataset than the GDP dataset, implying there may be some variance in performance based on the dataset. We speculate that the irregularity of the COVID-19 dataset, containing localised clusters that do not work well with distance-based methods such as kriging, was the cause for this. 
Apart from this, the performance rankings between the GDP and COVID-19 datasets were quite similar. This is a good outcome, not only because it is an encouraging sign that our results may generalise to other datasets as well, but also because the effective interpolation of virus exposure risk might be a very useful tool in virus containment efforts.

\chapter{Conclusion and future work}
\label{chap:conclusion}
In this thesis, 
we have explored the use of Markov reward processes (MRPs) to solve spatial interpolation problems. These MRPs were primarily aimed at addressing the limitation of existing methods, which suffer from having to trade-off between local and global spatial effects. We introduced three different variants of MRP interpolation (SD-MRP, O-MRP and WP-MRP), offering different combinations of practical applicability and robustness to non-stationarity and anisotropy.
We tested these methods on a GDP dataset for two cities in South Korea and Taiwan, with same-city, same-country and different-country transferability conditions. In our experiments, SD-MRP outperformed all baselines in $23$ out of $40$ conditions, and O-MRP did in $35$ out of $40$ conditions. The WP-MRP was unable to predict optimised weights with sufficient accuracy to outperform SD-MRP, but still remained competitive in most cases and outperformed all baselines in $15$ conditions.

In future work, 
it would be interesting to further study WP-MRP; perhaps different models or different explanatory variables could be more effective at predicting optimal weights. Of course, other variants of MRP interpolation, beyond the three proposed in this thesis, also offer a promising avenue for future work.

Overall, our results have clearly shown the potential of MRP-based spatial interpolation, and we thus hope to see it studied further and to see it broadly applied in practice. 

\chapter*{Ethical statement}
\label{chap:ethical_statement}
The COVID-19 dataset provided by DACON is quite rare, as it contains individual trajectories of infected COVID patients prior to their diagnosis. Although such a dataset could clearly be of great use to the South Korean pandemic response, as well as that of other countries looking to learn from this data, there are obvious privacy concerns to raise for this dataset. We believe our use of this dataset to have been justified by the strict standards of responsible AI one should expect in the AI research community, particularly as we transformed the dataset from individual trajectories (which could lead to personally identifiable information) to location-based aggregated counts of visits without considering any temporal aspects. While this format does not completely eliminate individual privacy concerns, we believe it is sufficiently addressed in this manner for the use of this dataset to be admissible in a thesis. One should also be wary of the use of spatial interpolation on this dataset, as well as others, for the purposes of local and regional discrimination. While data-driven decision making certainly has its merits, one should not forget that models are only models making predictions, and real-world policy being made on the basis of these predictions, particularly policy negatively affecting individuals or groups of individuals, should always be viewed with great caution and restraint.\\
Finally, while a discussion could (and should) certainly be had 
on the merits of using data that is available anyway  
for the improvement of the pandemic response around the world, weighed against the wariness against the use of data that one may be principally opposed to being collected in the first place, one should not forget the debt of gratitude owed to the South Koreans providing their data if it is indeed used for the improvement of the global pandemic response.


\bibliographystyle{unsrt}
\bibliography{sources.bib}

\chapter{Appendix}
\label{chap:appendix}
This appendix contains full results that would have taken up too much space to include in the main thesis. The full results given here consist of all conditions, of which $3$ examples were given in the main thesis. As in the thesis, we performed significance tests to rank all methods per condition, and highlighted our results based on this.

\section{Tables}

Tables of our full results are given below in Tables \ref{tab:gdp_taipei_taipei}-\ref{tab:covid_seoul_daegu} starting on the next page. Overall, these extended results show the same patterns as those discussed in the thesis.

\begin{table}[]
\centering
\resizebox{\textwidth}{!}{%
\begin{tabular}{|l|llllllllll|}
\hline
\multicolumn{1}{|c|}{}                                  & \multicolumn{10}{c|}{}                                                                                                                                                                                                                                                                                                                                                                                             \\
\multicolumn{1}{|c|}{\multirow{-2}{*}{}}                & \multicolumn{10}{c|}{\multirow{-2}{*}{\textbf{Test data:   Taipei GDP}}}                                                                                                                                                                                                                                                                                                                                           \\ \hline
\multicolumn{1}{|c|}{}                                  & \multicolumn{10}{c|}{\textbf{Training region: Taipei}}                                                                                                                                                                                                                                                                                                                                                             \\ \cline{2-11} 
\multicolumn{1}{|c|}{}                                  & \multicolumn{2}{c|}{p=0.1}                                                      & \multicolumn{2}{c|}{p=0.3}                                                    & \multicolumn{2}{c|}{p=0.5}                                                     & \multicolumn{2}{c|}{p=0.7}                                                     & \multicolumn{2}{c|}{p=0.9}                                                     \\ \cline{2-11} 
\multicolumn{1}{|c|}{\multirow{-3}{*}{\textbf{Method}}} & \multicolumn{1}{c}{$\mu$}                       & \multicolumn{1}{c}{$\sigma$}  & \multicolumn{1}{c}{$\mu$}                     & \multicolumn{1}{c}{$\sigma$}  & \multicolumn{1}{c}{$\mu$}                      & \multicolumn{1}{c}{$\sigma$}  & \multicolumn{1}{c}{$\mu$}                      & \multicolumn{1}{c}{$\sigma$}  & \multicolumn{1}{c}{$\mu$}                      & \multicolumn{1}{c|}{$\sigma$} \\ \hline
OK                                                      & \cellcolor[HTML]{D5E8D5}2.6301$^{(2,3)}$        & \cellcolor[HTML]{D5E8D5}0.233 & \cellcolor[HTML]{D5E8D5}2.66$^{(2,3)}$        & \cellcolor[HTML]{D5E8D5}0.272 & \cellcolor[HTML]{D5E8D5}2.629$^{(2,3)}$        & \cellcolor[HTML]{D5E8D5}0.244 & \cellcolor[HTML]{D5E8D5}2.672$^{(2,3)}$        & \cellcolor[HTML]{D5E8D5}0.328 & \cellcolor[HTML]{D5E8D5}\textbf{2.615$^{(1)}$} & \cellcolor[HTML]{D5E8D5}0.232 \\
UK                                                      & \cellcolor[HTML]{D5E8D5}2.6704$^{(2,3)}$        & \cellcolor[HTML]{D5E8D5}0.236 & \cellcolor[HTML]{D5E8D5}2.698$^{(2,3)}$       & \cellcolor[HTML]{D5E8D5}0.267 & \cellcolor[HTML]{D5E8D5}2.593$^{(2,3)}$        & \cellcolor[HTML]{D5E8D5}0.223 & \cellcolor[HTML]{D5E8D5}2.726$^{(2,3)}$        & \cellcolor[HTML]{D5E8D5}0.278 & \cellcolor[HTML]{D5E8D5}2.727$^{(2)}$          & \cellcolor[HTML]{D5E8D5}0.249 \\
Basic                                                   & 3.9025                                          & 0.108                         & 3.898                                         & 0.213                         & 3.819                                          & 0.355                         & 3.801                                          & 0.521                         & 3.662                                          & 0.857                         \\
SAR                                                     & 3.5036                                          & 0.532                         & 4.041                                         & 0.935                         & 4.121                                          & 0.488                         & 4.494                                          & 0.642                         & \cellcolor[HTML]{F8CECC}6.041$^{(\otimes)}$    & \cellcolor[HTML]{F8CECC}1.659 \\
MA                                                      & 3.5539                                          & 0.556                         & 3.625                                         & 0.551                         & 3.691                                          & 0.505                         & 3.876                                          & 0.722                         & 4.671                                          & 1.827                         \\
ARMA                                                    & 3.8732                                          & 0.569                         & 3.973                                         & 0.669                         & 4.089                                          & 0.728                         & 4.252                                          & 1.002                         & 4.391                                          & 0.689                         \\
CNN                                                     & \cellcolor[HTML]{F8CECC}5.4808$^{(\otimes)}$    & \cellcolor[HTML]{F8CECC}0.378 & \cellcolor[HTML]{F8CECC}5.427$^{(\otimes)}$   & \cellcolor[HTML]{F8CECC}0.567 & \cellcolor[HTML]{F8CECC}5.461$^{(\otimes)}$    & \cellcolor[HTML]{F8CECC}0.811 & \cellcolor[HTML]{F8CECC}5.349$^{(\otimes)}$    & \cellcolor[HTML]{F8CECC}1.039 & 5.337                                          & 2.259                         \\ \hline
SD-MRP                                                  & 4.2816                                          & 1.981                         & 4.863                                         & 1.119                         & 4.625                                          & 0.736                         & \cellcolor[HTML]{F8CECC}4.962$^{(\otimes)}$    & \cellcolor[HTML]{F8CECC}0.623 & 5.188                                          & 0.329                         \\
WP-MRP                                                  & 4.9688                                          & 2.243                         & \cellcolor[HTML]{F8CECC}5.556$^{(\otimes)}$   & \cellcolor[HTML]{F8CECC}1.077 & \cellcolor[HTML]{F8CECC}5.382$^{(\otimes)}$    & \cellcolor[HTML]{F8CECC}0.723 & \cellcolor[HTML]{F8CECC}5.223$^{(\otimes)}$    & \cellcolor[HTML]{F8CECC}0.523 & 5.296                                          & 0.263                         \\ \hline
O-MRP                                                   & \cellcolor[HTML]{D5E8D5}\textbf{0.9534$^{(1)}$} & \cellcolor[HTML]{D5E8D5}0.514 & \cellcolor[HTML]{D5E8D5}\textbf{1.46$^{(1)}$} & \cellcolor[HTML]{D5E8D5}0.334 & \cellcolor[HTML]{D5E8D5}\textbf{1.753$^{(1)}$} & \cellcolor[HTML]{D5E8D5}0.343 & \cellcolor[HTML]{D5E8D5}\textbf{2.205$^{(1)}$} & \cellcolor[HTML]{D5E8D5}0.311 & \cellcolor[HTML]{D5E8D5}3.021$^{(3)}$          & \cellcolor[HTML]{D5E8D5}0.746 \\ \hline
\end{tabular}%
}
\caption{Results per setting of $p$ per method for GDP interpolation, trained on Taipei and tested on Taipei. Top-3 methods were marked using $\mu^{(1)}$, $\mu^{(2)}$ and $\mu^{(3)}$ (green), and the worst performing method was marked with $\mu^{(\otimes)}$ (red), determined using statistical significance testing at $\alpha=0.05$. The lowest error per condition, regardless of statistical significance, was marked \textbf{bold}.}
\label{tab:gdp_taipei_taipei}
\end{table}

\begin{table}[]
\centering
\resizebox{\textwidth}{!}{%
\begin{tabular}{|l|llllllllll|}
\hline
\multicolumn{1}{|c|}{}                                  & \multicolumn{10}{c|}{}                                                                                                                                                                                                                                                                                                                                                                               \\
\multicolumn{1}{|c|}{\multirow{-2}{*}{}}                & \multicolumn{10}{c|}{\multirow{-2}{*}{\textbf{Test data:   Taipei GDP}}}                                                                                                                                                                                                                                                                                                                             \\ \hline
\multicolumn{1}{|c|}{}                                  & \multicolumn{10}{c|}{\textbf{Training region: Seoul}}                                                                                                                                                                                                                                                                                                                                                \\ \cline{2-11} 
\multicolumn{1}{|c|}{}                                  & \multicolumn{2}{c|}{p=0.1}                                                   & \multicolumn{2}{c|}{p=0.3}                                                  & \multicolumn{2}{c|}{p=0.5}                                                  & \multicolumn{2}{c|}{p=0.7}                                                  & \multicolumn{2}{c|}{p=0.9}                                                  \\ \cline{2-11} 
\multicolumn{1}{|c|}{\multirow{-3}{*}{\textbf{Method}}} & \multicolumn{1}{c}{$\mu$}                    & \multicolumn{1}{c}{$\sigma$}  & \multicolumn{1}{c}{$\mu$}                   & \multicolumn{1}{c}{$\sigma$}  & \multicolumn{1}{c}{$\mu$}                   & \multicolumn{1}{c}{$\sigma$}  & \multicolumn{1}{c}{$\mu$}                   & \multicolumn{1}{c}{$\sigma$}  & \multicolumn{1}{c}{$\mu$}                   & \multicolumn{1}{c|}{$\sigma$} \\ \hline
OK                                                      & \cellcolor[HTML]{FFFFFF}10.131               & \cellcolor[HTML]{FFFFFF}0.589 & \cellcolor[HTML]{FFFFFF}10.15               & \cellcolor[HTML]{FFFFFF}0.451 & \cellcolor[HTML]{FFFFFF}10.04               & \cellcolor[HTML]{FFFFFF}0.583 & \cellcolor[HTML]{FFFFFF}10.23               & \cellcolor[HTML]{FFFFFF}0.548 & \cellcolor[HTML]{FFFFFF}9.853               & \cellcolor[HTML]{FFFFFF}0.465 \\
UK                                                      & \cellcolor[HTML]{FFFFFF}10.038               & \cellcolor[HTML]{FFFFFF}0.835 & \cellcolor[HTML]{FFFFFF}10.13               & \cellcolor[HTML]{FFFFFF}0.591 & \cellcolor[HTML]{FFFFFF}10.35               & \cellcolor[HTML]{FFFFFF}0.659 & \cellcolor[HTML]{FFFFFF}10.39               & \cellcolor[HTML]{FFFFFF}0.531 & \cellcolor[HTML]{FFFFFF}10.33               & \cellcolor[HTML]{FFFFFF}0.74  \\
Basic                                                   & \cellcolor[HTML]{F8CECC}25.474$^{(\otimes)}$ & \cellcolor[HTML]{F8CECC}4.757 & \cellcolor[HTML]{F8CECC}26.35$^{(\otimes)}$ & \cellcolor[HTML]{F8CECC}1.213 & \cellcolor[HTML]{F8CECC}26.24$^{(\otimes)}$ & \cellcolor[HTML]{F8CECC}0.704 & \cellcolor[HTML]{F8CECC}26.48$^{(\otimes)}$ & \cellcolor[HTML]{F8CECC}0.602 & \cellcolor[HTML]{F8CECC}26.47$^{(\otimes)}$ & \cellcolor[HTML]{F8CECC}0.306 \\
SAR                                                     & \cellcolor[HTML]{FFFFFF}8.3827               & \cellcolor[HTML]{FFFFFF}1.719 & \cellcolor[HTML]{FFFFFF}9.37                & \cellcolor[HTML]{FFFFFF}4.045 & \cellcolor[HTML]{FFFFFF}8.996               & \cellcolor[HTML]{FFFFFF}1.585 & \cellcolor[HTML]{FFFFFF}9.695               & \cellcolor[HTML]{FFFFFF}4.467 & \cellcolor[HTML]{FFFFFF}8.731               & \cellcolor[HTML]{FFFFFF}1.298 \\
MA                                                      & \cellcolor[HTML]{FFFFFF}12.348               & \cellcolor[HTML]{FFFFFF}8.498 & \cellcolor[HTML]{FFFFFF}9.295               & \cellcolor[HTML]{FFFFFF}3.965 & \cellcolor[HTML]{FFFFFF}9.662               & \cellcolor[HTML]{FFFFFF}5.201 & \cellcolor[HTML]{FFFFFF}13.81               & \cellcolor[HTML]{FFFFFF}14.06 & \cellcolor[HTML]{FFFFFF}13.52               & \cellcolor[HTML]{FFFFFF}12.25 \\
ARMA                                                    & \cellcolor[HTML]{FFFFFF}10.922               & \cellcolor[HTML]{FFFFFF}5.503 & \cellcolor[HTML]{FFFFFF}10.77               & \cellcolor[HTML]{FFFFFF}6.012 & \cellcolor[HTML]{FFFFFF}10.94               & \cellcolor[HTML]{FFFFFF}7.303 & \cellcolor[HTML]{FFFFFF}11.39               & \cellcolor[HTML]{FFFFFF}7.272 & \cellcolor[HTML]{FFFFFF}11.8                & \cellcolor[HTML]{FFFFFF}4.77  \\
CNN                                                     & \cellcolor[HTML]{D5E8D5}5.6052$^{(3)}$       & \cellcolor[HTML]{D5E8D5}0.242 & \cellcolor[HTML]{D5E8D5}5.601$^{(3)}$       & \cellcolor[HTML]{D5E8D5}0.243 & \cellcolor[HTML]{FFFFFF}5.6                 & \cellcolor[HTML]{FFFFFF}0.245 & \cellcolor[HTML]{FFFFFF}5.6                 & \cellcolor[HTML]{FFFFFF}0.245 & \cellcolor[HTML]{D5E8D5}5.655$^{(3)}$       & \cellcolor[HTML]{D5E8D5}0.238 \\ \hline
SD-MRP                                                  & \cellcolor[HTML]{D5E8D5}4.0505$^{(2)}$       & \cellcolor[HTML]{D5E8D5}1.348 & \cellcolor[HTML]{D5E8D5}4.251$^{(2)}$       & \cellcolor[HTML]{D5E8D5}0.725 & \cellcolor[HTML]{D5E8D5}4.564$^{(2)}$       & \cellcolor[HTML]{D5E8D5}0.508 & \cellcolor[HTML]{D5E8D5}4.646$^{(1,2)}$     & \cellcolor[HTML]{D5E8D5}0.606 & \cellcolor[HTML]{D5E8D5}5.308$^{(2)}$       & \cellcolor[HTML]{D5E8D5}0.557 \\
WP-MRP                                                  & \cellcolor[HTML]{D5E8D5}6.0888$^{(3)}$       & \cellcolor[HTML]{D5E8D5}2.464 & \cellcolor[HTML]{FFFFFF}5.542               & \cellcolor[HTML]{FFFFFF}1.057 & \cellcolor[HTML]{D5E8D5}5.322$^{(3)}$       & \cellcolor[HTML]{D5E8D5}0.781 & \cellcolor[HTML]{D5E8D5}5.226$^{(3)}$       & \cellcolor[HTML]{D5E8D5}0.532 & \cellcolor[HTML]{FFFFFF}5.205               & \cellcolor[HTML]{FFFFFF}0.297 \\ \hline
O-MRP                                                   & \cellcolor[HTML]{D5E8D5}1.2719$^{(1)}$       & \cellcolor[HTML]{D5E8D5}0.708 & \cellcolor[HTML]{D5E8D5}2.388$^{(1)}$       & \cellcolor[HTML]{D5E8D5}0.437 & \cellcolor[HTML]{D5E8D5}3.554$^{(1)}$       & \cellcolor[HTML]{D5E8D5}0.604 & \cellcolor[HTML]{D5E8D5}4.378$^{(1,2)}$     & \cellcolor[HTML]{D5E8D5}0.557 & \cellcolor[HTML]{D5E8D5}4.818$^{(1)}$       & \cellcolor[HTML]{D5E8D5}0.472 \\ \hline
\end{tabular}%
}
\caption{Results per setting of $p$ per method for GDP interpolation, trained on Seoul and tested on Taipei. Top-3 methods were marked using $\mu^{(1)}$, $\mu^{(2)}$ and $\mu^{(3)}$ (green), and the worst performing method was marked with $\mu^{(\otimes)}$ (red), determined using statistical significance testing at $\alpha=0.05$. The lowest error per condition, regardless of statistical significance, was marked \textbf{bold}.}
\label{tab:gdp_seoul_taipei}
\end{table}

\begin{table}[]
\centering
\resizebox{\textwidth}{!}{%
\begin{tabular}{|l|llllllllll|}
\hline
\multicolumn{1}{|c|}{}                                  & \multicolumn{10}{c|}{}                                                                                                                                                                                                                                                                                                                                                                                                    \\
\multicolumn{1}{|c|}{\multirow{-2}{*}{}}                & \multicolumn{10}{c|}{\multirow{-2}{*}{\textbf{Test data:   Taipei GDP}}}                                                                                                                                                                                                                                                                                                                                                  \\ \hline
\multicolumn{1}{|c|}{}                                  & \multicolumn{10}{c|}{\textbf{Training region: Taichung}}                                                                                                                                                                                                                                                                                                                                                                  \\ \cline{2-11} 
\multicolumn{1}{|c|}{}                                  & \multicolumn{2}{c|}{p=0.1}                                                     & \multicolumn{2}{c|}{p=0.3}                                                         & \multicolumn{2}{c|}{p=0.5}                                                     & \multicolumn{2}{c|}{p=0.7}                                                       & \multicolumn{2}{c|}{p=0.9}                                                      \\ \cline{2-11} 
\multicolumn{1}{|c|}{\multirow{-3}{*}{\textbf{Method}}} & \multicolumn{1}{c}{$\mu$}                      & \multicolumn{1}{c}{$\sigma$}  & \multicolumn{1}{c}{$\mu$}                          & \multicolumn{1}{c}{$\sigma$}  & \multicolumn{1}{c}{$\mu$}                      & \multicolumn{1}{c}{$\sigma$}  & \multicolumn{1}{c}{$\mu$}                        & \multicolumn{1}{c}{$\sigma$}  & \multicolumn{1}{c}{$\mu$}                       & \multicolumn{1}{c|}{$\sigma$} \\ \hline
OK                                                      & \cellcolor[HTML]{D5E8D5}2.6273$^{(2,3)}$       & \cellcolor[HTML]{D5E8D5}0.276 & \cellcolor[HTML]{D5E8D5}2.585$^{(1,2,3)}$          & \cellcolor[HTML]{D5E8D5}0.269 & \cellcolor[HTML]{D5E8D5}\textbf{2.521$^{(1)}$} & \cellcolor[HTML]{D5E8D5}0.233 & \cellcolor[HTML]{D5E8D5}\textbf{2.702$^{(1,2)}$} & \cellcolor[HTML]{D5E8D5}0.234 & \cellcolor[HTML]{D5E8D5}2.684$^{(1,2)}$         & \cellcolor[HTML]{D5E8D5}0.252 \\
UK                                                      & \cellcolor[HTML]{D5E8D5}2.6603$^{(2,3)}$       & \cellcolor[HTML]{D5E8D5}0.24  & \cellcolor[HTML]{D5E8D5}2.628$^{(1,2,3)}$          & \cellcolor[HTML]{D5E8D5}0.267 & \cellcolor[HTML]{D5E8D5}2.698$^{(2)}$          & \cellcolor[HTML]{D5E8D5}0.382 & \cellcolor[HTML]{D5E8D5}2.707$^{(1,2})$          & \cellcolor[HTML]{D5E8D5}0.284 & \cellcolor[HTML]{D5E8D5}\textbf{2.62$^{(1,2)}$} & \cellcolor[HTML]{D5E8D5}0.223 \\
Basic                                                   & \cellcolor[HTML]{FFFFFF}5.1524                 & \cellcolor[HTML]{FFFFFF}1.705 & \cellcolor[HTML]{FFFFFF}5.173                      & \cellcolor[HTML]{FFFFFF}0.895 & \cellcolor[HTML]{FFFFFF}5.245                  & \cellcolor[HTML]{FFFFFF}0.753 & \cellcolor[HTML]{FFFFFF}5.148                    & \cellcolor[HTML]{FFFFFF}0.381 & \cellcolor[HTML]{FFFFFF}5.08                    & \cellcolor[HTML]{FFFFFF}0.24  \\
SAR                                                     & \cellcolor[HTML]{FFFFFF}6.2848                 & \cellcolor[HTML]{FFFFFF}2.396 & \cellcolor[HTML]{FFFFFF}6.917                      & \cellcolor[HTML]{FFFFFF}4.324 & \cellcolor[HTML]{F8CECC}9.434$^{(\otimes})$    & \cellcolor[HTML]{F8CECC}13.95 & \cellcolor[HTML]{F8CECC}6.023$^{(\otimes)}$      & \cellcolor[HTML]{F8CECC}0.381 & \cellcolor[HTML]{F8CECC}6.065$^{(\otimes})$     & \cellcolor[HTML]{F8CECC}0.852 \\
MA                                                      & \cellcolor[HTML]{F8CECC}7.5805$^{(\otimes)}$   & \cellcolor[HTML]{F8CECC}5.457 & \cellcolor[HTML]{F8CECC}9.742$^{(\otimes)}$        & \cellcolor[HTML]{F8CECC}11.33 & \cellcolor[HTML]{F8CECC}7.996$^({\otimes})$    & \cellcolor[HTML]{F8CECC}4.825 & \cellcolor[HTML]{F8CECC}9.533$^{(\otimes)}$      & \cellcolor[HTML]{F8CECC}15.58 & \cellcolor[HTML]{FFFFFF}13.05                   & \cellcolor[HTML]{FFFFFF}17.24 \\
ARMA                                                    & \cellcolor[HTML]{FFFFFF}5.7533                 & \cellcolor[HTML]{FFFFFF}0.884 & \cellcolor[HTML]{FFFFFF}5.822                      & \cellcolor[HTML]{FFFFFF}0.898 & \cellcolor[HTML]{FFFFFF}5.852                  & \cellcolor[HTML]{FFFFFF}1.067 & \cellcolor[HTML]{F8CECC}6.041$^{(\otimes})$      & \cellcolor[HTML]{F8CECC}0.843 & \cellcolor[HTML]{FFFFFF}6.008                   & \cellcolor[HTML]{FFFFFF}1.072 \\
CNN                                                     & \cellcolor[HTML]{FFFFFF}6.1006                 & \cellcolor[HTML]{FFFFFF}2.251 & \cellcolor[HTML]{FFFFFF}5.914                      & \cellcolor[HTML]{FFFFFF}1.26  & \cellcolor[HTML]{FFFFFF}5.824                  & \cellcolor[HTML]{FFFFFF}1.454 & \cellcolor[HTML]{FFFFFF}5.683                    & \cellcolor[HTML]{FFFFFF}1.359 & \cellcolor[HTML]{FFFFFF}6.137                   & \cellcolor[HTML]{FFFFFF}2.114 \\ \hline
SD-MRP                                                  & \cellcolor[HTML]{FFFFFF}3.9926                 & \cellcolor[HTML]{FFFFFF}1.451 & \cellcolor[HTML]{FFFFFF}4.441                      & \cellcolor[HTML]{FFFFFF}0.784 & \cellcolor[HTML]{FFFFFF}4.375                  & \cellcolor[HTML]{FFFFFF}0.409 & \cellcolor[HTML]{FFFFFF}4.516                    & \cellcolor[HTML]{FFFFFF}0.471 & \cellcolor[HTML]{D5E8D5}4.913$^{(3)}$           & \cellcolor[HTML]{D5E8D5}0.442 \\
WP-MRP                                                  & \cellcolor[HTML]{FFFFFF}5.4675                 & \cellcolor[HTML]{FFFFFF}2.538 & \cellcolor[HTML]{FFFFFF}5.934                      & \cellcolor[HTML]{FFFFFF}1.924 & \cellcolor[HTML]{FFFFFF}5.284                  & \cellcolor[HTML]{FFFFFF}0.962 & \cellcolor[HTML]{FFFFFF}5.343                    & \cellcolor[HTML]{FFFFFF}0.718 & \cellcolor[HTML]{FFFFFF}5.183                   & \cellcolor[HTML]{FFFFFF}0.319 \\ \hline
O-MRP                                                   & \cellcolor[HTML]{D5E8D5}\textbf{1.356$^{(1)}$} & \cellcolor[HTML]{D5E8D5}0.49  & \cellcolor[HTML]{D5E8D5}\textbf{2.472$^{(1,2,3)}$} & \cellcolor[HTML]{D5E8D5}0.573 & \cellcolor[HTML]{D5E8D5}3.513$^{(3)}$          & \cellcolor[HTML]{D5E8D5}0.531 & \cellcolor[HTML]{D5E8D5}4.264$^{(3)}$            & \cellcolor[HTML]{D5E8D5}0.477 & \cellcolor[HTML]{D5E8D5}4.909$^{(3)}$           & \cellcolor[HTML]{D5E8D5}0.485 \\ \hline
\end{tabular}%
}
\caption{Results per setting of $p$ per method for GDP interpolation, trained on Taichung and tested on Taipei. Top-3 methods were marked using $\mu^{(1)}$, $\mu^{(2)}$ and $\mu^{(3)}$ (green), and the worst performing method was marked with $\mu^{(\otimes)}$ (red), determined using statistical significance testing at $\alpha=0.05$. The lowest error per condition, regardless of statistical significance, was marked \textbf{bold}.}
\label{tab:gdp_taichung_taipei2}
\end{table}

\begin{table}[]
\centering
\resizebox{\textwidth}{!}{%
\begin{tabular}{|l|llllllllll|}
\hline
\multicolumn{1}{|c|}{}                                  & \multicolumn{10}{c|}{}                                                                                                                                                                                                                                                                                                                                                                                               \\
\multicolumn{1}{|c|}{\multirow{-2}{*}{}}                & \multicolumn{10}{c|}{\multirow{-2}{*}{\textbf{Test data:   Daegu GDP}}}                                                                                                                                                                                                                                                                                                                                              \\ \hline
\multicolumn{1}{|c|}{}                                  & \multicolumn{10}{c|}{\textbf{Training region: Daegu}}                                                                                                                                                                                                                                                                                                                                                                \\ \cline{2-11} 
\multicolumn{1}{|c|}{}                                  & \multicolumn{2}{c|}{p=0.1}                                                      & \multicolumn{2}{c|}{p=0.3}                                                     & \multicolumn{2}{c|}{p=0.5}                                                    & \multicolumn{2}{c|}{p=0.7}                                                       & \multicolumn{2}{c|}{p=0.9}                                                     \\ \cline{2-11} 
\multicolumn{1}{|c|}{\multirow{-3}{*}{\textbf{Method}}} & \multicolumn{1}{c}{$\mu$}                       & \multicolumn{1}{c}{$\sigma$}  & \multicolumn{1}{c}{$\mu$}                      & \multicolumn{1}{c}{$\sigma$}  & \multicolumn{1}{c}{$\mu$}                     & \multicolumn{1}{c}{$\sigma$}  & \multicolumn{1}{c}{$\mu$}                        & \multicolumn{1}{c}{$\sigma$}  & \multicolumn{1}{c}{$\mu$}                      & \multicolumn{1}{c|}{$\sigma$} \\ \hline
OK                                                      & \cellcolor[HTML]{D5E8D5}2.5744$^{(3)}$          & \cellcolor[HTML]{D5E8D5}0.277 & \cellcolor[HTML]{FFFFFF}2.676                  & \cellcolor[HTML]{FFFFFF}0.332 & \cellcolor[HTML]{D5E8D5}2.69$^{(3)}$          & \cellcolor[HTML]{D5E8D5}0.33  & \cellcolor[HTML]{D5E8D5}2.611$^{(3)}$            & \cellcolor[HTML]{D5E8D5}0.255 & \cellcolor[HTML]{D5E8D5}2.733$^{(3)}$          & \cellcolor[HTML]{D5E8D5}0.264 \\
UK                                                      & \cellcolor[HTML]{FFFFFF}2.7399                  & \cellcolor[HTML]{FFFFFF}0.253 & \cellcolor[HTML]{FFFFFF}2.629                  & \cellcolor[HTML]{FFFFFF}0.29  & \cellcolor[HTML]{D5E8D5}2.652$^{(3)}$         & \cellcolor[HTML]{D5E8D5}0.269 & \cellcolor[HTML]{FFFFFF}2.726                    & \cellcolor[HTML]{FFFFFF}0.261 & \cellcolor[HTML]{D5E8D5}2.634$^{(3)}$          & \cellcolor[HTML]{D5E8D5}0.242 \\
Basic                                                   & \cellcolor[HTML]{D5E8D5}0.8961$^{(2)}$          & \cellcolor[HTML]{D5E8D5}0.054 & \cellcolor[HTML]{D5E8D5}0.905$^{(2)}$          & \cellcolor[HTML]{D5E8D5}0.084 & \cellcolor[HTML]{D5E8D5}0.889$^{(2)}$         & \cellcolor[HTML]{D5E8D5}0.174 & \cellcolor[HTML]{D5E8D5}\textbf{0.901$^{(1,2)}$} & \cellcolor[HTML]{D5E8D5}0.269 & \cellcolor[HTML]{D5E8D5}\textbf{1.055$^{(1)}$} & \cellcolor[HTML]{D5E8D5}0.738 \\
SAR                                                     & \cellcolor[HTML]{FFFFFF}2.4277                  & \cellcolor[HTML]{FFFFFF}0.97  & \cellcolor[HTML]{FFFFFF}2.962                  & \cellcolor[HTML]{FFFFFF}1.213 & \cellcolor[HTML]{FFFFFF}2.894                 & \cellcolor[HTML]{FFFFFF}0.561 & \cellcolor[HTML]{FFFFFF}4.674                    & \cellcolor[HTML]{FFFFFF}9.476 & \cellcolor[HTML]{FFFFFF}3.87                   & \cellcolor[HTML]{FFFFFF}1.635 \\
MA                                                      & \cellcolor[HTML]{F8CECC}3.5353$^{(\otimes)}$    & \cellcolor[HTML]{F8CECC}0.476 & \cellcolor[HTML]{F8CECC}3.584$^{(\otimes)}$    & \cellcolor[HTML]{F8CECC}0.738 & \cellcolor[HTML]{F8CECC}3.793$^{(\otimes)}$   & \cellcolor[HTML]{F8CECC}0.659 & \cellcolor[HTML]{F8CECC}4.127$^{(\otimes)}$      & \cellcolor[HTML]{F8CECC}1.524 & \cellcolor[HTML]{F8CECC}8.519$^{(\otimes)}$    & \cellcolor[HTML]{F8CECC}12.18 \\
ARMA                                                    & \cellcolor[HTML]{F8CECC}3.7675$^{(\otimes)}$    & \cellcolor[HTML]{F8CECC}0.985 & \cellcolor[HTML]{F8CECC}3.898$^{(\otimes)}$    & \cellcolor[HTML]{F8CECC}1.103 & \cellcolor[HTML]{F8CECC}4.125$^{(\otimes)}$   & \cellcolor[HTML]{F8CECC}1.065 & \cellcolor[HTML]{F8CECC}4.205$^{(\otimes)}$      & \cellcolor[HTML]{F8CECC}1.091 & \cellcolor[HTML]{FFFFFF}4.829                  & \cellcolor[HTML]{FFFFFF}5.114 \\
CNN                                                     & \cellcolor[HTML]{FFFFFF}2.9861                  & \cellcolor[HTML]{FFFFFF}0.201 & \cellcolor[HTML]{FFFFFF}2.967                  & \cellcolor[HTML]{FFFFFF}0.336 & \cellcolor[HTML]{FFFFFF}3.186                 & \cellcolor[HTML]{FFFFFF}1.789 & \cellcolor[HTML]{FFFFFF}3.841                    & \cellcolor[HTML]{FFFFFF}2.779 & \cellcolor[HTML]{FFFFFF}3.605                  & \cellcolor[HTML]{FFFFFF}1.206 \\ \hline
SD-MRP                                                  & \cellcolor[HTML]{FFFFFF}2.4405                  & \cellcolor[HTML]{FFFFFF}1.324 & \cellcolor[HTML]{D5E8D5}2.198$^{(3)}$          & \cellcolor[HTML]{D5E8D5}0.793 & \cellcolor[HTML]{D5E8D5}2.459$^{(3)}$         & \cellcolor[HTML]{D5E8D5}0.658 & \cellcolor[HTML]{FFFFFF}2.59                     & \cellcolor[HTML]{FFFFFF}0.433 & \cellcolor[HTML]{D5E8D5}2.745$^{(3)}$          & \cellcolor[HTML]{D5E8D5}0.346 \\
WP-MRP                                                  & \cellcolor[HTML]{FFFFFF}2.4715                  & \cellcolor[HTML]{FFFFFF}1.359 & \cellcolor[HTML]{FFFFFF}3.005                  & \cellcolor[HTML]{FFFFFF}0.951 & \cellcolor[HTML]{FFFFFF}3.03                  & \cellcolor[HTML]{FFFFFF}0.62  & \cellcolor[HTML]{FFFFFF}2.861                    & \cellcolor[HTML]{FFFFFF}0.297 & \cellcolor[HTML]{FFFFFF}2.939                  & \cellcolor[HTML]{FFFFFF}0.218 \\ \hline
O-MRP                                                   & \cellcolor[HTML]{D5E8D5}\textbf{0.4538$^{(1)}$} & \cellcolor[HTML]{D5E8D5}0.257 & \cellcolor[HTML]{D5E8D5}\textbf{0.652$^{(1)}$} & \cellcolor[HTML]{D5E8D5}0.171 & \cellcolor[HTML]{D5E8D5}\textbf{0.78$^{(1)}$} & \cellcolor[HTML]{D5E8D5}0.137 & \cellcolor[HTML]{D5E8D5}0.977$^{(1,2)}$          & \cellcolor[HTML]{D5E8D5}0.19  & \cellcolor[HTML]{D5E8D5}1.377$^{(2)}$          & \cellcolor[HTML]{D5E8D5}0.45  \\ \hline
\end{tabular}%
}
\caption{Results per setting of $p$ per method for GDP interpolation, trained on Daegu and tested on Daegu. Top-3 methods were marked using $\mu^{(1)}$, $\mu^{(2)}$ and $\mu^{(3)}$ (green), and the worst performing method was marked with $\mu^{(\otimes)}$ (red), determined using statistical significance testing at $\alpha=0.05$. The lowest error per condition, regardless of statistical significance, was marked \textbf{bold}.}
\label{tab:gdp_daegu_daegu}
\end{table}

\begin{table}[]
\centering
\resizebox{\textwidth}{!}{%
\begin{tabular}{|l|llllllllll|}
\hline
\multicolumn{1}{|c|}{}                                  & \multicolumn{10}{c|}{}                                                                                                                                                                                                                                                                                                                                                                                              \\
\multicolumn{1}{|c|}{\multirow{-2}{*}{}}                & \multicolumn{10}{c|}{\multirow{-2}{*}{\textbf{Test data:   Daegu GDP}}}                                                                                                                                                                                                                                                                                                                                             \\ \hline
\multicolumn{1}{|c|}{}                                  & \multicolumn{10}{c|}{\textbf{Training region: Seoul}}                                                                                                                                                                                                                                                                                                                                                               \\ \cline{2-11} 
\multicolumn{1}{|c|}{}                                  & \multicolumn{2}{c|}{p=0.1}                                                      & \multicolumn{2}{c|}{p=0.3}                                                     & \multicolumn{2}{c|}{p=0.5}                                                     & \multicolumn{2}{c|}{p=0.7}                                                     & \multicolumn{2}{c|}{p=0.9}                                                     \\ \cline{2-11} 
\multicolumn{1}{|c|}{\multirow{-3}{*}{\textbf{Method}}} & \multicolumn{1}{c}{$\mu$}                       & \multicolumn{1}{c}{$\sigma$}  & \multicolumn{1}{c}{$\mu$}                      & \multicolumn{1}{c}{$\sigma$}  & \multicolumn{1}{c}{$\mu$}                      & \multicolumn{1}{c}{$\sigma$}  & \multicolumn{1}{c}{$\mu$}                      & \multicolumn{1}{c}{$\sigma$}  & \multicolumn{1}{c}{$\mu$}                      & \multicolumn{1}{c|}{$\sigma$} \\ \hline
OK                                                      & \cellcolor[HTML]{F8CECC}10.129$^{(\otimes)}$    & \cellcolor[HTML]{F8CECC}0.767 & \cellcolor[HTML]{F8CECC}10.03$^{(\otimes)}$    & \cellcolor[HTML]{F8CECC}0.664 & \cellcolor[HTML]{F8CECC}10.27$^{(\otimes)}$    & \cellcolor[HTML]{F8CECC}0.68  & \cellcolor[HTML]{F8CECC}9.808$^{(\otimes)}$    & \cellcolor[HTML]{F8CECC}0.595 & \cellcolor[HTML]{F8CECC}10.19$^{(\otimes)}$    & \cellcolor[HTML]{F8CECC}0.749 \\
UK                                                      & \cellcolor[HTML]{F8CECC}10.166$^{(\otimes)}$    & \cellcolor[HTML]{F8CECC}0.615 & \cellcolor[HTML]{F8CECC}10.3$^{(\otimes)}$     & \cellcolor[HTML]{F8CECC}0.618 & \cellcolor[HTML]{F8CECC}9.971$^{(\otimes)}$    & \cellcolor[HTML]{F8CECC}0.647 & \cellcolor[HTML]{F8CECC}10.05$^{(\otimes)}$    & \cellcolor[HTML]{F8CECC}0.701 & \cellcolor[HTML]{F8CECC}10.21$^{(\otimes)}$    & \cellcolor[HTML]{F8CECC}0.947 \\
Basic                                                   & \cellcolor[HTML]{FFFFFF}3.7723                  & \cellcolor[HTML]{FFFFFF}1.457 & \cellcolor[HTML]{FFFFFF}3.757                  & \cellcolor[HTML]{FFFFFF}0.553 & \cellcolor[HTML]{FFFFFF}3.833                  & \cellcolor[HTML]{FFFFFF}0.413 & \cellcolor[HTML]{FFFFFF}3.842                  & \cellcolor[HTML]{FFFFFF}0.247 & \cellcolor[HTML]{FFFFFF}3.867                  & \cellcolor[HTML]{FFFFFF}0.111 \\
SAR                                                     & \cellcolor[HTML]{FFFFFF}4.7495                  & \cellcolor[HTML]{FFFFFF}1.606 & \cellcolor[HTML]{FFFFFF}3.805                  & \cellcolor[HTML]{FFFFFF}1.068 & \cellcolor[HTML]{FFFFFF}4.203                  & \cellcolor[HTML]{FFFFFF}1.422 & \cellcolor[HTML]{FFFFFF}4.812                  & \cellcolor[HTML]{FFFFFF}1.577 & \cellcolor[HTML]{FFFFFF}4.331                  & \cellcolor[HTML]{FFFFFF}1.369 \\
MA                                                      & \cellcolor[HTML]{FFFFFF}5.1548                  & \cellcolor[HTML]{FFFFFF}2.078 & \cellcolor[HTML]{FFFFFF}6.567                  & \cellcolor[HTML]{FFFFFF}4.967 & \cellcolor[HTML]{FFFFFF}6.085                  & \cellcolor[HTML]{FFFFFF}2.404 & \cellcolor[HTML]{FFFFFF}4.847                  & \cellcolor[HTML]{FFFFFF}1.836 & \cellcolor[HTML]{FFFFFF}5.334                  & \cellcolor[HTML]{FFFFFF}3.829 \\
ARMA                                                    & \cellcolor[HTML]{FFFFFF}4.5039                  & \cellcolor[HTML]{FFFFFF}1.181 & \cellcolor[HTML]{FFFFFF}4.675                  & \cellcolor[HTML]{FFFFFF}1.319 & \cellcolor[HTML]{FFFFFF}5.062                  & \cellcolor[HTML]{FFFFFF}1.326 & \cellcolor[HTML]{FFFFFF}5.412                  & \cellcolor[HTML]{FFFFFF}1.763 & \cellcolor[HTML]{FFFFFF}5.164                  & \cellcolor[HTML]{FFFFFF}1.559 \\
CNN                                                     & \cellcolor[HTML]{FFFFFF}3.0891                  & \cellcolor[HTML]{FFFFFF}0.165 & \cellcolor[HTML]{D5E8D5}3.091$^{(3)}$          & \cellcolor[HTML]{D5E8D5}0.165 & \cellcolor[HTML]{D5E8D5}3.089$^{(3)}$          & \cellcolor[HTML]{D5E8D5}0.165 & \cellcolor[HTML]{FFFFFF}3.089                  & \cellcolor[HTML]{FFFFFF}0.165 & \cellcolor[HTML]{D5E8D5}3.086$^{(3)}$          & \cellcolor[HTML]{D5E8D5}0.164 \\ \hline
SD-MRP                                                  & \cellcolor[HTML]{D5E8D5}2.1037$^{(2)}$          & \cellcolor[HTML]{D5E8D5}1.202 & \cellcolor[HTML]{D5E8D5}2.177$^{(2)}$          & \cellcolor[HTML]{D5E8D5}0.577 & \cellcolor[HTML]{D5E8D5}2.085$^{(2)}$          & \cellcolor[HTML]{D5E8D5}0.347 & \cellcolor[HTML]{D5E8D5}2.616${(2)}$           & \cellcolor[HTML]{D5E8D5}0.307 & \cellcolor[HTML]{D5E8D5}2.911$^{(2)}$          & \cellcolor[HTML]{D5E8D5}0.338 \\
WP-MRP                                                  & \cellcolor[HTML]{D5E8D5}2.7458$^{(3)}$          & \cellcolor[HTML]{D5E8D5}1.534 & \cellcolor[HTML]{D5E8D5}3.02$^{(3)}$           & \cellcolor[HTML]{D5E8D5}0.855 & \cellcolor[HTML]{D5E8D5}3.162$^{(3)}$          & \cellcolor[HTML]{D5E8D5}0.651 & \cellcolor[HTML]{D5E8D5}2.901$^{(3)}$          & \cellcolor[HTML]{D5E8D5}0.376 & \cellcolor[HTML]{FFFFFF}3.017                  & \cellcolor[HTML]{FFFFFF}0.228 \\ \hline
O-MRP                                                   & \cellcolor[HTML]{D5E8D5}\textbf{0.5288$^{(1)}$} & \cellcolor[HTML]{D5E8D5}0.327 & \cellcolor[HTML]{D5E8D5}\textbf{0.982$^{(1)}$} & \cellcolor[HTML]{D5E8D5}0.227 & \cellcolor[HTML]{D5E8D5}\textbf{1.444$^{(1)}$} & \cellcolor[HTML]{D5E8D5}0.252 & \cellcolor[HTML]{D5E8D5}\textbf{2.055$^{(1)}$} & \cellcolor[HTML]{D5E8D5}0.296 & \cellcolor[HTML]{D5E8D5}\textbf{2.346$^{(1)}$} & \cellcolor[HTML]{D5E8D5}0.352 \\ \hline
\end{tabular}%
}
\caption{Results per setting of $p$ per method for GDP interpolation, trained on Seoul and tested on Daegu. Top-3 methods were marked using $\mu^{(1)}$, $\mu^{(2)}$ and $\mu^{(3)}$ (green), and the worst performing method was marked with $\mu^{(\otimes)}$ (red), determined using statistical significance testing at $\alpha=0.05$. The lowest error per condition, regardless of statistical significance, was marked \textbf{bold}.}
\label{tab:gdp_seoul_daegu2}
\end{table}

\begin{table}[]
\centering
\resizebox{\textwidth}{!}{%
\begin{tabular}{|l|llllllllll|}
\hline
\multicolumn{1}{|c|}{}                                  & \multicolumn{10}{c|}{}                                                                                                                                                                                                                                                                                                                                                                                             \\
\multicolumn{1}{|c|}{\multirow{-2}{*}{}}                & \multicolumn{10}{c|}{\multirow{-2}{*}{\textbf{Test data:   Daegu GDP}}}                                                                                                                                                                                                                                                                                                                                            \\ \hline
\multicolumn{1}{|c|}{}                                  & \multicolumn{10}{c|}{\textbf{Training region: Taichung}}                                                                                                                                                                                                                                                                                                                                                           \\ \cline{2-11} 
\multicolumn{1}{|c|}{}                                  & \multicolumn{2}{c|}{p=0.1}                                                     & \multicolumn{2}{c|}{p=0.3}                                                     & \multicolumn{2}{c|}{p=0.5}                                                     & \multicolumn{2}{c|}{p=0.7}                                                     & \multicolumn{2}{c|}{p=0.9}                                                     \\ \cline{2-11} 
\multicolumn{1}{|c|}{\multirow{-3}{*}{\textbf{Method}}} & \multicolumn{1}{c}{$\mu$}                      & \multicolumn{1}{c}{$\sigma$}  & \multicolumn{1}{c}{$\mu$}                      & \multicolumn{1}{c}{$\sigma$}  & \multicolumn{1}{c}{$\mu$}                      & \multicolumn{1}{c}{$\sigma$}  & \multicolumn{1}{c}{$\mu$}                      & \multicolumn{1}{c}{$\sigma$}  & \multicolumn{1}{c}{$\mu$}                      & \multicolumn{1}{c|}{$\sigma$} \\ \hline
OK                                                      & \cellcolor[HTML]{D5E8D5}2.5227$^{(3)}$         & \cellcolor[HTML]{D5E8D5}0.268 & \cellcolor[HTML]{D5E8D5}2.621$^{(3)}$          & \cellcolor[HTML]{D5E8D5}0.273 & \cellcolor[HTML]{D5E8D5}2.615$^{(3)}$          & \cellcolor[HTML]{D5E8D5}0.217 & \cellcolor[HTML]{D5E8D5}2.591$^{(3)}$          & \cellcolor[HTML]{D5E8D5}0.224 & \cellcolor[HTML]{D5E8D5}2.675$^{(2,3)}$        & \cellcolor[HTML]{D5E8D5}0.234 \\
UK                                                      & \cellcolor[HTML]{FFFFFF}2.6946                 & \cellcolor[HTML]{FFFFFF}0.262 & \cellcolor[HTML]{D5E8D5}2.656$^{(3)}$          & \cellcolor[HTML]{D5E8D5}0.311 & \cellcolor[HTML]{FFFFFF}2.676                  & \cellcolor[HTML]{FFFFFF}0.308 & \cellcolor[HTML]{D5E8D5}2.665$^{(3)}$          & \cellcolor[HTML]{D5E8D5}0.22  & \cellcolor[HTML]{D5E8D5}2.702$^{(2,3)}$        & \cellcolor[HTML]{D5E8D5}0.278 \\
Basic                                                   & \cellcolor[HTML]{FFFFFF}3.3167                 & \cellcolor[HTML]{FFFFFF}1.177 & \cellcolor[HTML]{FFFFFF}3.408                  & \cellcolor[HTML]{FFFFFF}0.721 & \cellcolor[HTML]{FFFFFF}3.522                  & \cellcolor[HTML]{FFFFFF}0.527 & \cellcolor[HTML]{FFFFFF}3.63                   & \cellcolor[HTML]{FFFFFF}0.316 & \cellcolor[HTML]{FFFFFF}3.511                  & \cellcolor[HTML]{FFFFFF}0.155 \\
SAR                                                     & \cellcolor[HTML]{FFFFFF}3.1231                 & \cellcolor[HTML]{FFFFFF}0.342 & \cellcolor[HTML]{FFFFFF}3.179                  & \cellcolor[HTML]{FFFFFF}0.37  & \cellcolor[HTML]{FFFFFF}3.502                  & \cellcolor[HTML]{FFFFFF}1.871 & \cellcolor[HTML]{FFFFFF}3.428                  & \cellcolor[HTML]{FFFFFF}1.05  & \cellcolor[HTML]{FFFFFF}3.377                  & \cellcolor[HTML]{FFFFFF}1.29  \\
MA                                                      & \cellcolor[HTML]{F8CECC}4.3585$^{(\otimes)}$   & \cellcolor[HTML]{F8CECC}1.575 & \cellcolor[HTML]{F8CECC}9.844$^{(\otimes)}$    & \cellcolor[HTML]{F8CECC}12.13 & \cellcolor[HTML]{F8CECC}5.664$^{(\otimes)}$    & \cellcolor[HTML]{F8CECC}4.31  & \cellcolor[HTML]{F8CECC}8.666$^{(\otimes)}$    & \cellcolor[HTML]{F8CECC}16.13 & \cellcolor[HTML]{F8CECC}4.754$^{(\otimes)}$    & \cellcolor[HTML]{F8CECC}4.063 \\
ARMA                                                    & \cellcolor[HTML]{FFFFFF}3.8306                 & \cellcolor[HTML]{FFFFFF}1.151 & \cellcolor[HTML]{FFFFFF}3.641                  & \cellcolor[HTML]{FFFFFF}0.452 & \cellcolor[HTML]{FFFFFF}3.944                  & \cellcolor[HTML]{FFFFFF}0.773 & \cellcolor[HTML]{F8CECC}4.249$^{(\otimes)}$    & \cellcolor[HTML]{F8CECC}2.128 & \cellcolor[HTML]{F8CECC}3.979$^{(\otimes)}$    & \cellcolor[HTML]{F8CECC}0.359 \\
CNN                                                     & \cellcolor[HTML]{FFFFFF}3.1007                 & \cellcolor[HTML]{FFFFFF}0.17  & \cellcolor[HTML]{FFFFFF}3.127                  & \cellcolor[HTML]{FFFFFF}0.22  & \cellcolor[HTML]{FFFFFF}3.182                  & \cellcolor[HTML]{FFFFFF}0.405 & \cellcolor[HTML]{FFFFFF}3.158                  & \cellcolor[HTML]{FFFFFF}0.524 & \cellcolor[HTML]{FFFFFF}3.09                   & \cellcolor[HTML]{FFFFFF}0.152 \\ \hline
SD-MRP                                                  & \cellcolor[HTML]{D5E8D5}2.0466$^{(2)}$         & \cellcolor[HTML]{D5E8D5}1.031 & \cellcolor[HTML]{D5E8D5}2.101$^{(2)}$          & \cellcolor[HTML]{D5E8D5}0.602 & \cellcolor[HTML]{D5E8D5}2.183$^{(2)}$          & \cellcolor[HTML]{D5E8D5}0.453 & \cellcolor[HTML]{D5E8D5}2.428$^{(2)}$          & \cellcolor[HTML]{D5E8D5}0.359 & \cellcolor[HTML]{D5E8D5}2.689$^{(2,3)}$        & \cellcolor[HTML]{D5E8D5}0.275 \\
WP-MRP                                                  & \cellcolor[HTML]{FFFFFF}3.0431                 & \cellcolor[HTML]{FFFFFF}1.467 & \cellcolor[HTML]{FFFFFF}3.077                  & \cellcolor[HTML]{FFFFFF}0.847 & \cellcolor[HTML]{FFFFFF}2.879                  & \cellcolor[HTML]{FFFFFF}0.593 & \cellcolor[HTML]{FFFFFF}3.099                  & \cellcolor[HTML]{FFFFFF}0.331 & \cellcolor[HTML]{FFFFFF}3.059                  & \cellcolor[HTML]{FFFFFF}0.176 \\ \hline
O-MRP                                                   & \cellcolor[HTML]{D5E8D5}\textbf{0.462$^{(1)}$} & \cellcolor[HTML]{D5E8D5}0.243 & \cellcolor[HTML]{D5E8D5}\textbf{0.938$^{(1)}$} & \cellcolor[HTML]{D5E8D5}0.259 & \cellcolor[HTML]{D5E8D5}\textbf{1.498$^{(1)}$} & \cellcolor[HTML]{D5E8D5}0.251 & \cellcolor[HTML]{D5E8D5}\textbf{2.022$^{(1)}$} & \cellcolor[HTML]{D5E8D5}0.267 & \cellcolor[HTML]{D5E8D5}\textbf{2.493$^{(1)}$} & \cellcolor[HTML]{D5E8D5}0.328 \\ \hline
\end{tabular}%
}
\caption{Results per setting of $p$ per method for GDP interpolation, trained on Taichung and tested on Daegu. Top-3 methods were marked using $\mu^{(1)}$, $\mu^{(2)}$ and $\mu^{(3)}$ (green), and the worst performing method was marked with $\mu^{(\otimes)}$ (red), determined using statistical significance testing at $\alpha=0.05$. The lowest error per condition, regardless of statistical significance, was marked \textbf{bold}. }
\label{tab:gdp_taichung_daegu}
\end{table}

\begin{table}[]
\centering
\resizebox{\textwidth}{!}{%
\begin{tabular}{|l|llllllllll|}
\hline
\multicolumn{1}{|c|}{}                                  & \multicolumn{10}{c|}{}                                                                                                                                                                                                                                                                                                                                                                                              \\
\multicolumn{1}{|c|}{\multirow{-2}{*}{}}                & \multicolumn{10}{c|}{\multirow{-2}{*}{\textbf{Test data:   Daegu COVID-19}}}                                                                                                                                                                                                                                                                                                                                        \\ \hline
\multicolumn{1}{|c|}{}                                  & \multicolumn{10}{c|}{\textbf{Training region: Daegu}}                                                                                                                                                                                                                                                                                                                                                               \\ \cline{2-11} 
\multicolumn{1}{|c|}{}                                  & \multicolumn{2}{c|}{p=0.1}                                                      & \multicolumn{2}{c|}{p=0.3}                                                     & \multicolumn{2}{c|}{p=0.5}                                                     & \multicolumn{2}{c|}{p=0.7}                                                     & \multicolumn{2}{c|}{p=0.9}                                                     \\ \cline{2-11} 
\multicolumn{1}{|c|}{\multirow{-3}{*}{\textbf{Method}}} & \multicolumn{1}{c}{$\mu$}                       & \multicolumn{1}{c}{$\sigma$}  & \multicolumn{1}{c}{$\mu$}                      & \multicolumn{1}{c}{$\sigma$}  & \multicolumn{1}{c}{$\mu$}                      & \multicolumn{1}{c}{$\sigma$}  & \multicolumn{1}{c}{$\mu$}                      & \multicolumn{1}{c}{$\sigma$}  & \multicolumn{1}{c}{$\mu$}                      & \multicolumn{1}{c|}{$\sigma$} \\ \hline
OK                                                      & \cellcolor[HTML]{F8CECC}12.365$^{(\otimes)}$    & \cellcolor[HTML]{F8CECC}1.342 & \cellcolor[HTML]{F8CECC}13.64$^{(\otimes)}$    & \cellcolor[HTML]{F8CECC}1.475 & \cellcolor[HTML]{F8CECC}13.04$^{(\otimes)}$    & \cellcolor[HTML]{F8CECC}1.513 & \cellcolor[HTML]{F8CECC}13.17$^{(\otimes)}$    & \cellcolor[HTML]{F8CECC}1.341 & \cellcolor[HTML]{F8CECC}13.4$^{(\otimes)}$     & \cellcolor[HTML]{F8CECC}1.202 \\
UK                                                      & \cellcolor[HTML]{F8CECC}12.546$^{(\otimes)}$    & \cellcolor[HTML]{F8CECC}1.511 & \cellcolor[HTML]{F8CECC}13.09$^{(\otimes)}$    & \cellcolor[HTML]{F8CECC}1.415 & \cellcolor[HTML]{F8CECC}13.18$^{(\otimes)}$    & \cellcolor[HTML]{F8CECC}1.313 & \cellcolor[HTML]{F8CECC}13.13$^{(\otimes)}$    & \cellcolor[HTML]{F8CECC}1.55  & \cellcolor[HTML]{F8CECC}13.48$^{(\otimes)}$    & \cellcolor[HTML]{F8CECC}1.444 \\
Basic                                                   & \cellcolor[HTML]{FFFFFF}2.5102                  & \cellcolor[HTML]{FFFFFF}0.234 & \cellcolor[HTML]{FFFFFF}2.56                   & \cellcolor[HTML]{FFFFFF}0.461 & \cellcolor[HTML]{FFFFFF}2.435                  & \cellcolor[HTML]{FFFFFF}0.606 & \cellcolor[HTML]{FFFFFF}2.492                  & \cellcolor[HTML]{FFFFFF}1.006 & \cellcolor[HTML]{FFFFFF}2.279                  & \cellcolor[HTML]{FFFFFF}1.625 \\
SAR                                                     & \cellcolor[HTML]{D5E8D5}1.1801$^{(2)}$          & \cellcolor[HTML]{D5E8D5}0.836 & \cellcolor[HTML]{D5E8D5}1.638$^{(2,3)}$        & \cellcolor[HTML]{D5E8D5}0.679 & \cellcolor[HTML]{D5E8D5}1.63$^{(3)}$           & \cellcolor[HTML]{D5E8D5}0.629 & \cellcolor[HTML]{FFFFFF}2.437                  & \cellcolor[HTML]{FFFFFF}2.029 & \cellcolor[HTML]{FFFFFF}2.402                  & \cellcolor[HTML]{FFFFFF}1.632 \\
MA                                                      & \cellcolor[HTML]{FFFFFF}1.8745                  & \cellcolor[HTML]{FFFFFF}0.395 & \cellcolor[HTML]{FFFFFF}2.239                  & \cellcolor[HTML]{FFFFFF}1.085 & \cellcolor[HTML]{FFFFFF}3.912                  & \cellcolor[HTML]{FFFFFF}7.824 & \cellcolor[HTML]{FFFFFF}3.807                  & \cellcolor[HTML]{FFFFFF}3.683 & \cellcolor[HTML]{F8CECC}26.69$^{(\otimes)}$    & \cellcolor[HTML]{F8CECC}83    \\
ARMA                                                    & \cellcolor[HTML]{FFFFFF}2.3224                  & \cellcolor[HTML]{FFFFFF}1.26  & \cellcolor[HTML]{FFFFFF}2.355                  & \cellcolor[HTML]{FFFFFF}0.901 & \cellcolor[HTML]{FFFFFF}2.31                   & \cellcolor[HTML]{FFFFFF}0.675 & \cellcolor[HTML]{FFFFFF}2.682                  & \cellcolor[HTML]{FFFFFF}1.478 & \cellcolor[HTML]{FFFFFF}3.453                  & \cellcolor[HTML]{FFFFFF}2.227 \\
CNN                                                     & \cellcolor[HTML]{D5E8D5}1.6711$^{(3)}$          & \cellcolor[HTML]{D5E8D5}0.247 & \cellcolor[HTML]{D5E8D5}1.794$^{(2,3)}$        & \cellcolor[HTML]{D5E8D5}0.376 & \cellcolor[HTML]{FFFFFF}1.9                    & \cellcolor[HTML]{FFFFFF}0.662 & \cellcolor[HTML]{FFFFFF}1.879                  & \cellcolor[HTML]{FFFFFF}1.083 & \cellcolor[HTML]{D5E8D5}1.819$^{(2)}$          & \cellcolor[HTML]{D5E8D5}1.97  \\ \hline
SD-MRP                                                  & \cellcolor[HTML]{FFFFFF}2.1936                  & \cellcolor[HTML]{FFFFFF}2.399 & \cellcolor[HTML]{FFFFFF}1.925                  & \cellcolor[HTML]{FFFFFF}1.014 & \cellcolor[HTML]{D5E8D5}1.61$^{(2)}$           & \cellcolor[HTML]{D5E8D5}0.699 & \cellcolor[HTML]{D5E8D5}1.708$^{(3)}$          & \cellcolor[HTML]{D5E8D5}0.397 & \cellcolor[HTML]{D5E8D5}1.652$^{(3)}$          & \cellcolor[HTML]{D5E8D5}0.248 \\
WP-MRP                                                  & \cellcolor[HTML]{FFFFFF}2.0035                  & \cellcolor[HTML]{FFFFFF}2.427 & \cellcolor[HTML]{D5E8D5}1.948$^{(2,3)}$        & \cellcolor[HTML]{D5E8D5}2.024 & \cellcolor[HTML]{D5E8D5}1.632$^{(3)}$          & \cellcolor[HTML]{D5E8D5}0.679 & \cellcolor[HTML]{D5E8D5}1.585$^{(2)}$          & \cellcolor[HTML]{D5E8D5}0.452 & \cellcolor[HTML]{FFFFFF}1.67                   & \cellcolor[HTML]{FFFFFF}0.245 \\ \hline
O-MRP                                                   & \cellcolor[HTML]{D5E8D5}\textbf{0.2647$^{(1)}$} & \cellcolor[HTML]{D5E8D5}0.311 & \cellcolor[HTML]{D5E8D5}\textbf{0.611$^{(1)}$} & \cellcolor[HTML]{D5E8D5}0.311 & \cellcolor[HTML]{D5E8D5}\textbf{0.757$^{(1)}$} & \cellcolor[HTML]{D5E8D5}0.366 & \cellcolor[HTML]{D5E8D5}\textbf{1.008$^{(1)}$} & \cellcolor[HTML]{D5E8D5}0.313 & \cellcolor[HTML]{D5E8D5}\textbf{1.197$^{(1)}$} & \cellcolor[HTML]{D5E8D5}0.253 \\ \hline
\end{tabular}%
}
\caption{Results per setting of $p$ per method for COVID-19 interpolation, trained on Daegu and tested on Daegu. Top-3 methods were marked using $\mu^{(1)}$, $\mu^{(2)}$ and $\mu^{(3)}$ (green), and the worst performing method was marked with $\mu^{(\otimes)}$ (red), determined using statistical significance testing at $\alpha=0.05$. The lowest error per condition, regardless of statistical significance, was marked \textbf{bold}. }
\label{tab:covid_daegu_daegu2}
\end{table}

\begin{table}[]
\centering
\resizebox{\textwidth}{!}{%
\begin{tabular}{|l|llllllllll|}
\hline
\multicolumn{1}{|c|}{}                                  & \multicolumn{10}{c|}{}                                                                                                                                                                                                                                                                                                                                                                                              \\
\multicolumn{1}{|c|}{\multirow{-2}{*}{}}                & \multicolumn{10}{c|}{\multirow{-2}{*}{\textbf{Test data:   Daegu COVID-19}}}                                                                                                                                                                                                                                                                                                                                        \\ \hline
\multicolumn{1}{|c|}{}                                  & \multicolumn{10}{c|}{\textbf{Training region: Seoul}}                                                                                                                                                                                                                                                                                                                                                               \\ \cline{2-11} 
\multicolumn{1}{|c|}{}                                  & \multicolumn{2}{c|}{p=0.1}                                                      & \multicolumn{2}{c|}{p=0.3}                                                     & \multicolumn{2}{c|}{p=0.5}                                                     & \multicolumn{2}{c|}{p=0.7}                                                     & \multicolumn{2}{c|}{p=0.9}                                                     \\ \cline{2-11} 
\multicolumn{1}{|c|}{\multirow{-3}{*}{\textbf{Method}}} & \multicolumn{1}{c}{$\mu$}                       & \multicolumn{1}{c}{$\sigma$}  & \multicolumn{1}{c}{$\mu$}                      & \multicolumn{1}{c}{$\sigma$}  & \multicolumn{1}{c}{$\mu$}                      & \multicolumn{1}{c}{$\sigma$}  & \multicolumn{1}{c}{$\mu$}                      & \multicolumn{1}{c}{$\sigma$}  & \multicolumn{1}{c}{$\mu$}                      & \multicolumn{1}{c|}{$\sigma$} \\ \hline
OK                                                      & \cellcolor[HTML]{F8CECC}13.167$^{(\otimes)}$    & \cellcolor[HTML]{F8CECC}1.303 & \cellcolor[HTML]{FFFFFF}12.85                  & \cellcolor[HTML]{FFFFFF}1.759 & \cellcolor[HTML]{F8CECC}12.92$^{(\otimes)}$    & \cellcolor[HTML]{F8CECC}1.495 & \cellcolor[HTML]{F8CECC}13.02$^{(\otimes)}$    & \cellcolor[HTML]{F8CECC}1.518 & \cellcolor[HTML]{F8CECC}13.36$^{(\otimes)}$    & \cellcolor[HTML]{F8CECC}1.399 \\
UK                                                      & \cellcolor[HTML]{F8CECC}13.056$^{(\otimes)}$    & \cellcolor[HTML]{F8CECC}1.681 & \cellcolor[HTML]{F8CECC}13.4$^{(\otimes)}$     & \cellcolor[HTML]{F8CECC}1.487 & \cellcolor[HTML]{F8CECC}13.03$^{(\otimes)}$    & \cellcolor[HTML]{F8CECC}1.676 & \cellcolor[HTML]{F8CECC}13.53$^{(\otimes)}$    & \cellcolor[HTML]{F8CECC}1.51  & \cellcolor[HTML]{F8CECC}13.21$^{(\otimes)}$    & \cellcolor[HTML]{F8CECC}1.56  \\
Basic                                                   & \cellcolor[HTML]{FFFFFF}6.3494                  & \cellcolor[HTML]{FFFFFF}1.914 & \cellcolor[HTML]{FFFFFF}7.157                  & \cellcolor[HTML]{FFFFFF}1.274 & \cellcolor[HTML]{FFFFFF}7.051                  & \cellcolor[HTML]{FFFFFF}0.911 & \cellcolor[HTML]{FFFFFF}7.168                  & \cellcolor[HTML]{FFFFFF}0.681 & \cellcolor[HTML]{FFFFFF}7.022                  & \cellcolor[HTML]{FFFFFF}0.767 \\
SAR                                                     & \cellcolor[HTML]{FFFFFF}5.4214                  & \cellcolor[HTML]{FFFFFF}3.4   & \cellcolor[HTML]{FFFFFF}4.508                  & \cellcolor[HTML]{FFFFFF}2.157 & \cellcolor[HTML]{FFFFFF}4.545                  & \cellcolor[HTML]{FFFFFF}2.124 & \cellcolor[HTML]{FFFFFF}5.092                  & \cellcolor[HTML]{FFFFFF}2.754 & \cellcolor[HTML]{FFFFFF}5.066                  & \cellcolor[HTML]{FFFFFF}3.261 \\
MA                                                      & \cellcolor[HTML]{FFFFFF}4.9665                  & \cellcolor[HTML]{FFFFFF}2.465 & \cellcolor[HTML]{FFFFFF}5.833                  & \cellcolor[HTML]{FFFFFF}5.087 & \cellcolor[HTML]{FFFFFF}5.485                  & \cellcolor[HTML]{FFFFFF}2.952 & \cellcolor[HTML]{FFFFFF}4.831                  & \cellcolor[HTML]{FFFFFF}2.675 & \cellcolor[HTML]{FFFFFF}4.441                  & \cellcolor[HTML]{FFFFFF}2.641 \\
ARMA                                                    & \cellcolor[HTML]{FFFFFF}4.4368                  & \cellcolor[HTML]{FFFFFF}1.941 & \cellcolor[HTML]{FFFFFF}3.967                  & \cellcolor[HTML]{FFFFFF}1.274 & \cellcolor[HTML]{FFFFFF}4.43                   & \cellcolor[HTML]{FFFFFF}1.725 & \cellcolor[HTML]{FFFFFF}5.326                  & \cellcolor[HTML]{FFFFFF}2.887 & \cellcolor[HTML]{FFFFFF}4.128                  & \cellcolor[HTML]{FFFFFF}1.777 \\
CNN                                                     & \cellcolor[HTML]{D5E8D5}1.8406$^{(2,3)}$        & \cellcolor[HTML]{D5E8D5}0.05  & \cellcolor[HTML]{D5E8D5}1.84$^{(2,3)}$         & \cellcolor[HTML]{D5E8D5}0.052 & \cellcolor[HTML]{D5E8D5}1.847$^{(2,3)}$        & \cellcolor[HTML]{D5E8D5}0.049 & \cellcolor[HTML]{D5E8D5}1.85$^{(3)}$           & \cellcolor[HTML]{D5E8D5}0.054 & \cellcolor[HTML]{FFFFFF}2.09                   & \cellcolor[HTML]{FFFFFF}0.076 \\ \hline
SD-MRP                                                  & \cellcolor[HTML]{D5E8D5}1.7902$^{(2,3)}$        & \cellcolor[HTML]{D5E8D5}1.745 & \cellcolor[HTML]{D5E8D5}1.862$^{(2,3)}$        & \cellcolor[HTML]{D5E8D5}0.82  & \cellcolor[HTML]{D5E8D5}1.81$^{(2,3)}$         & \cellcolor[HTML]{D5E8D5}0.467 & \cellcolor[HTML]{FFFFFF}1.968                  & \cellcolor[HTML]{FFFFFF}0.23  & \cellcolor[HTML]{D5E8D5}1.809$^{(3)}$          & \cellcolor[HTML]{D5E8D5}0.125 \\
WP-MRP                                                  & \cellcolor[HTML]{D5E8D5}2.0563$^{(2,3)}$        & \cellcolor[HTML]{D5E8D5}2.958 & \cellcolor[HTML]{D5E8D5}1.683$^{(2,3)}$        & \cellcolor[HTML]{D5E8D5}0.912 & \cellcolor[HTML]{D5E8D5}3.962$^{(2,3)}$        & \cellcolor[HTML]{D5E8D5}15.4  & \cellcolor[HTML]{D5E8D5}1.603$^{(2)}$          & \cellcolor[HTML]{D5E8D5}0.479 & \cellcolor[HTML]{D5E8D5}1.664$^{(2)}$          & \cellcolor[HTML]{D5E8D5}0.221 \\ \hline
O-MRP                                                   & \cellcolor[HTML]{D5E8D5}\textbf{0.3441$^{(1)}$} & \cellcolor[HTML]{D5E8D5}0.402 & \cellcolor[HTML]{D5E8D5}\textbf{0.745$^{(1)}$} & \cellcolor[HTML]{D5E8D5}0.382 & \cellcolor[HTML]{D5E8D5}\textbf{1.219$^{(1)}$} & \cellcolor[HTML]{D5E8D5}0.448 & \cellcolor[HTML]{D5E8D5}\textbf{1.461$^{(1)}$} & \cellcolor[HTML]{D5E8D5}0.347 & \cellcolor[HTML]{D5E8D5}\textbf{1.613$^{(1)}$} & \cellcolor[HTML]{D5E8D5}0.152 \\ \hline
\end{tabular}%
}
\caption{Results per setting of $p$ per method for COVID-19 interpolation, trained on Seoul and tested on Daegu. Top-3 methods were marked using $\mu^{(1)}$, $\mu^{(2)}$ and $\mu^{(3)}$ (green), and the worst performing method was marked with $\mu^{(\otimes)}$ (red), determined using statistical significance testing at $\alpha=0.05$. The lowest error per condition, regardless of statistical significance, was marked \textbf{bold}. }
\label{tab:covid_seoul_daegu}
\end{table}

\newpage
\section{Graphs}
We would also like to use this appendix to provide access to line graphs for all combinations of training and test sets, for both datasets. In the thesis, only two of these figured were included as examples. In this section of the appendix, every combination of (target variable dataset, training set, test set) has one graph for each of the following sets of methods:
\begin{itemize}
    \item \textbf{All}: all $10$ methods available
    \item \textbf{Purely spatial methods}: ordinary kriging, universal kriging, SD-MRP and O-MRP
    \item \textbf{(Spatial) regression methods}: basic regression, SAR, MA, ARMA, CNN and WP-MRP
    \item \textbf{MRP methods}: SD-MRP, WP-MRP and O-MRP
    \item \textbf{Most competitive}: ordinary kriging, universal kriging, SD-MRP and O-MRP
\end{itemize}

The figures are generally consistent with the example figure we included in the thesis. However, we can see the lower errors for kriging-based methods reflected in the results for GDP interpolation trained on Taichung and tested on Taipei in Figure 8.12, as well as some strange outliers for high $p$ for MA on COVID-19 interpolation trained and tested on Daegu. It was also unexpected to see the errors of SD-MRP and WP-MRP decrease, rather than increase, in Figure 8.34.

\subsection{GDP}

This section will cover the interpolation of GDP tested on Taipei and Daegu. All figures will use scientific notations for the y-axis for easy readability.

\subsubsection{Test region: Taipei}

In general, the results for Taipei were not as favourable for MRPs as the results for Daegu. The most notable example of this is the performance of the two kriging methods, whose lines tend to be lower than all others except O-MRP. Moreover, because the kriging methods seem so unaffected by the setting of $p$, in several cases they even overtake O-MRP for higher $p$. Only when trained on Taipei did the kriging methods perform poorly.

\goodbreak

\textbf{Training region: Taichung}

\begin{figure}[H]
  \centering
  \includegraphics[width=0.55\linewidth]{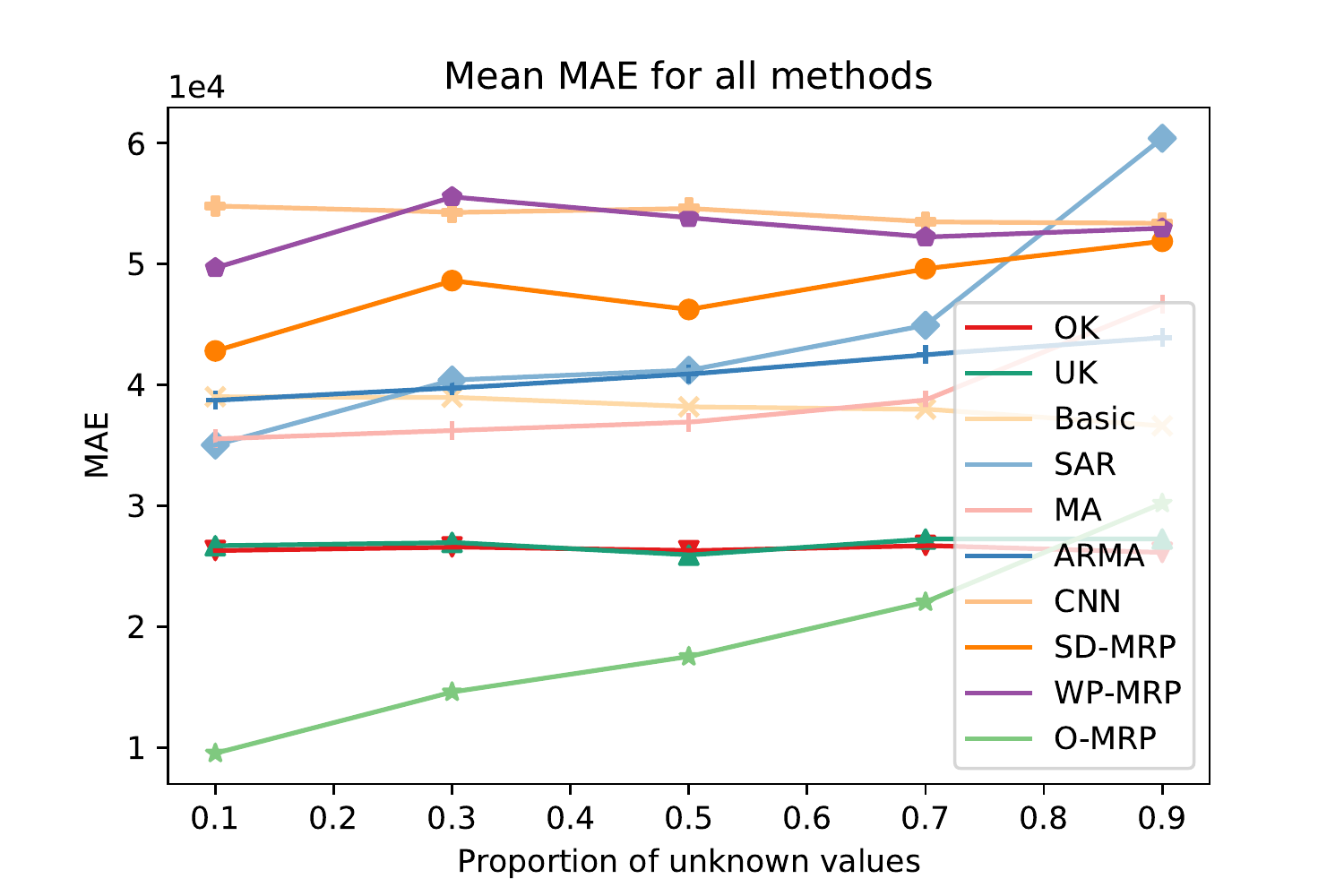}
  \caption{Mean MAE as a function of $p$ for GDP interpolation trained on Taipei and tested on Taipei: all methods.}
\end{figure}\par 

\begin{figure}[H]
  \centering
  \includegraphics[width=0.55\linewidth]{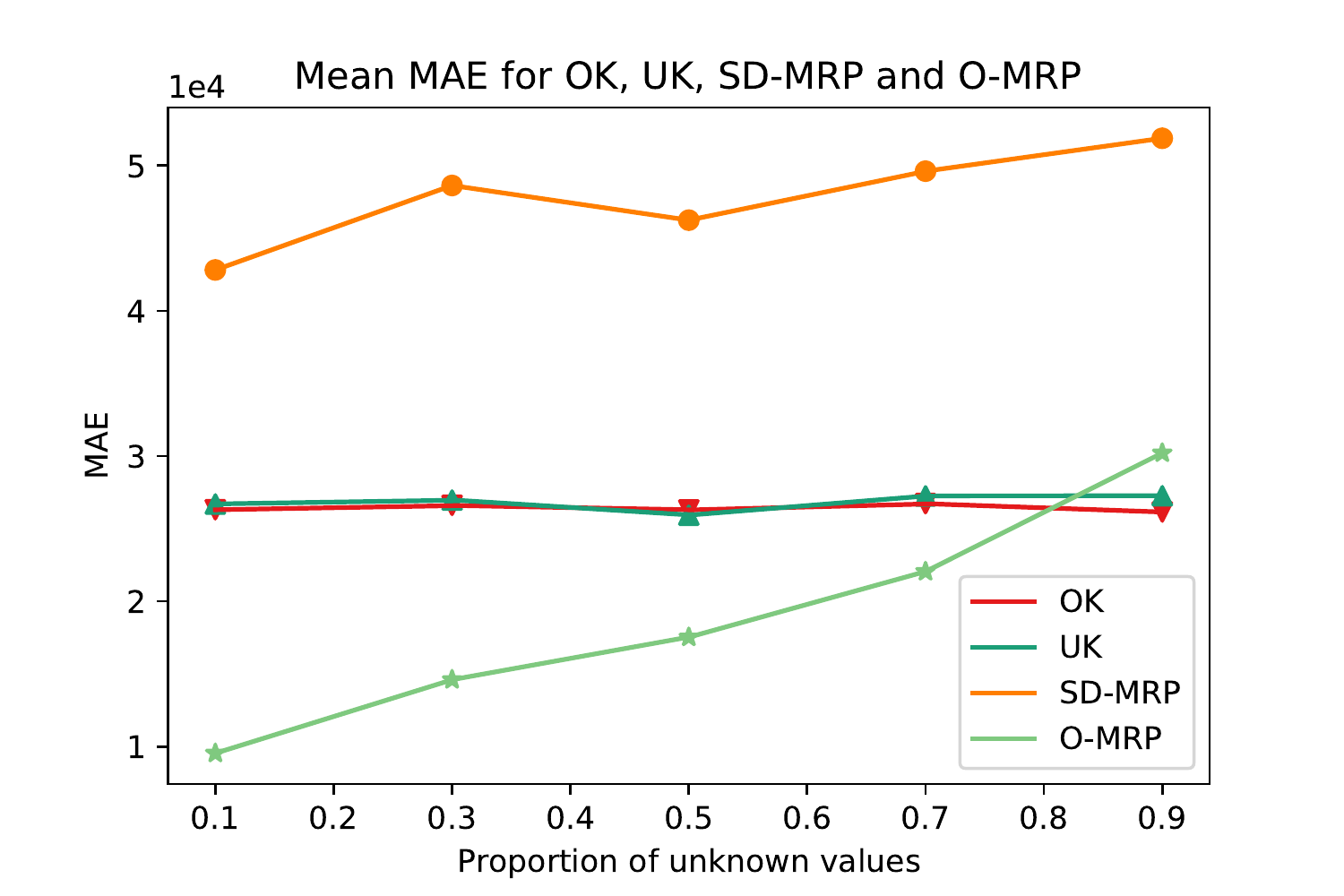}
  \caption{Mean MAE as a function of $p$ for GDP interpolation trained on Taipei and tested on Taipei: purely spatial methods.}
\end{figure}\par 

\begin{figure}[H]
  \centering
  \includegraphics[width=0.55\linewidth]{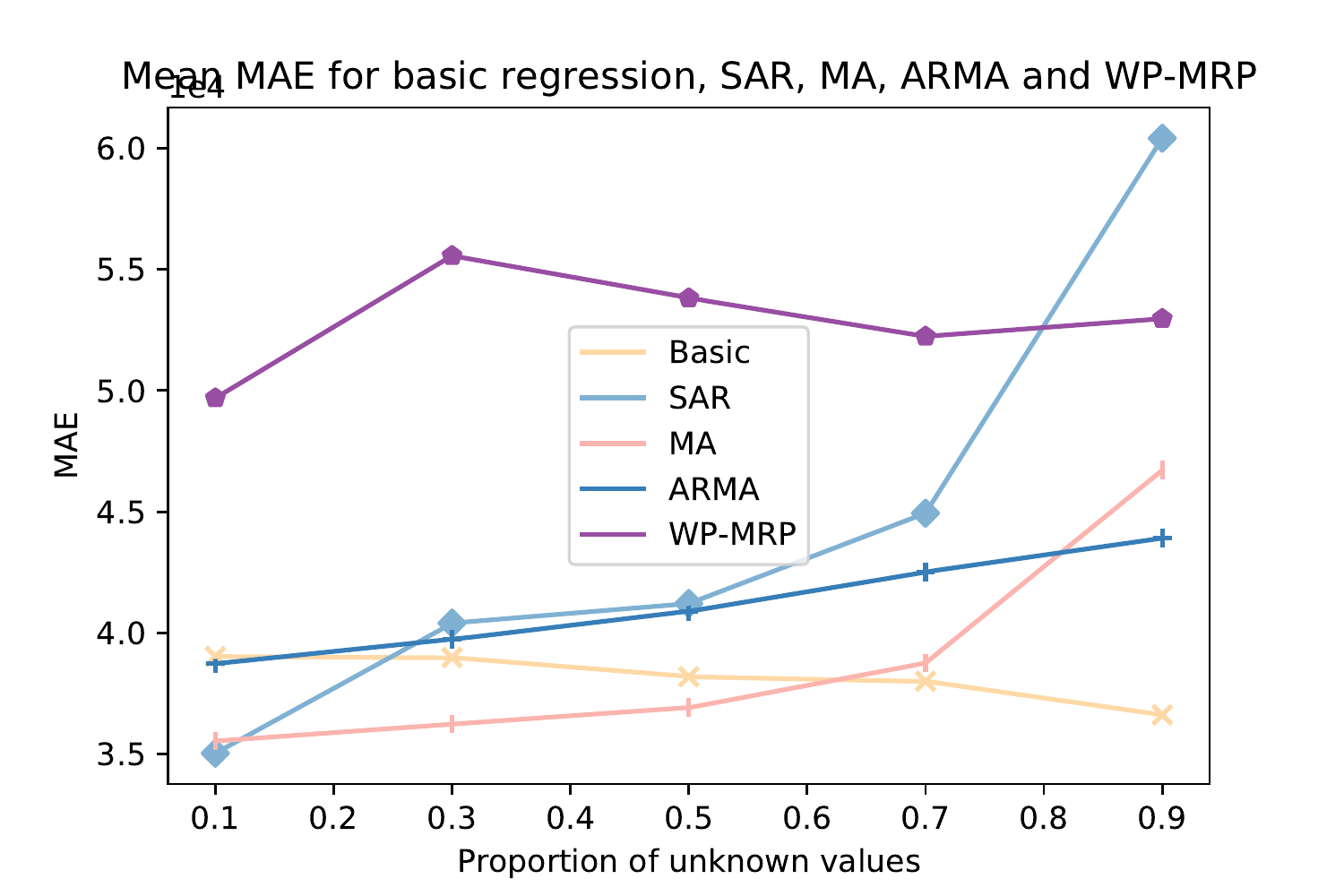}
  \caption{Mean MAE as a function of $p$ for GDP interpolation trained on Taipei and tested on Taipei: (spatial) regression methods.}
\end{figure}\par 

\begin{figure}[H]
  \centering
  \includegraphics[width=0.55\linewidth]{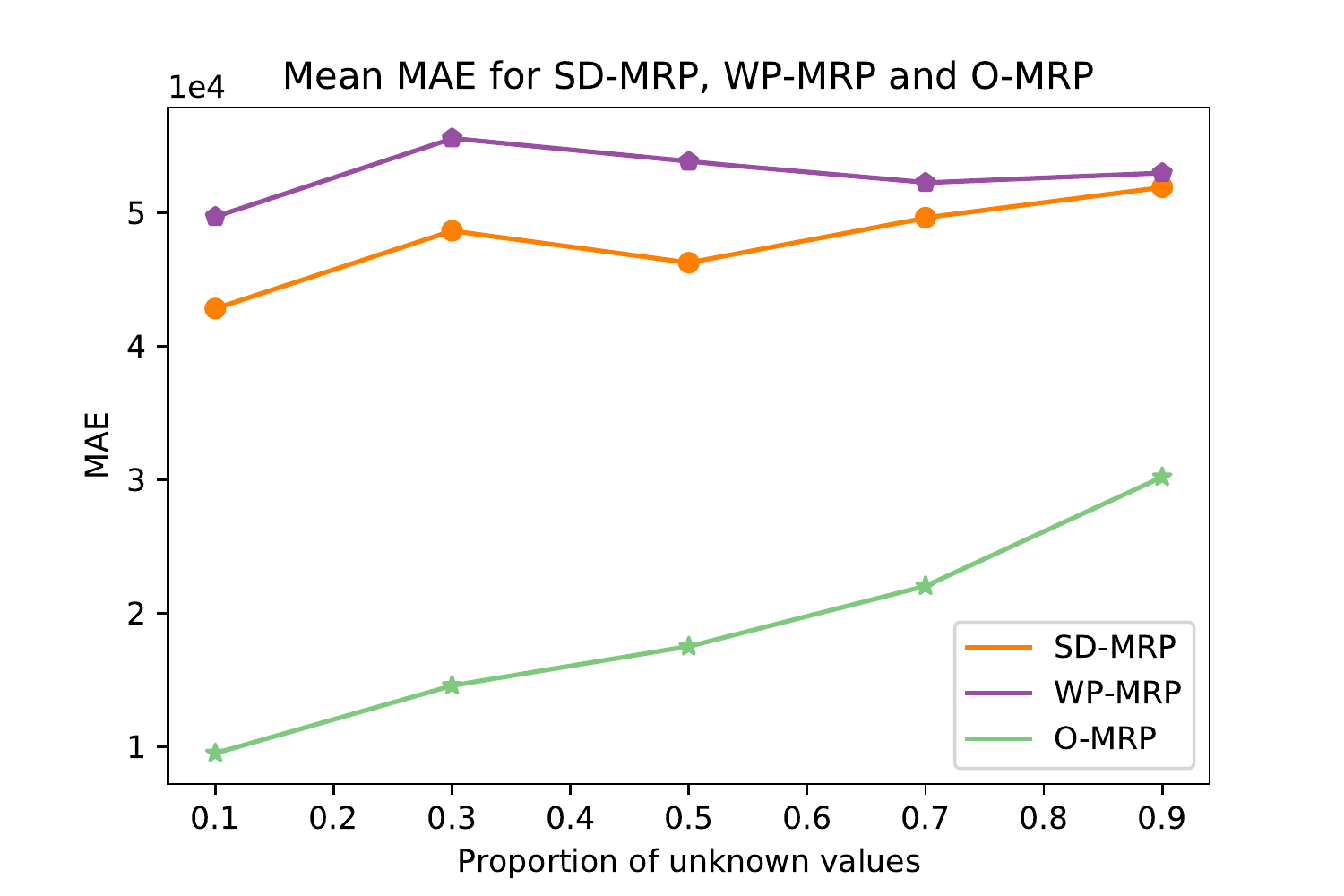}
  \caption{Mean MAE as a function of $p$ for GDP interpolation trained on Taipei and tested on Taipei: MRP methods.}
\end{figure}\par 

\begin{figure}[H]
  \centering
  \includegraphics[width=0.55\linewidth]{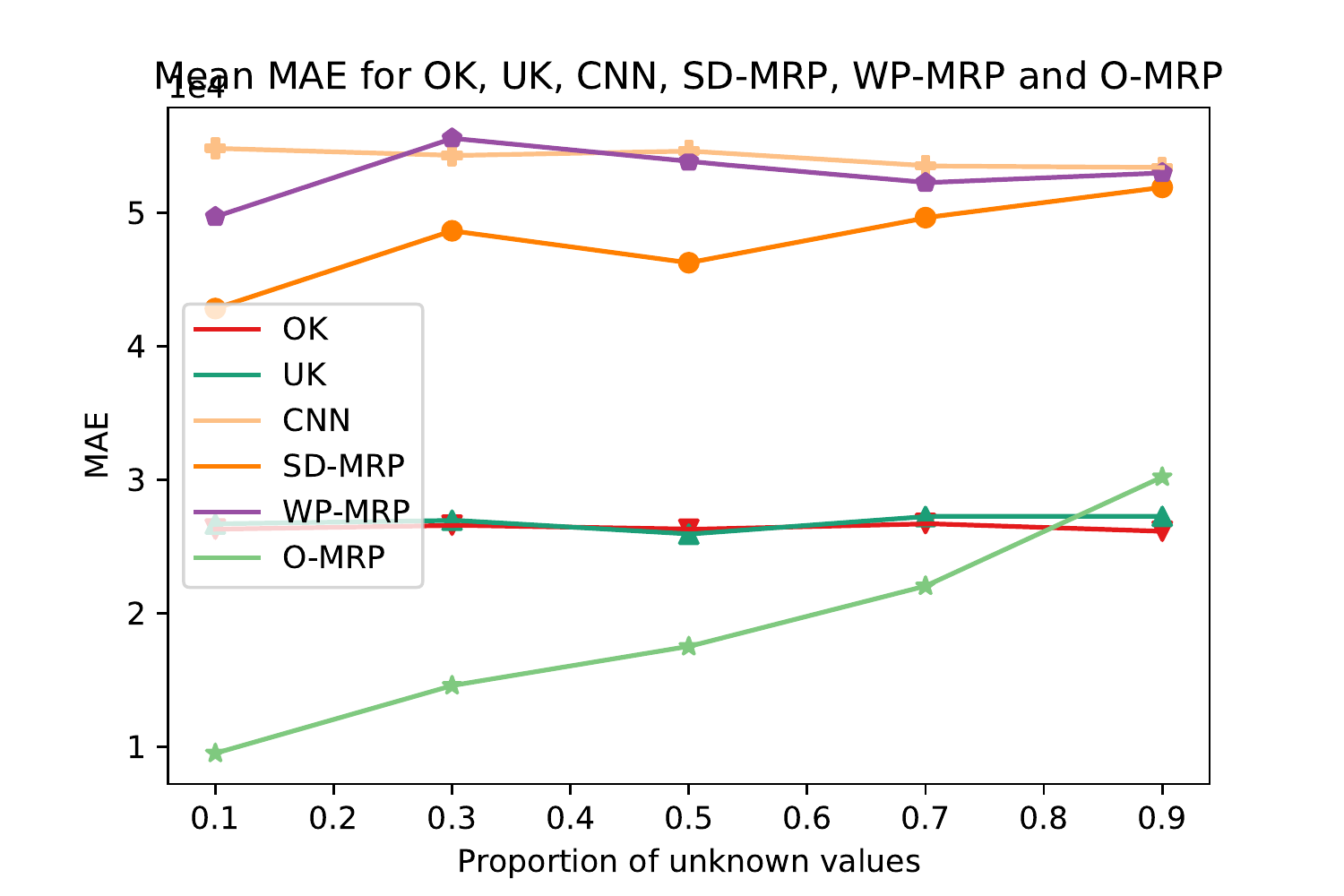}
  \caption{Mean MAE as a function of $p$ for GDP interpolation trained on Taipei and tested on Taipei: most competitive methods.}
\end{figure}\par

\textbf{Training region: Seoul}

\begin{figure}[H]
  \centering
  \includegraphics[width=0.55\linewidth]{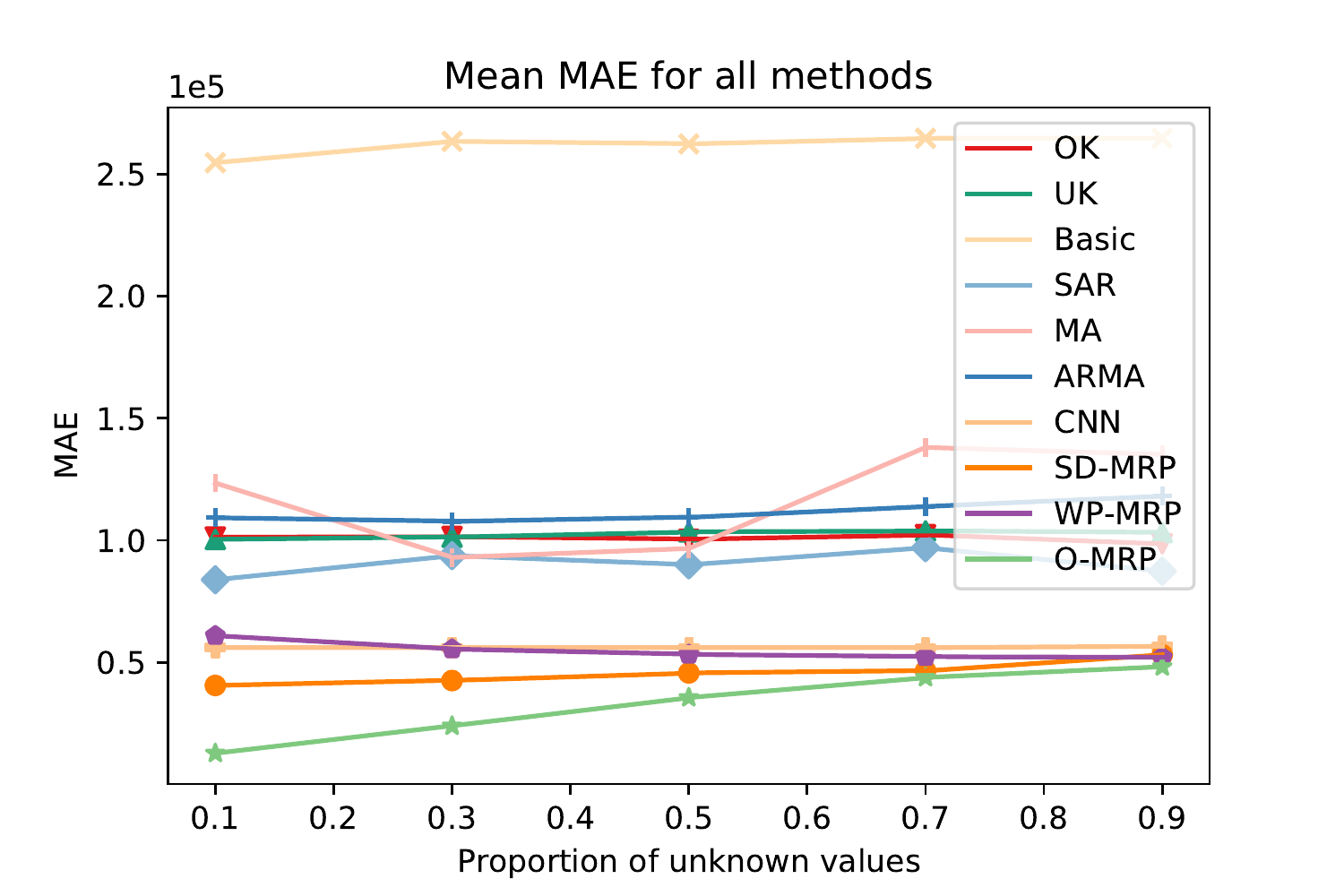}
  \caption{Mean MAE as a function of $p$ for GDP interpolation trained on Seoul and tested on Taipei: all methods.}
\end{figure}\par 

\begin{figure}[H]
  \centering
  \includegraphics[width=0.55\linewidth]{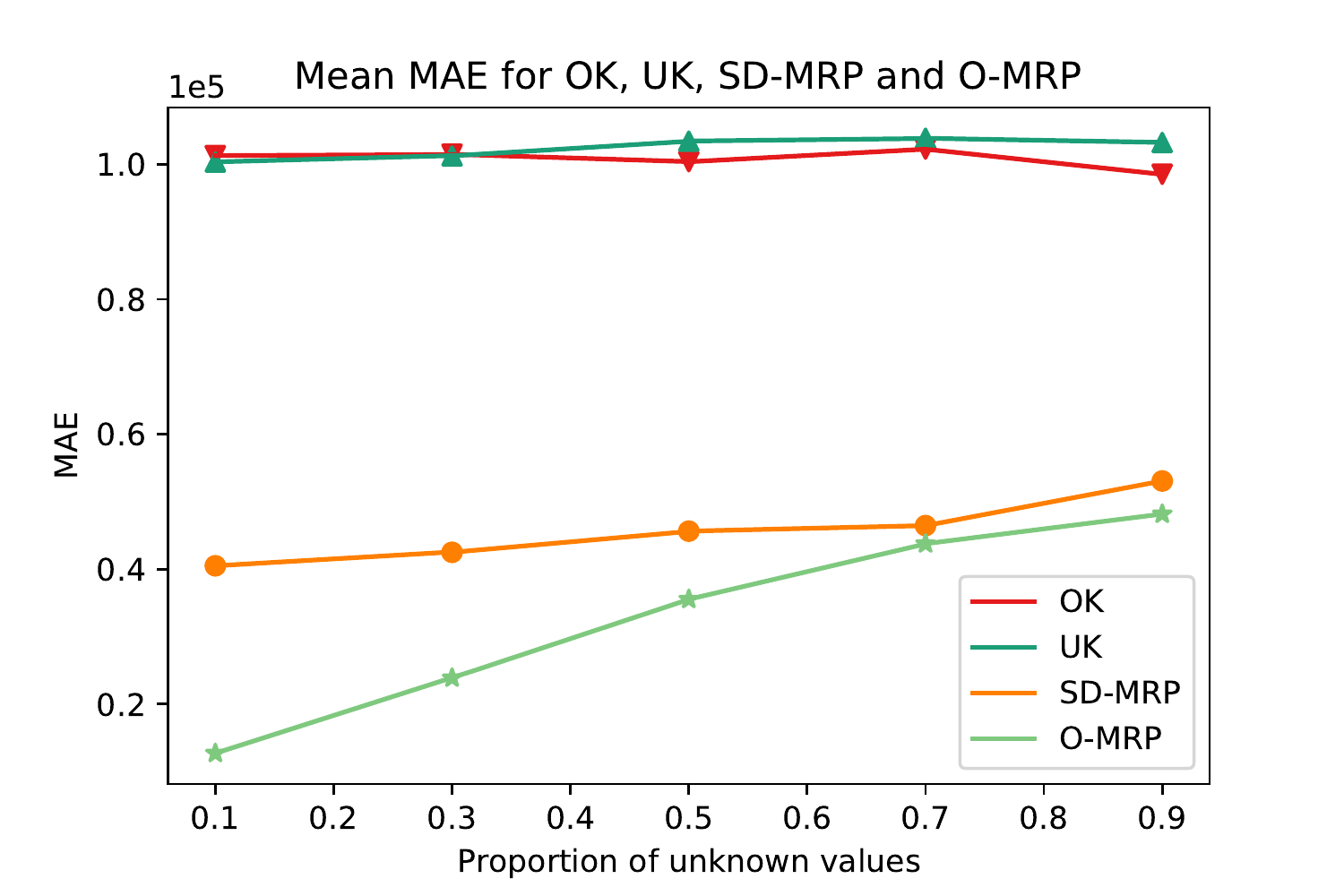}
  \caption{Mean MAE as a function of $p$ for GDP interpolation trained on Seoul and tested on Taipei: purely spatial methods.}
\end{figure}\par 

\begin{figure}[H]
  \centering
  \includegraphics[width=0.55\linewidth]{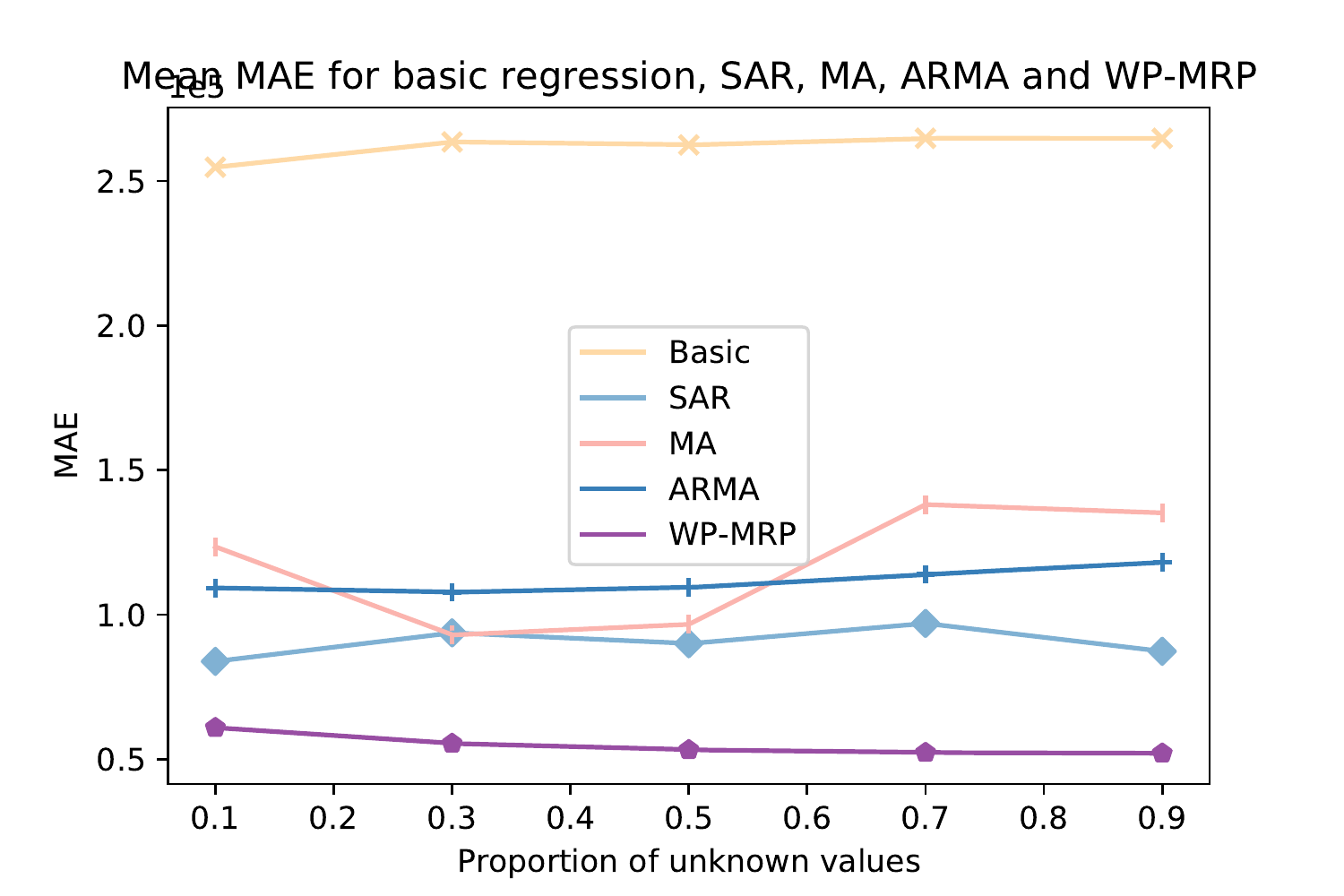}
  \caption{Mean MAE as a function of $p$ for GDP interpolation trained on Seoul and tested on Taipei: (spatial) regression methods.}
\end{figure}\par 

\begin{figure}[H]
  \centering
  \includegraphics[width=0.55\linewidth]{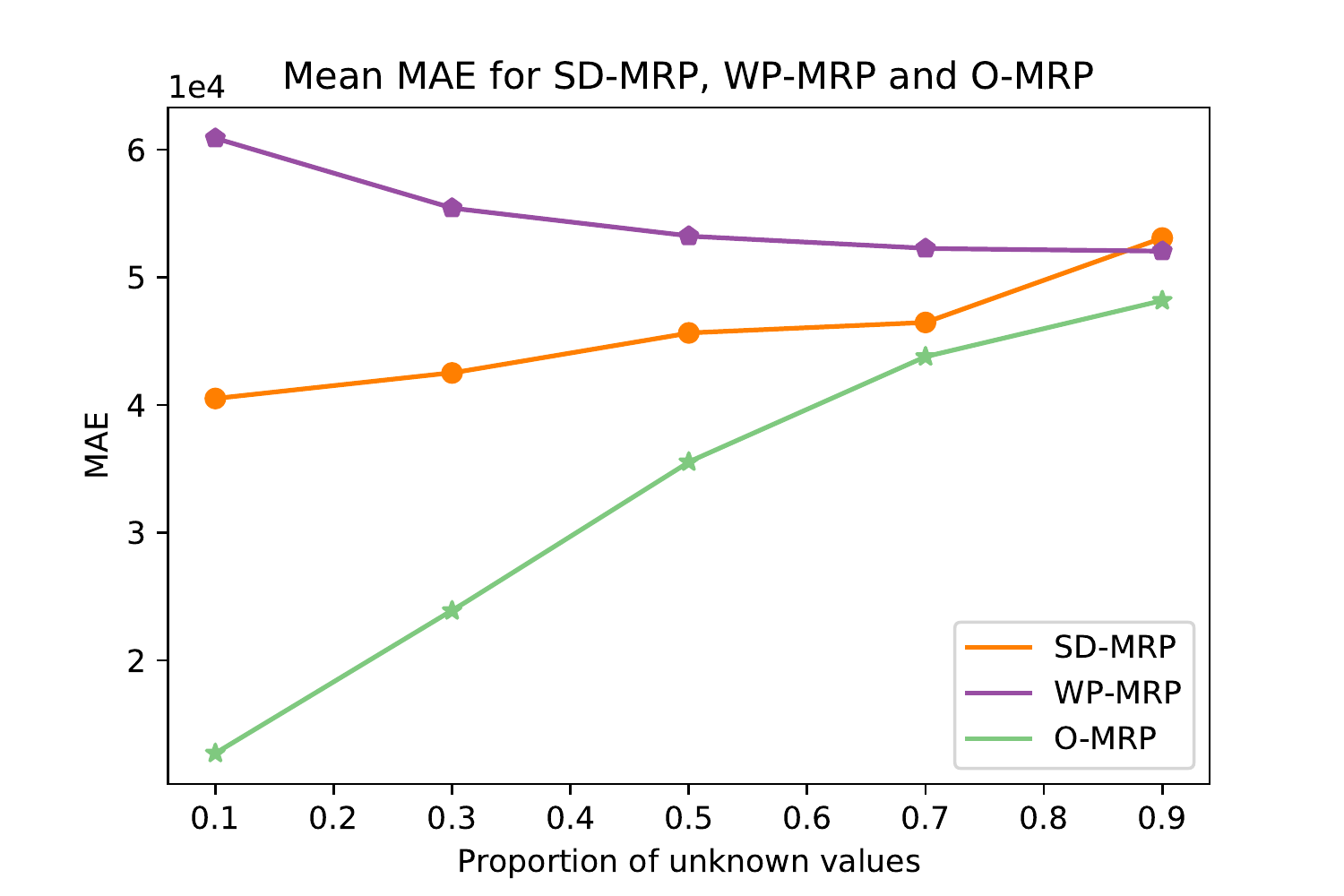}
  \caption{Mean MAE as a function of $p$ for GDP interpolation trained on Seoul and tested on Taipei: MRP methods.}
\end{figure}\par 

\begin{figure}[H]
  \centering
  \includegraphics[width=0.55\linewidth]{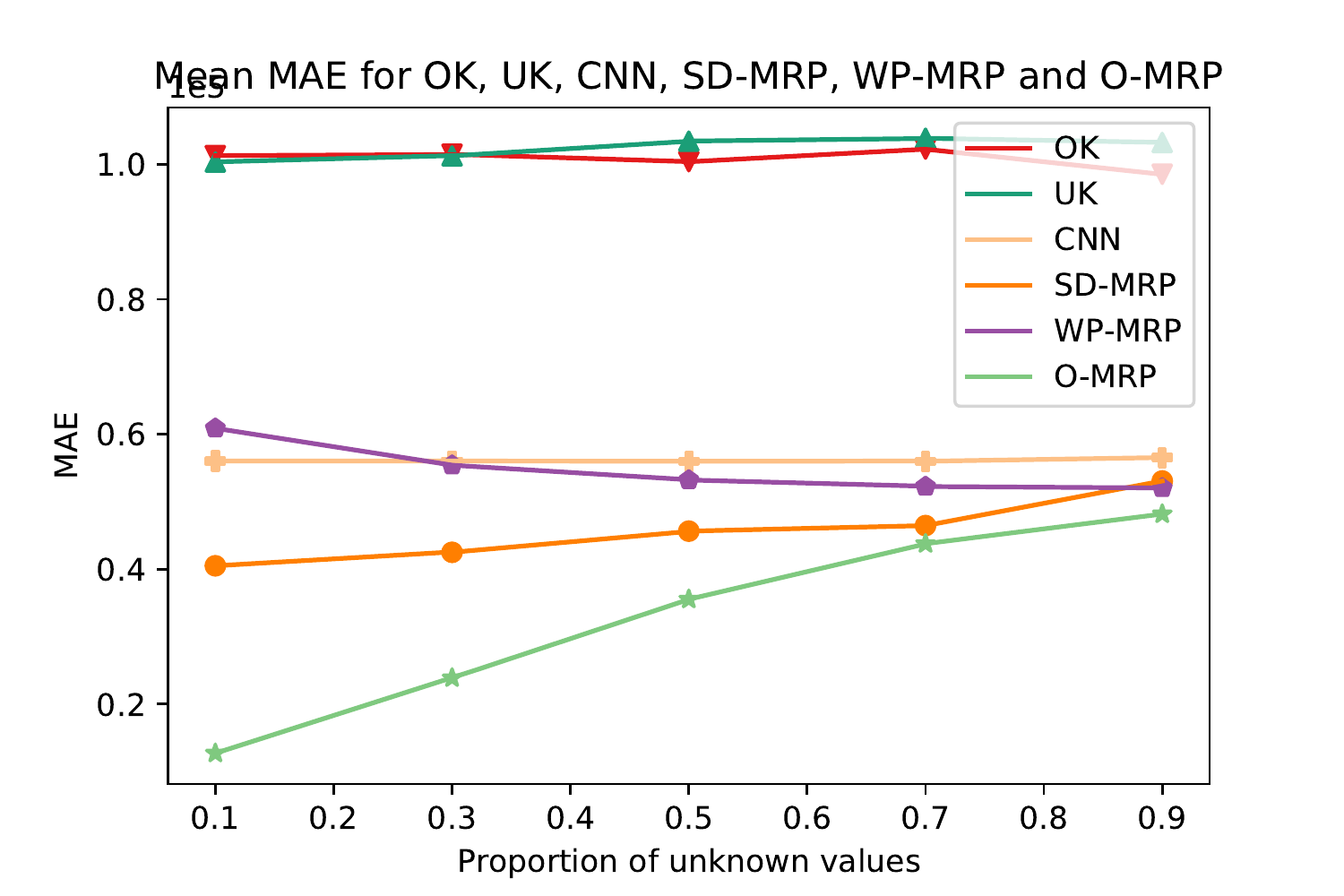}
  \caption{Mean MAE as a function of $p$ for GDP interpolation trained on Seoul and tested on Taipei: most competitive methods.}
\end{figure}\par

\textbf{Training region: Taichung}

\begin{figure}[H]
  \centering
  \includegraphics[width=0.55\linewidth]{../Media/appendix/GDP/Taichung_Taipei/all.pdf}
  \caption{Mean MAE as a function of $p$ for GDP interpolation trained on Taichung and tested on Taipei: all methods.}
\end{figure}\par 

\begin{figure}[H]
  \centering
  \includegraphics[width=0.55\linewidth]{../Media/appendix/GDP/Taichung_Taipei/label.pdf}
  \caption{Mean MAE as a function of $p$ for GDP interpolation trained on Taichung and tested on Taipei: purely spatial methods.}
\end{figure}\par 

\begin{figure}[H]
  \centering
  \includegraphics[width=0.55\linewidth]{../Media/appendix/GDP/Taichung_Taipei/reg.pdf}
  \caption{Mean MAE as a function of $p$ for GDP interpolation trained on Taichung and tested on Taipei: (spatial) regression methods.}
\end{figure}\par 

\begin{figure}[H]
  \centering
  \includegraphics[width=0.55\linewidth]{../Media/appendix/GDP/Taichung_Taipei/mrp.pdf}
  \caption{Mean MAE as a function of $p$ for GDP interpolation trained on Taichung and tested on Taipei: MRP methods.}
\end{figure}\par 

\begin{figure}[H]
  \centering
  \includegraphics[width=0.55\linewidth]{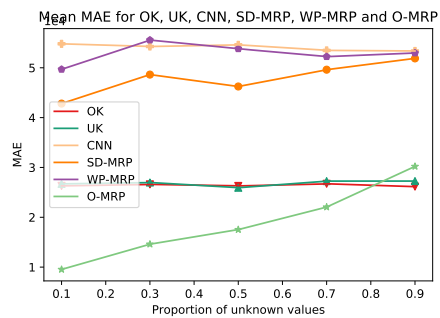}
  \caption{Mean MAE as a function of $p$ for GDP interpolation trained on Taichung and tested on Taipei: most competitive methods.}
\end{figure}\par

\subsubsection{Test region: Daegu}

The results for Daegu were more favourable to MRPs than the results for Taipei. In almost all graphs involving MRP methods, they can generally be found at the bottom of the figure with the lowest errors. Additionally, at times (such as in Figure 28) MA performed highly erratically when tested on Daegu.\\

\textbf{Training region: Daegu}

\begin{figure}[H]
  \centering
  \includegraphics[width=0.55\linewidth]{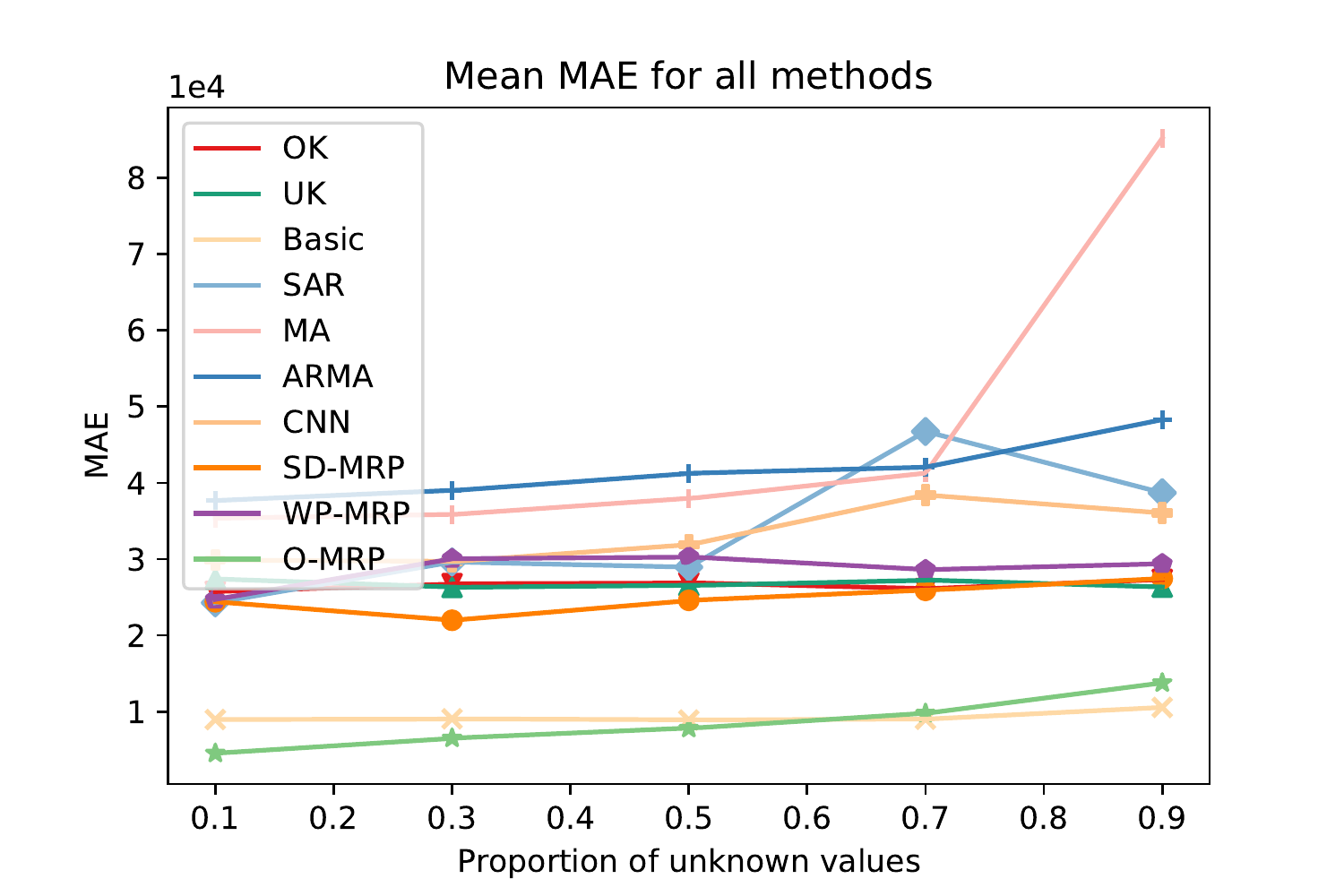}
  \caption{Mean MAE as a function of $p$ for GDP interpolation trained on Daegu and tested on Daegu: all methods.}
\end{figure}\par 

\begin{figure}[H]
  \centering
  \includegraphics[width=0.55\linewidth]{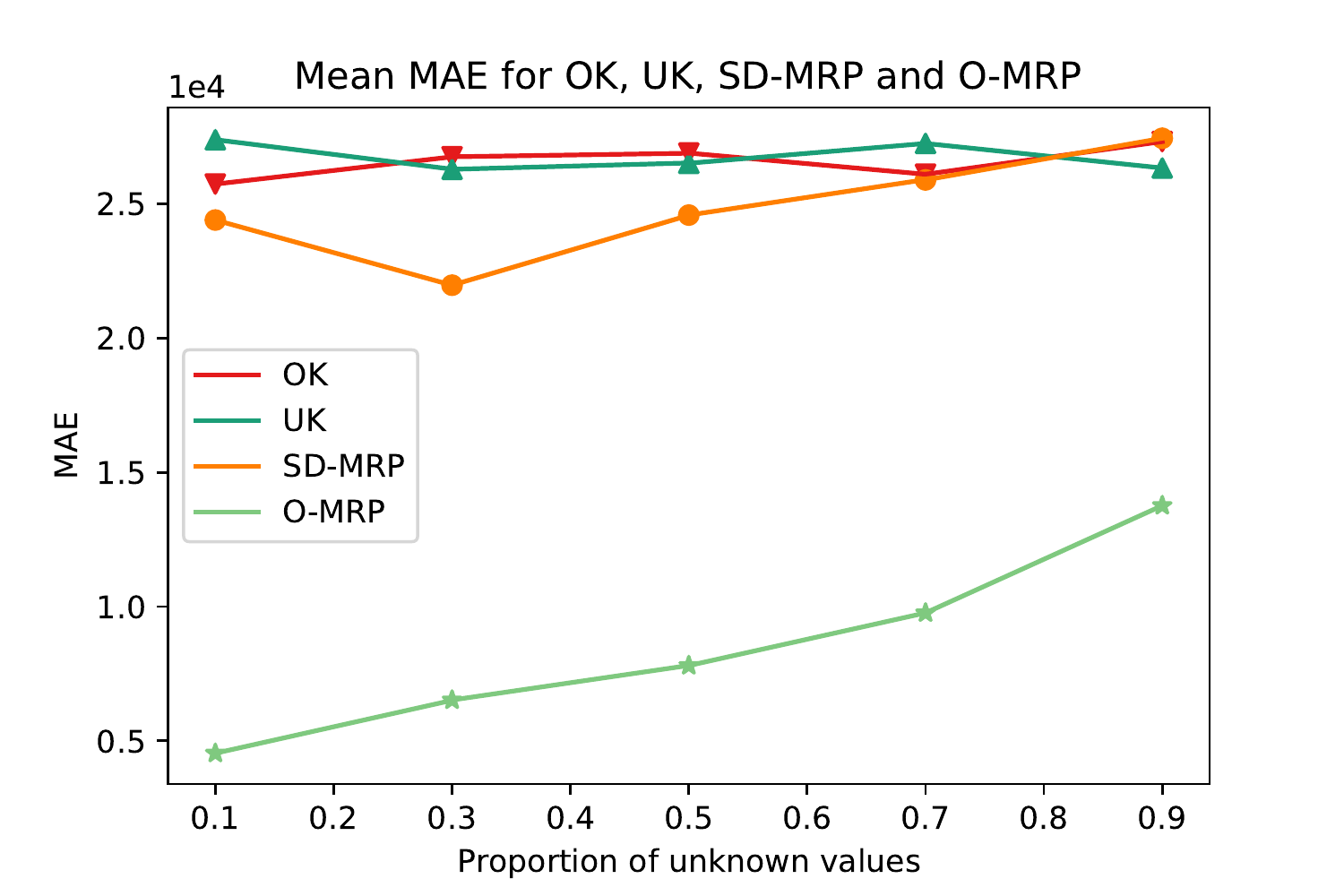}
  \caption{Mean MAE as a function of $p$ for GDP interpolation trained on Daegu and tested on Daegu: purely spatial methods.}
\end{figure}\par 

\begin{figure}[H]
  \centering
  \includegraphics[width=0.55\linewidth]{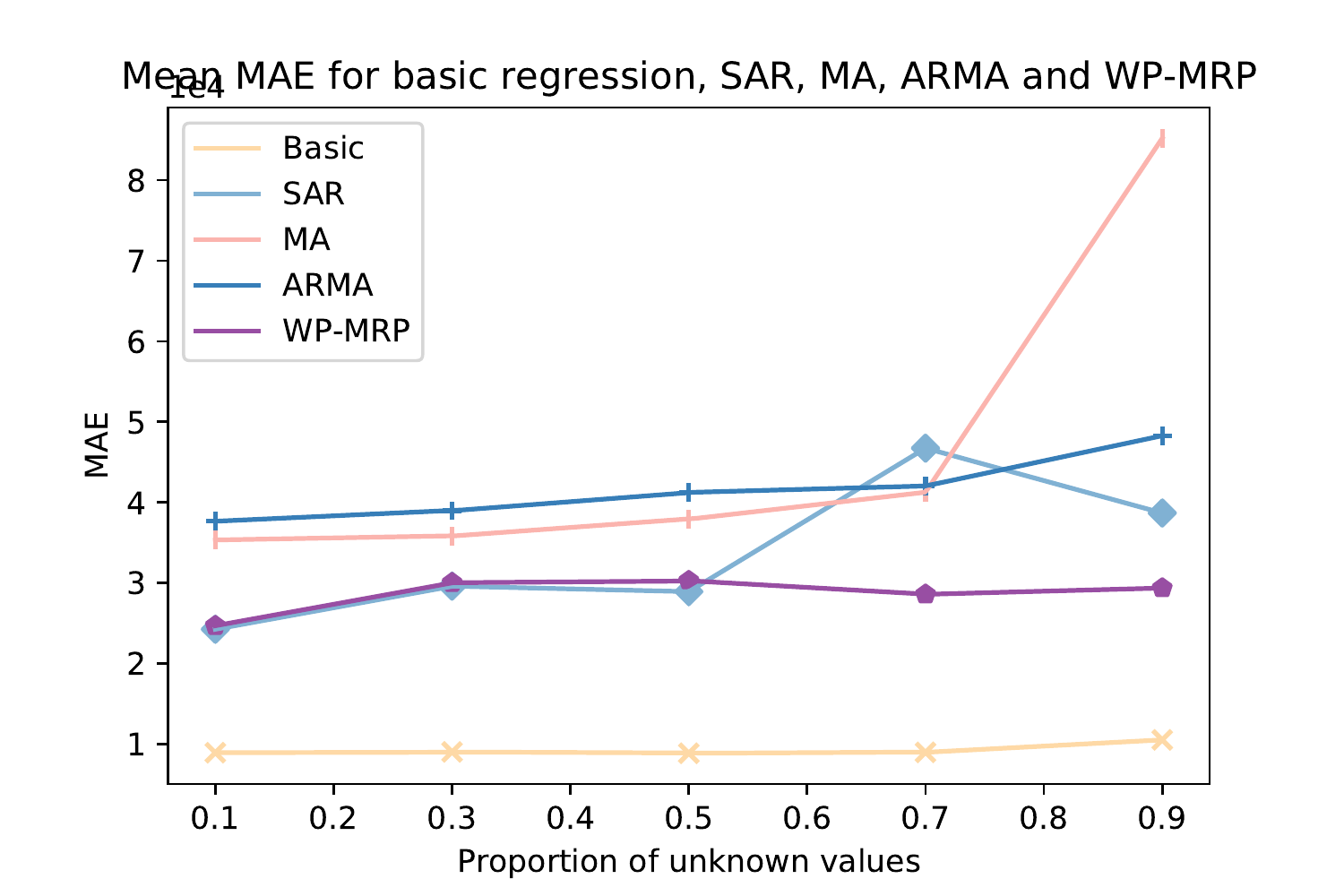}
  \caption{Mean MAE as a function of $p$ for GDP interpolation trained on Daegu and tested on Daegu: (spatial) regression methods.}
\end{figure}\par 

\begin{figure}[H]
  \centering
  \includegraphics[width=0.55\linewidth]{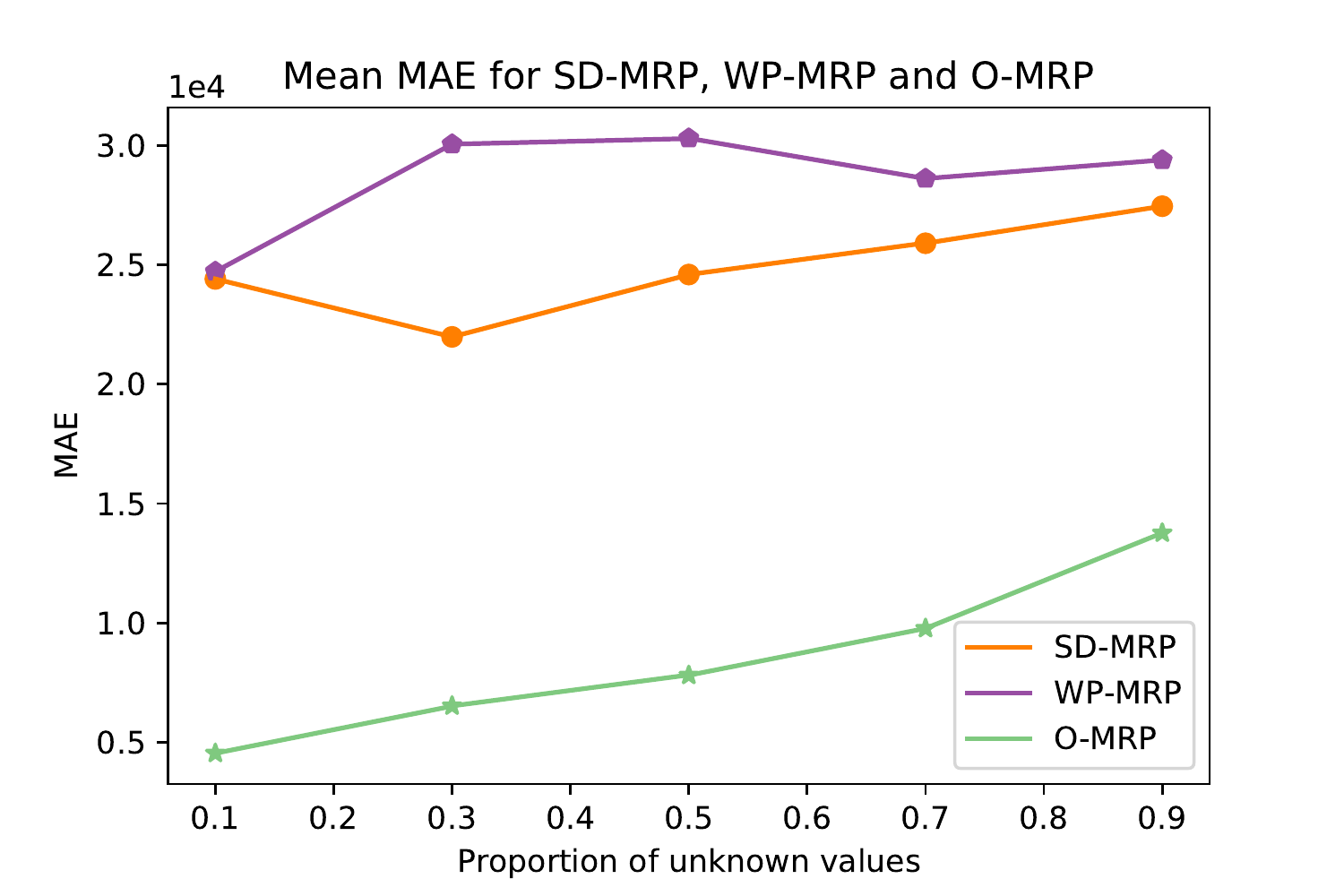}
  \caption{Mean MAE as a function of $p$ for GDP interpolation trained on Daegu and tested on Daegu: MRP methods.}
\end{figure}\par 

\begin{figure}[H]
  \centering
  \includegraphics[width=0.55\linewidth]{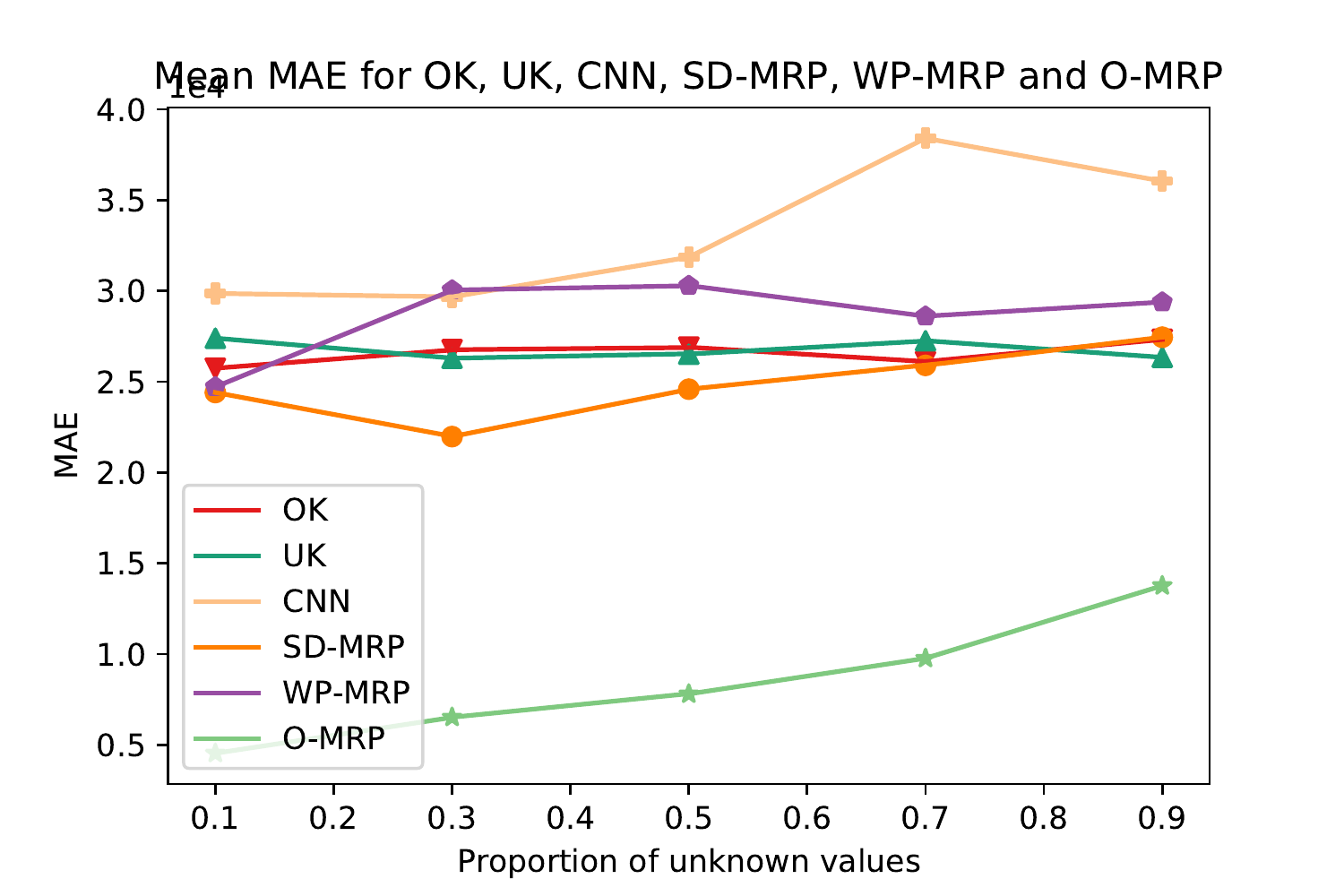}
  \caption{Mean MAE as a function of $p$ for GDP interpolation trained on Daegu and tested on Daegu: most competitive methods.}
\end{figure}\par 

\goodbreak
\textbf{Training region: Seoul}

\begin{figure}[H]
  \centering
  \includegraphics[width=0.55\linewidth]{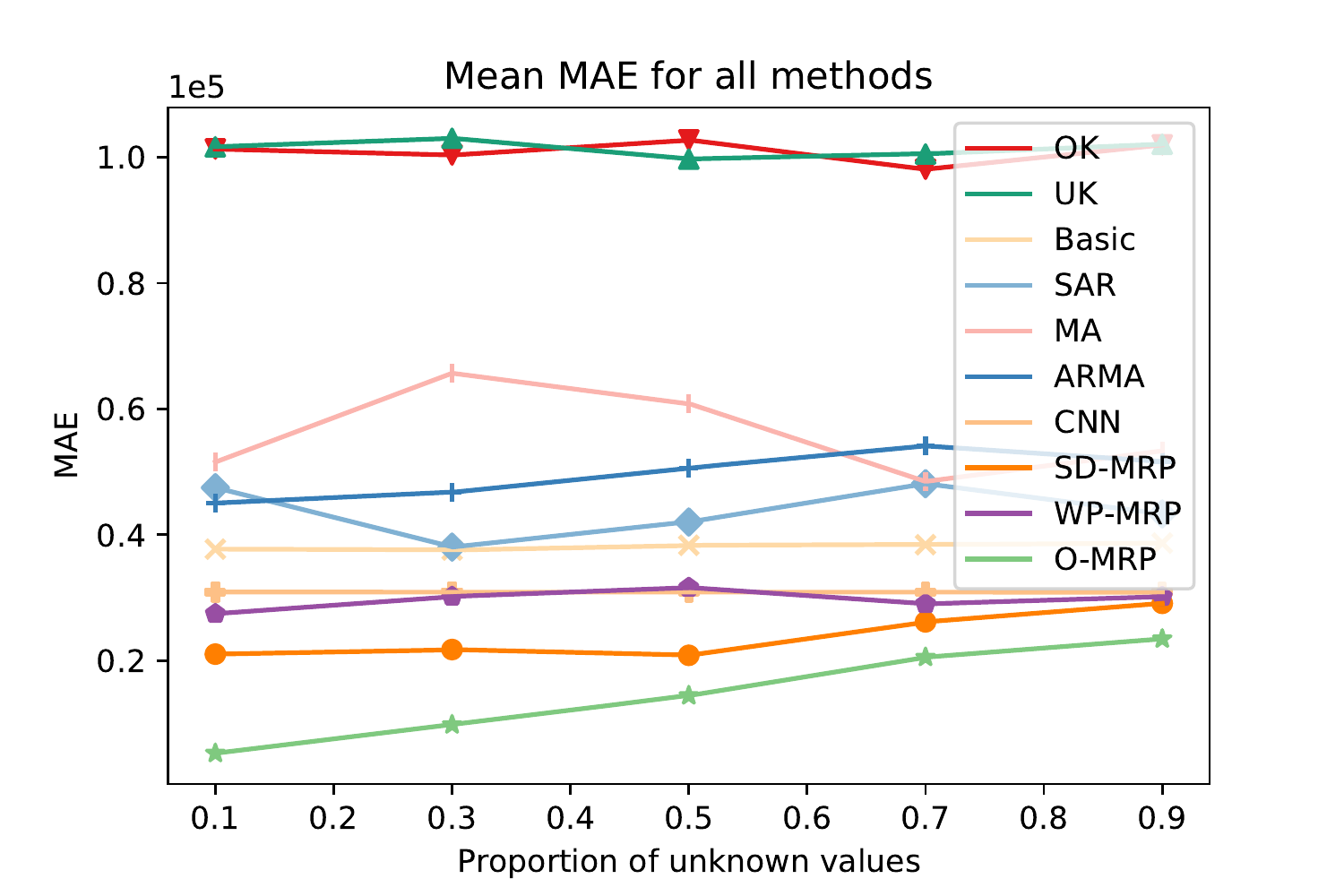}
  \caption{Mean MAE as a function of $p$ for GDP interpolation trained on Seoul and tested on Daegu: all methods.}
\end{figure}\par 

\begin{figure}[H]
  \centering
  \includegraphics[width=0.55\linewidth]{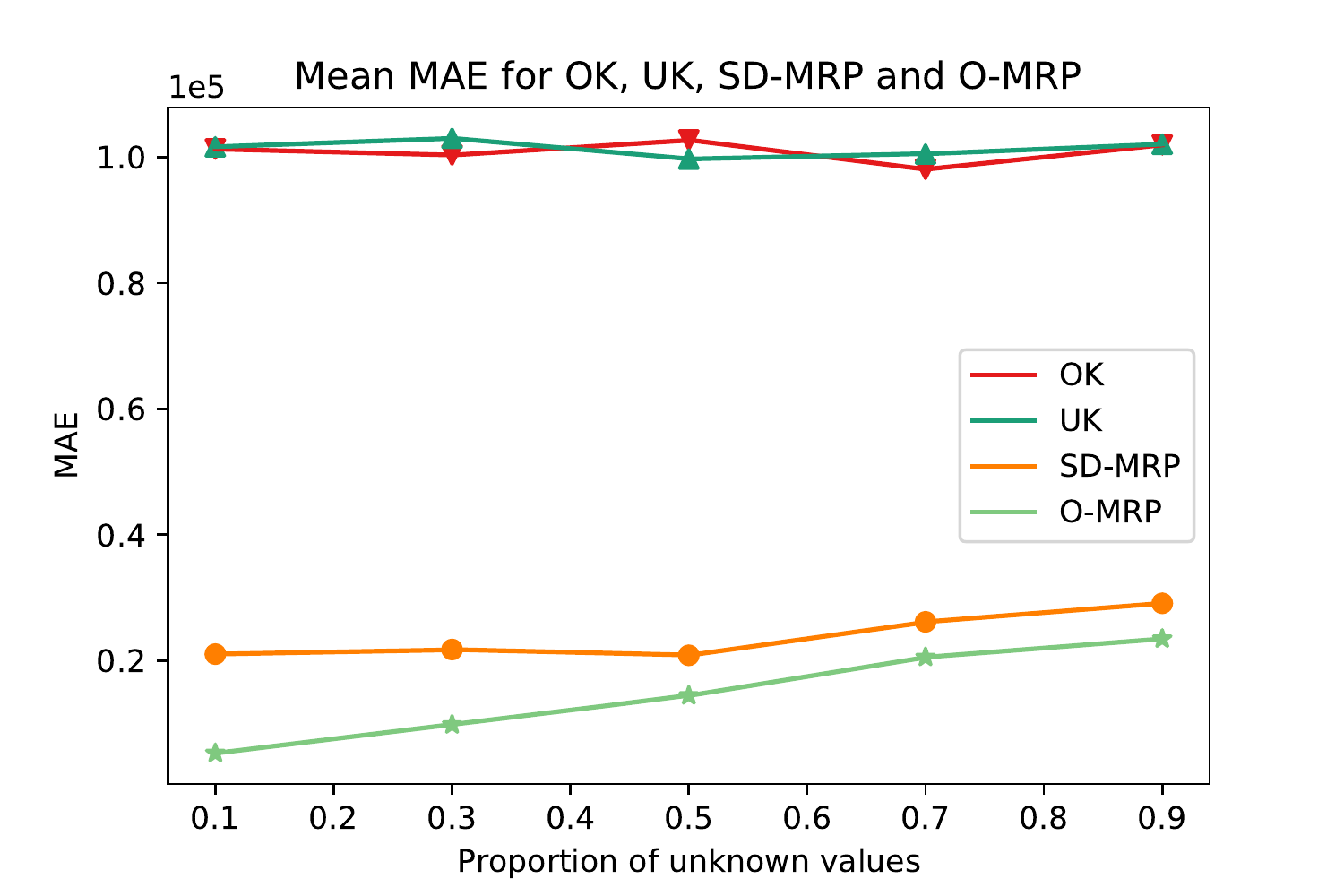}
  \caption{Mean MAE as a function of $p$ for GDP interpolation trained on Seoul and tested on Daegu: purely spatial methods.}
\end{figure}\par 

\begin{figure}[H]
  \centering
  \includegraphics[width=0.55\linewidth]{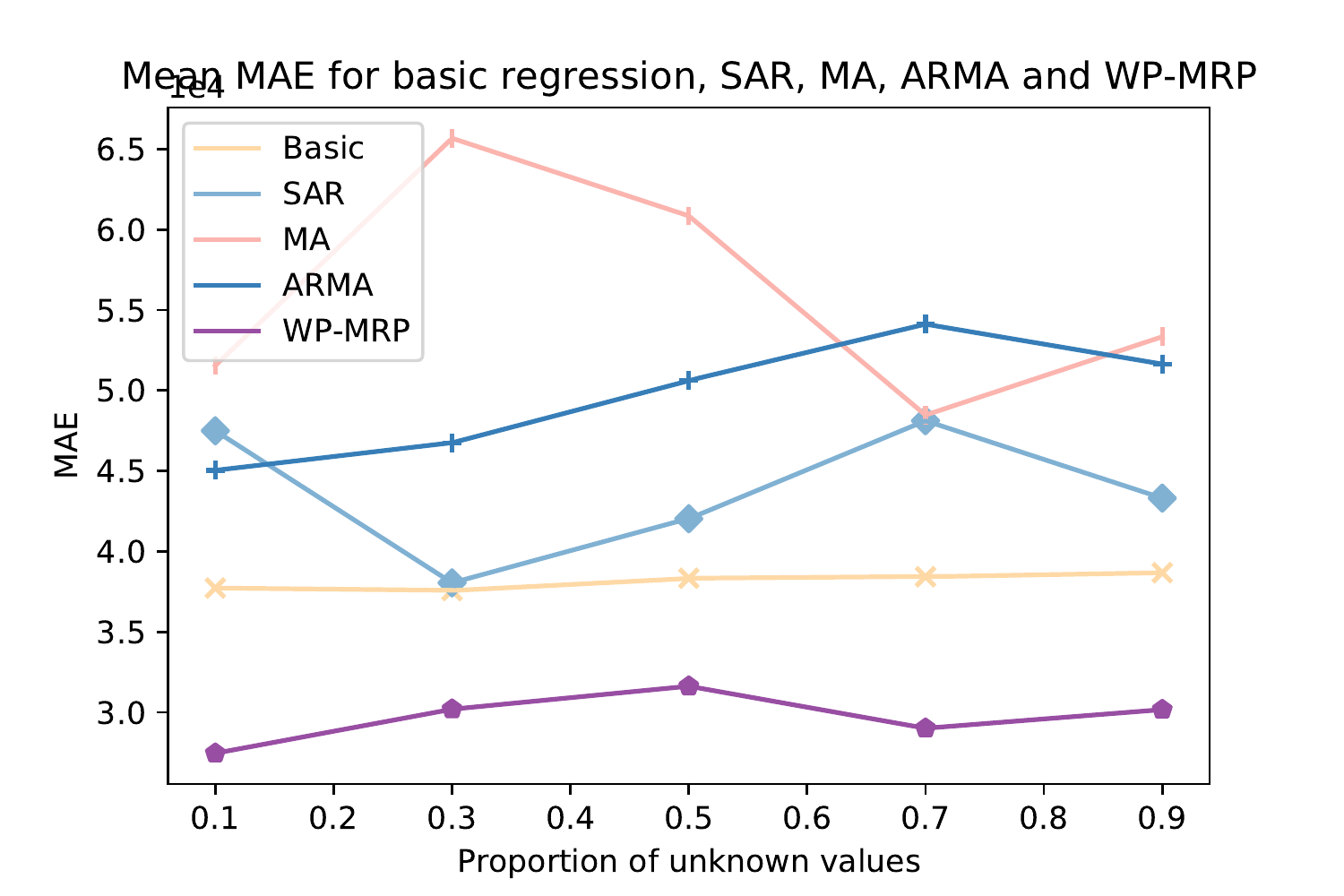}
  \caption{Mean MAE as a function of $p$ for GDP interpolation trained on Seoul and tested on Daegu: (spatial) regression methods.}
\end{figure}\par 

\begin{figure}[H]
  \centering
  \includegraphics[width=0.55\linewidth]{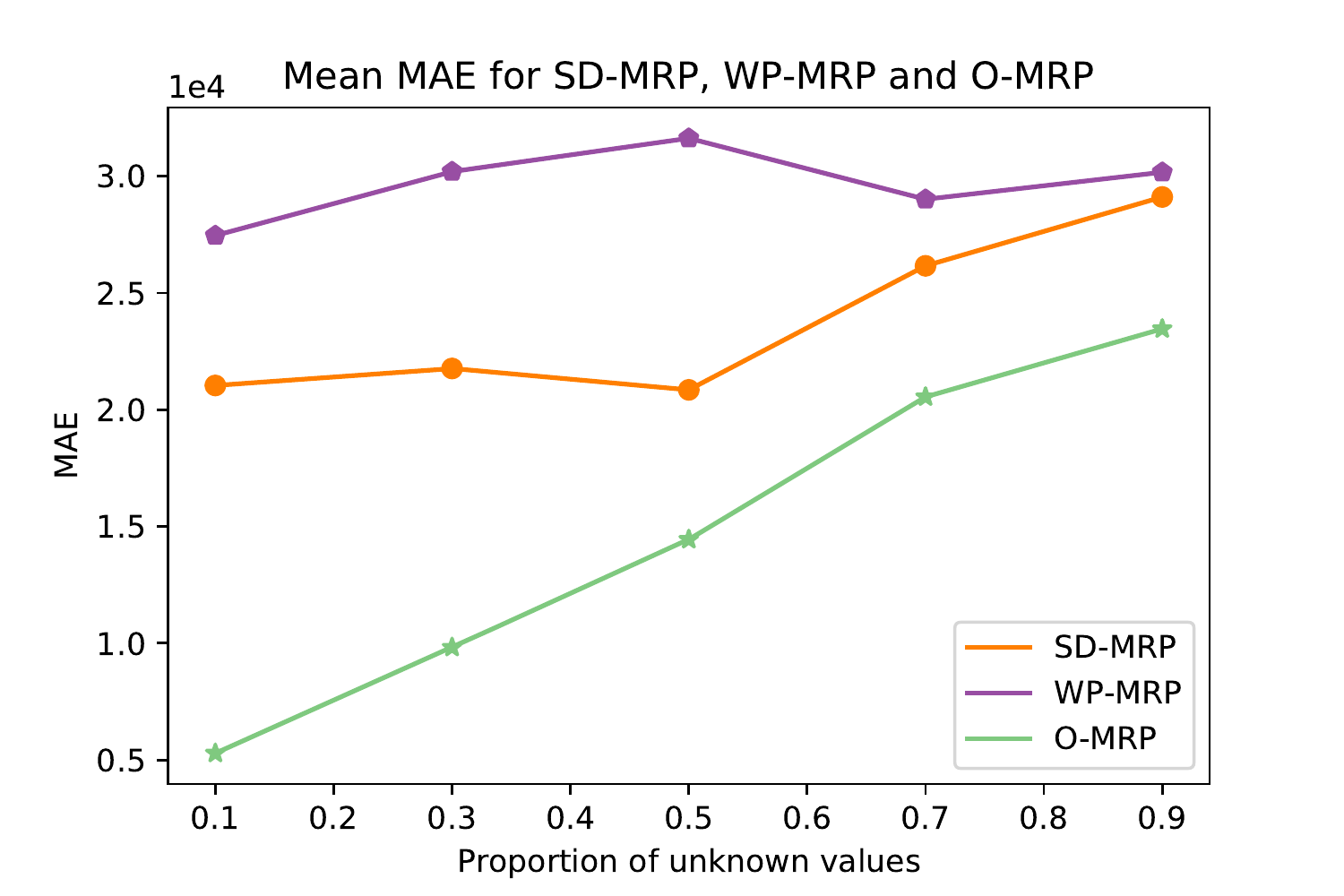}
  \caption{Mean MAE as a function of $p$ for GDP interpolation trained on Seoul and tested on Daegu: MRP methods.}
\end{figure}\par 

\begin{figure}[H]
  \centering
  \includegraphics[width=0.55\linewidth]{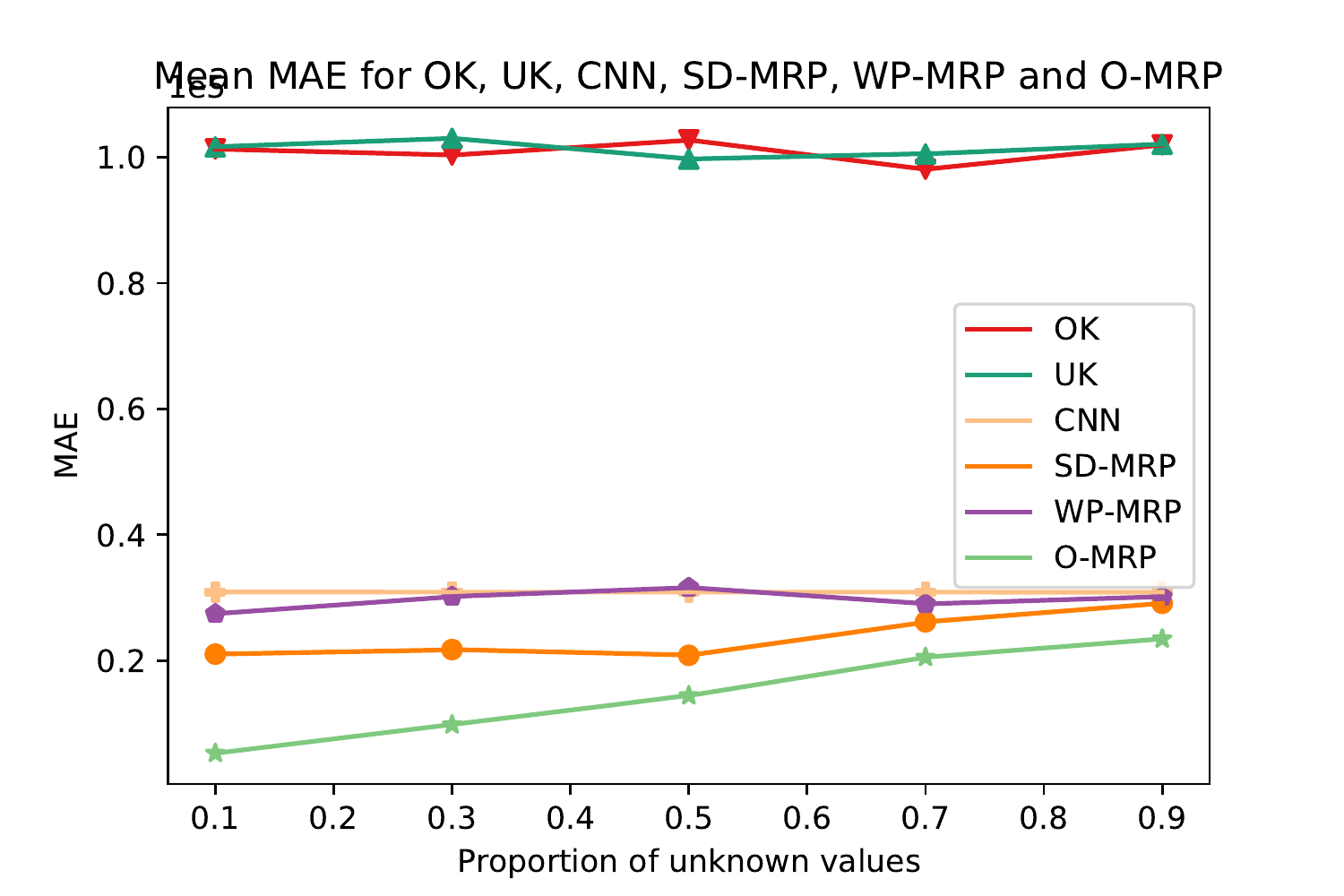}
  \caption{Mean MAE as a function of $p$ for GDP interpolation trained on Seoul and tested on Daegu: most competitive methods.}
\end{figure}\par 

\textbf{Training region: Taichung}

\begin{figure}[H]
  \centering
  \includegraphics[width=0.55\linewidth]{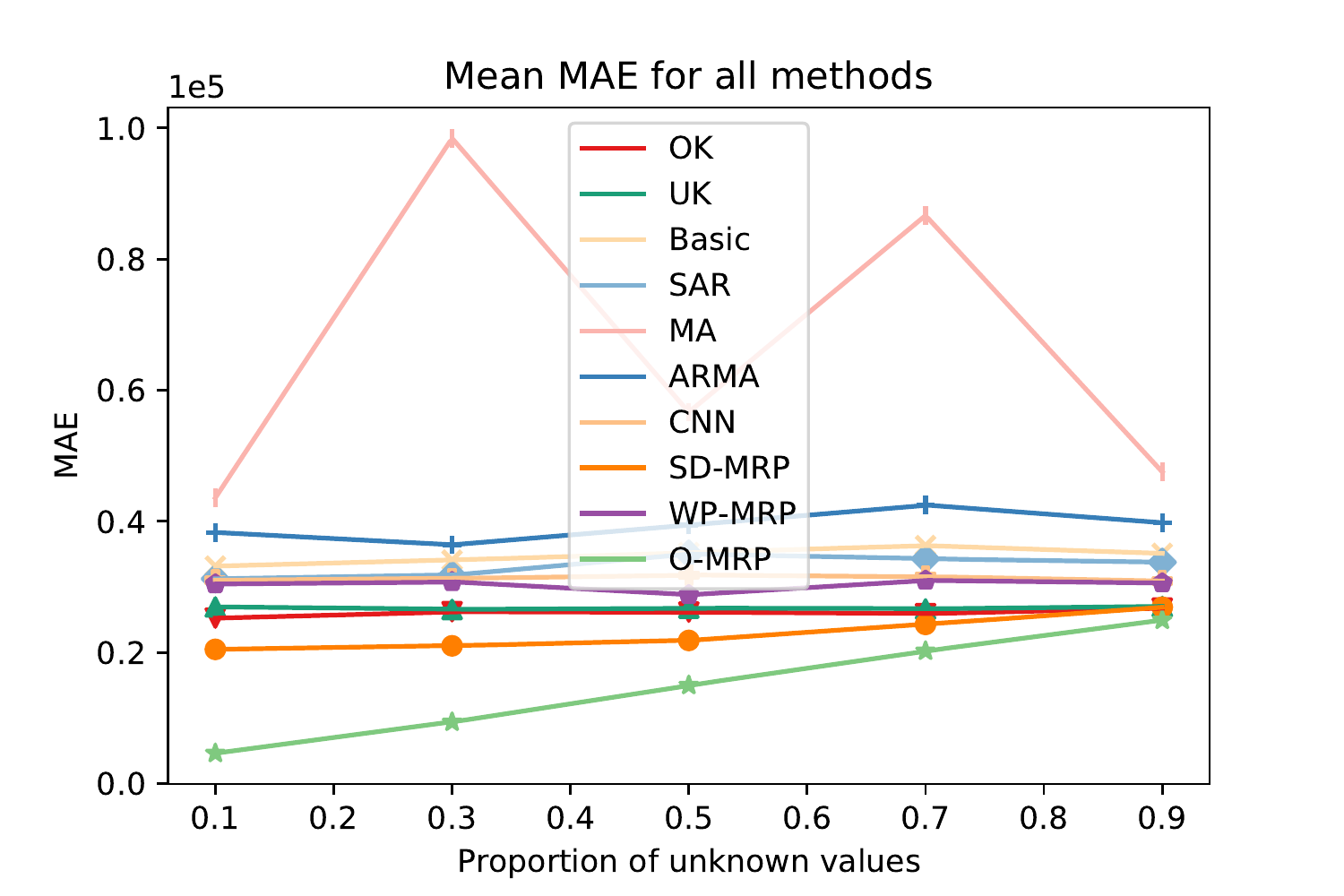}
  \caption{Mean MAE as a function of $p$ for GDP interpolation trained on Taichung and tested on Daegu: all methods.}
\end{figure}\par 

\begin{figure}[H]
  \centering
  \includegraphics[width=0.55\linewidth]{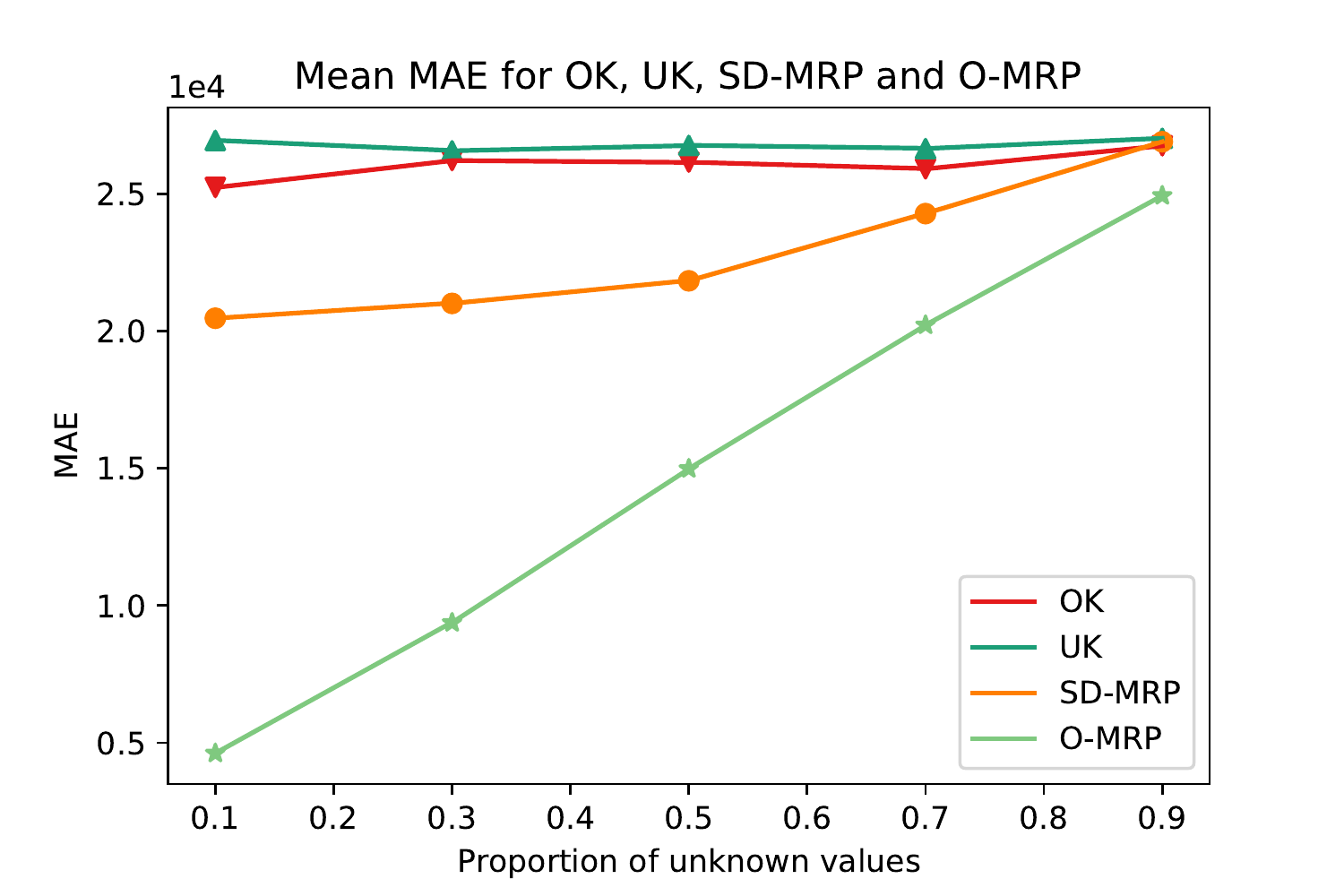}
  \caption{Mean MAE as a function of $p$ for GDP interpolation trained on Taichung and tested on Daegu: purely spatial methods.}
\end{figure}\par 

\begin{figure}[H]
  \centering
  \includegraphics[width=0.55\linewidth]{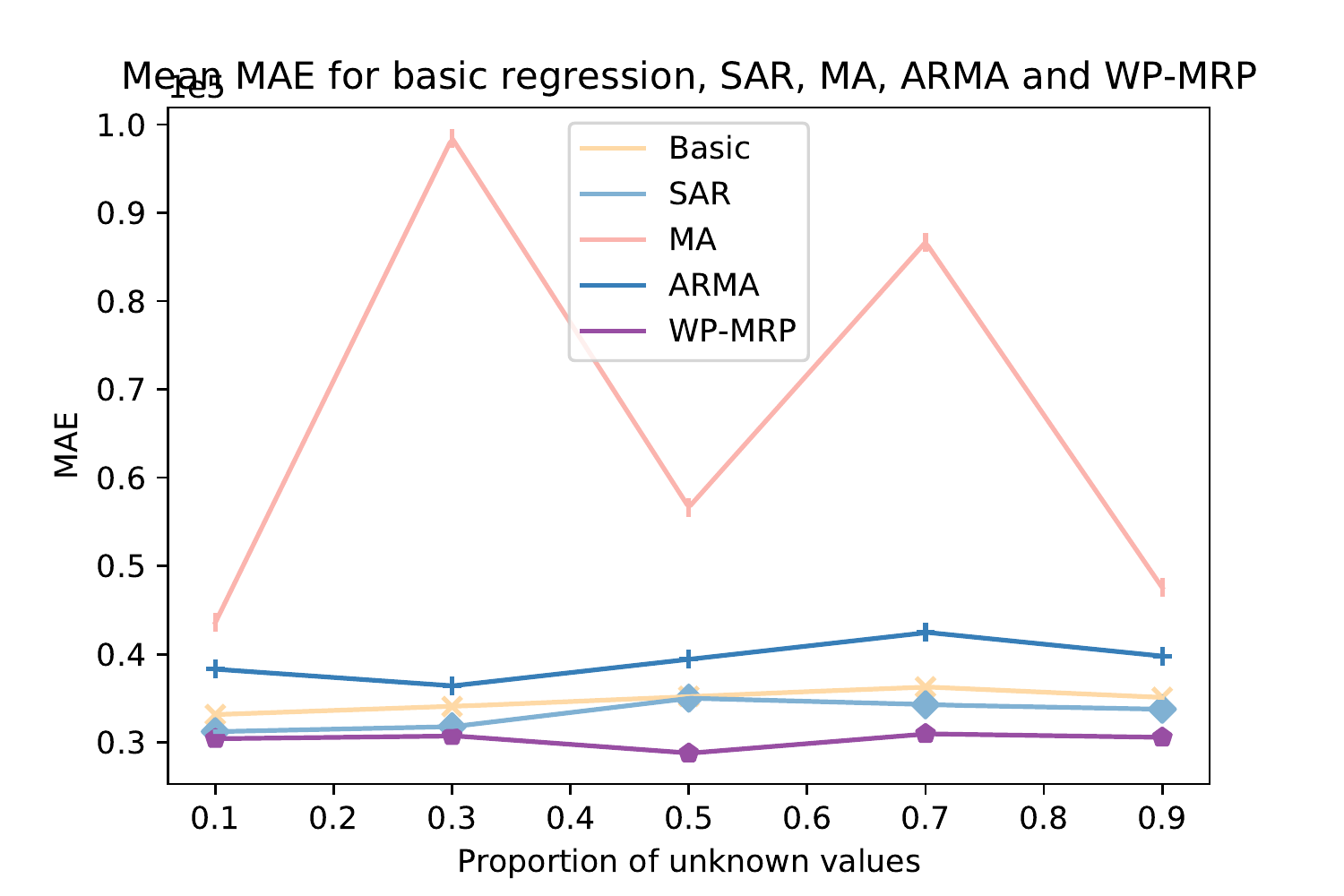}
  \caption{Mean MAE as a function of $p$ for GDP interpolation trained on Taichung and tested on Daegu: (spatial) regression methods.}
\end{figure}\par 

\begin{figure}[H]
  \centering
  \includegraphics[width=0.55\linewidth]{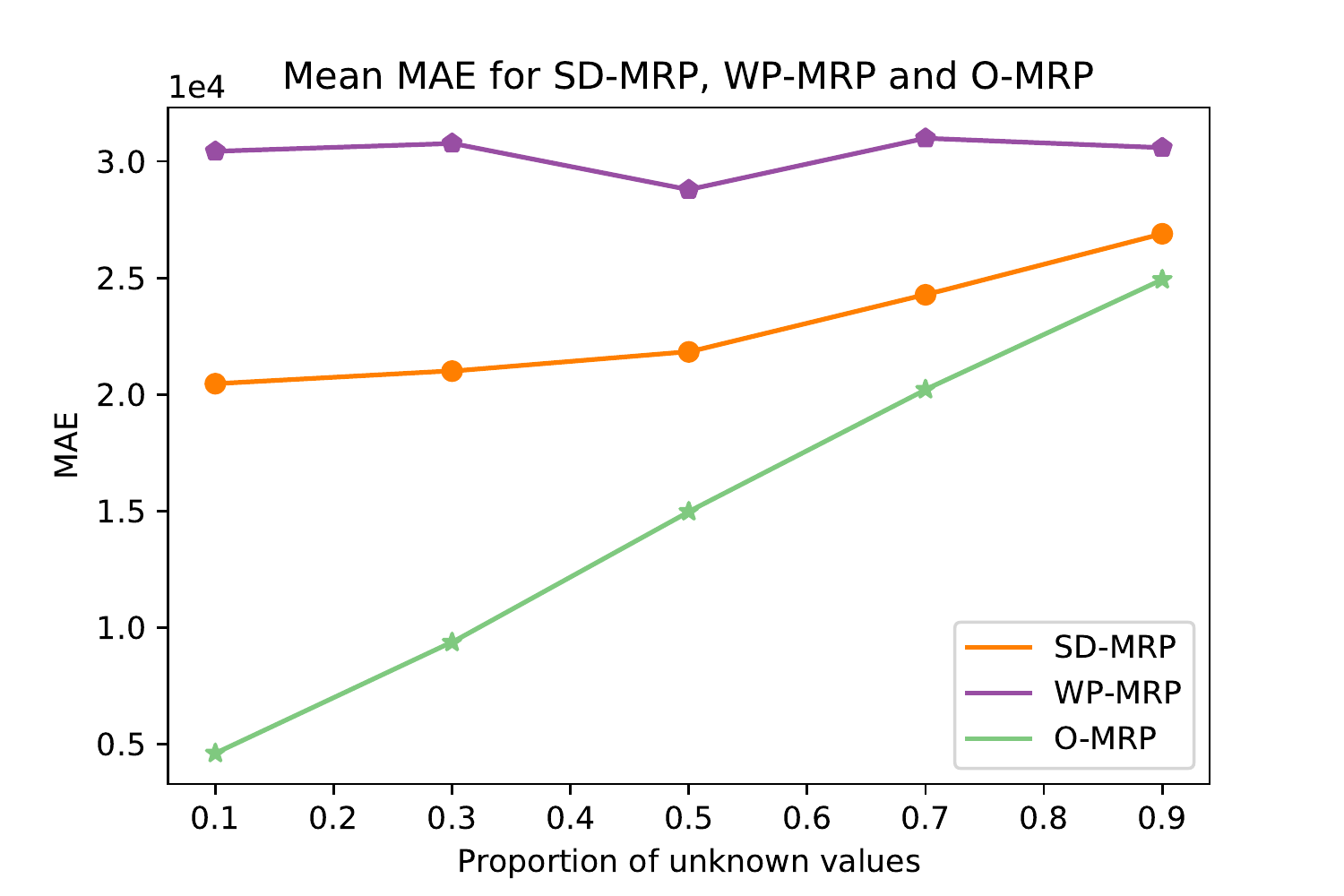}
  \caption{Mean MAE as a function of $p$ for GDP interpolation trained on Taichung and tested on Daegu: MRP methods.}
\end{figure}\par 

\begin{figure}[H]
  \centering
  \includegraphics[width=0.55\linewidth]{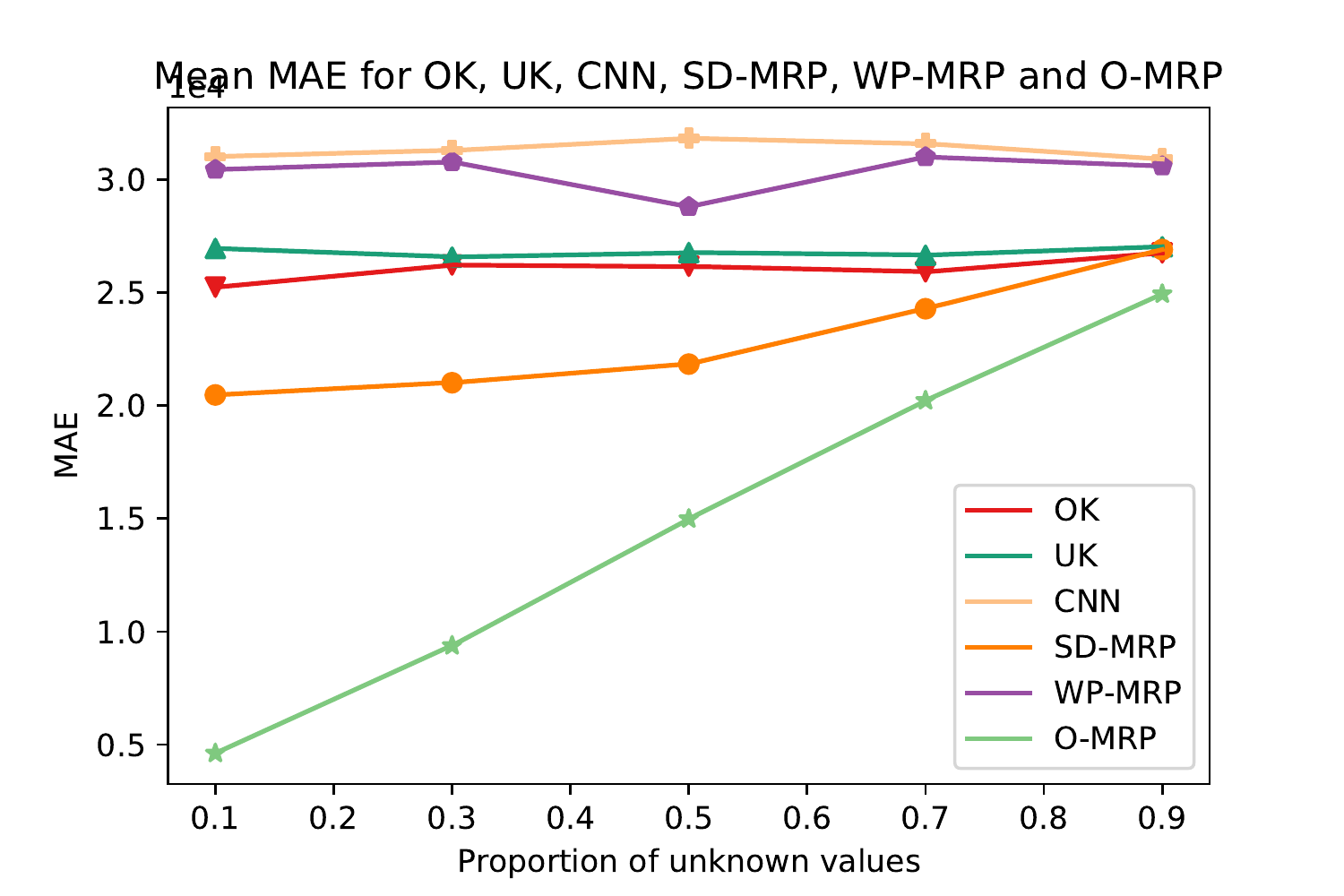}
  \caption{Mean MAE as a function of $p$ for GDP interpolation trained on Taichung and tested on Daegu: most competitive methods.}
\end{figure}\par

\subsection{COVID-19}

The results for the COVID-19 dataset were by far the most favourable to MRPs. In nearly all cases the MRP methods all performed better than the other baselines. Interestingly, though, basic regression also performed well when trained on Daegu. When trained on Seoul, however, this was no longer the case. As in other cases, we see some erratic performance from MA such as in Figure 33, and strangely we also see the errors of SD-MRP and WP-MRP decrease with $p$ in Figure 34. We are not sure what caused this, but it is certainly unusual.

\subsubsection{Test region: Daegu}

\textbf{Training region: Daegu}

\begin{figure}[H]
  \centering
  \includegraphics[width=0.55\linewidth]{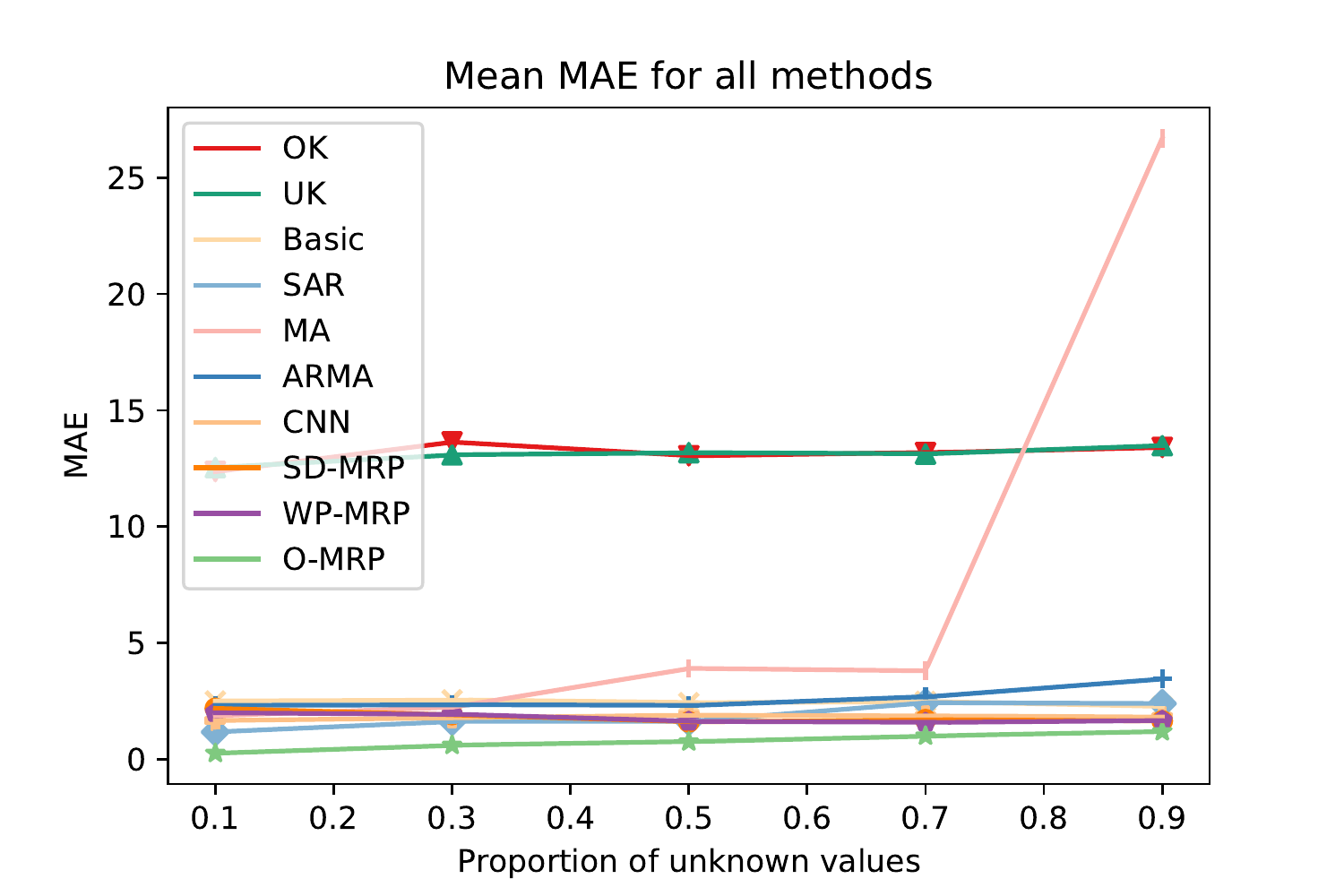}
  \caption{Mean MAE as a function of $p$ for COVID-19 interpolation trained on Daegu and tested on Daegu: all methods.}
\end{figure}\par 

\begin{figure}[H]
  \centering
  \includegraphics[width=0.55\linewidth]{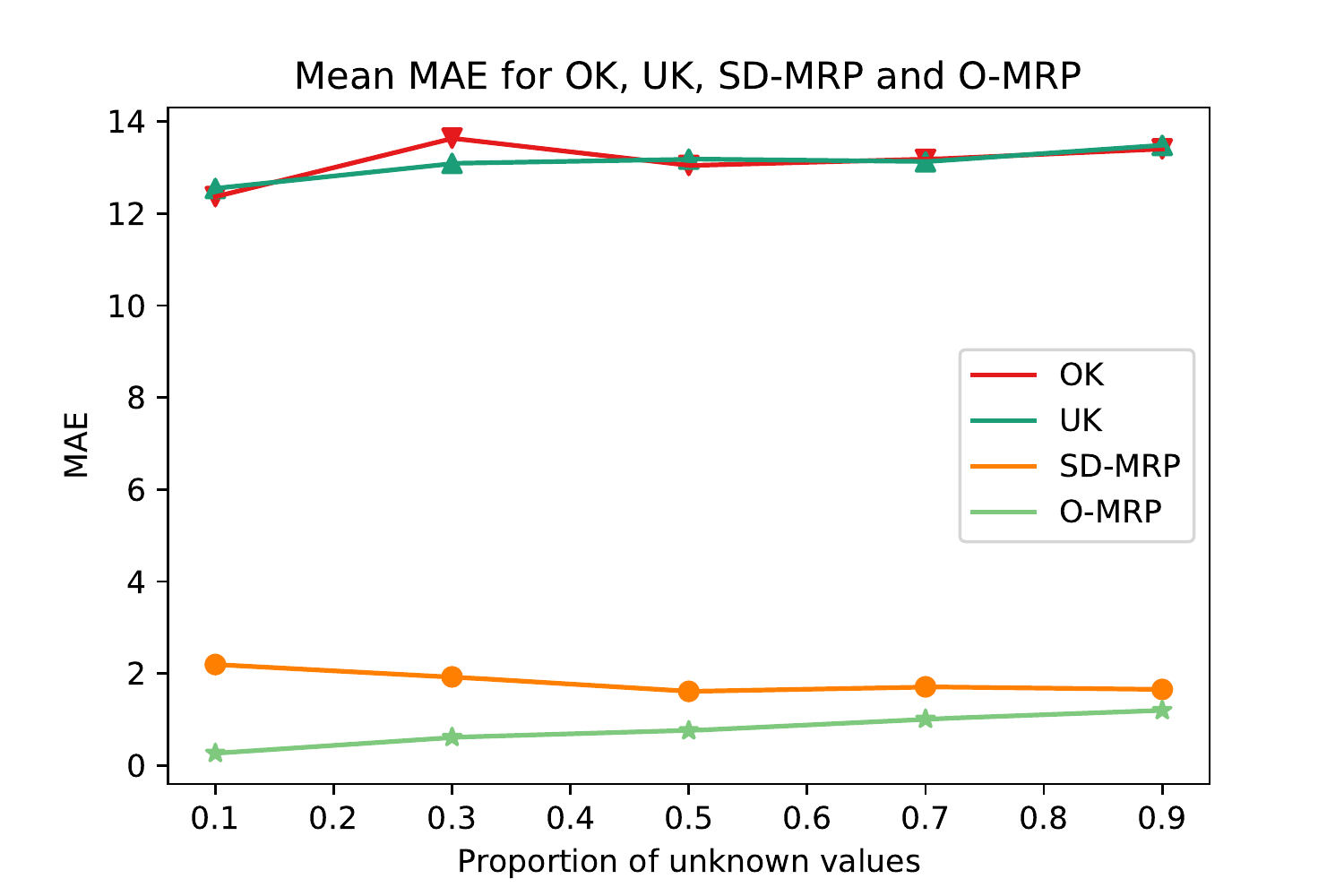}
  \caption{Mean MAE as a function of $p$ for COVID-19 interpolation trained on Daegu and tested on Daegu: purely spatial methods.}
\end{figure}\par 

\begin{figure}[H]
  \centering
  \includegraphics[width=0.55\linewidth]{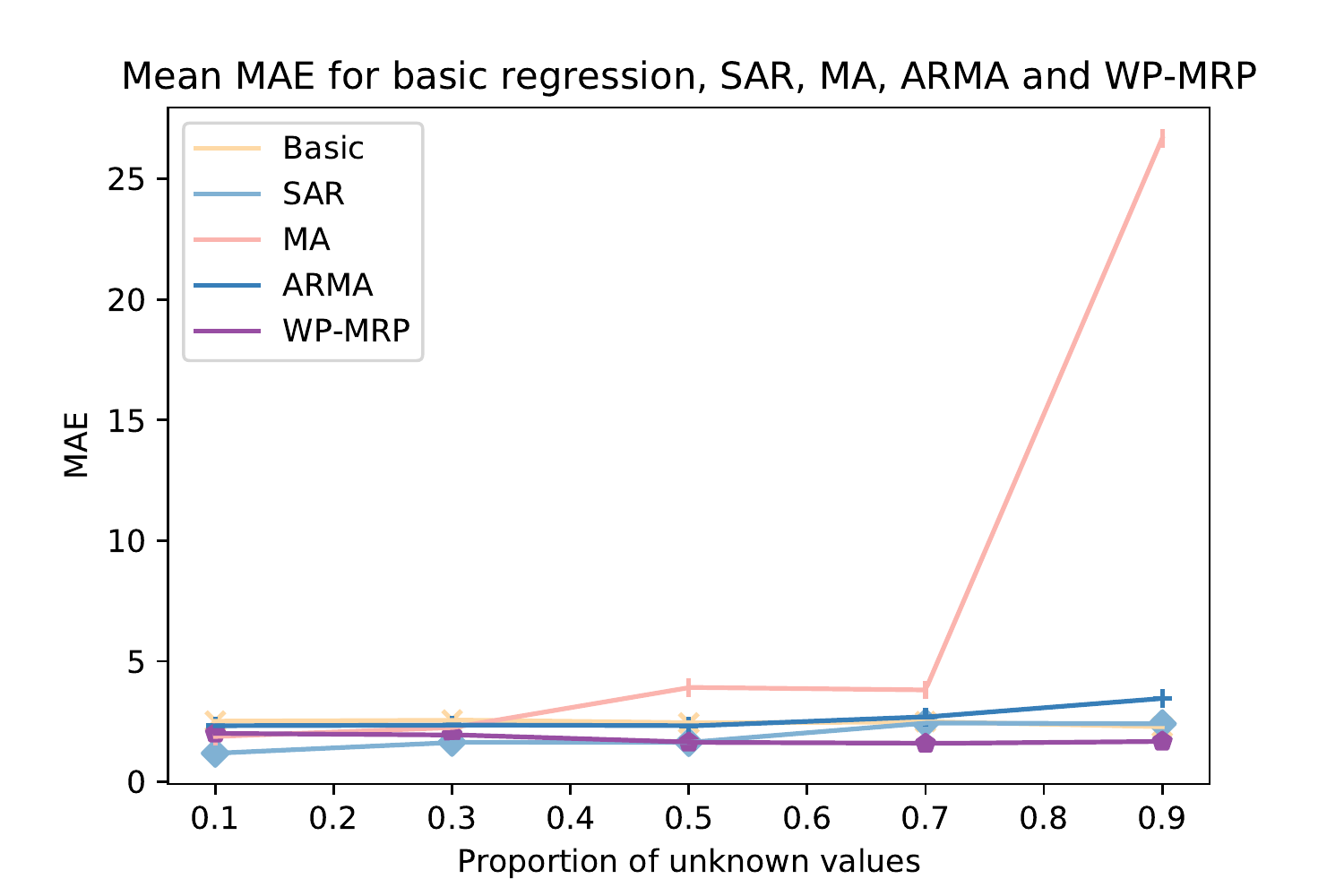}
  \caption{Mean MAE as a function of $p$ for COVID-19 interpolation trained on Daegu and tested on Daegu: (spatial) regression methods.}
\end{figure}\par 

\begin{figure}[H]
  \centering
  \includegraphics[width=0.55\linewidth]{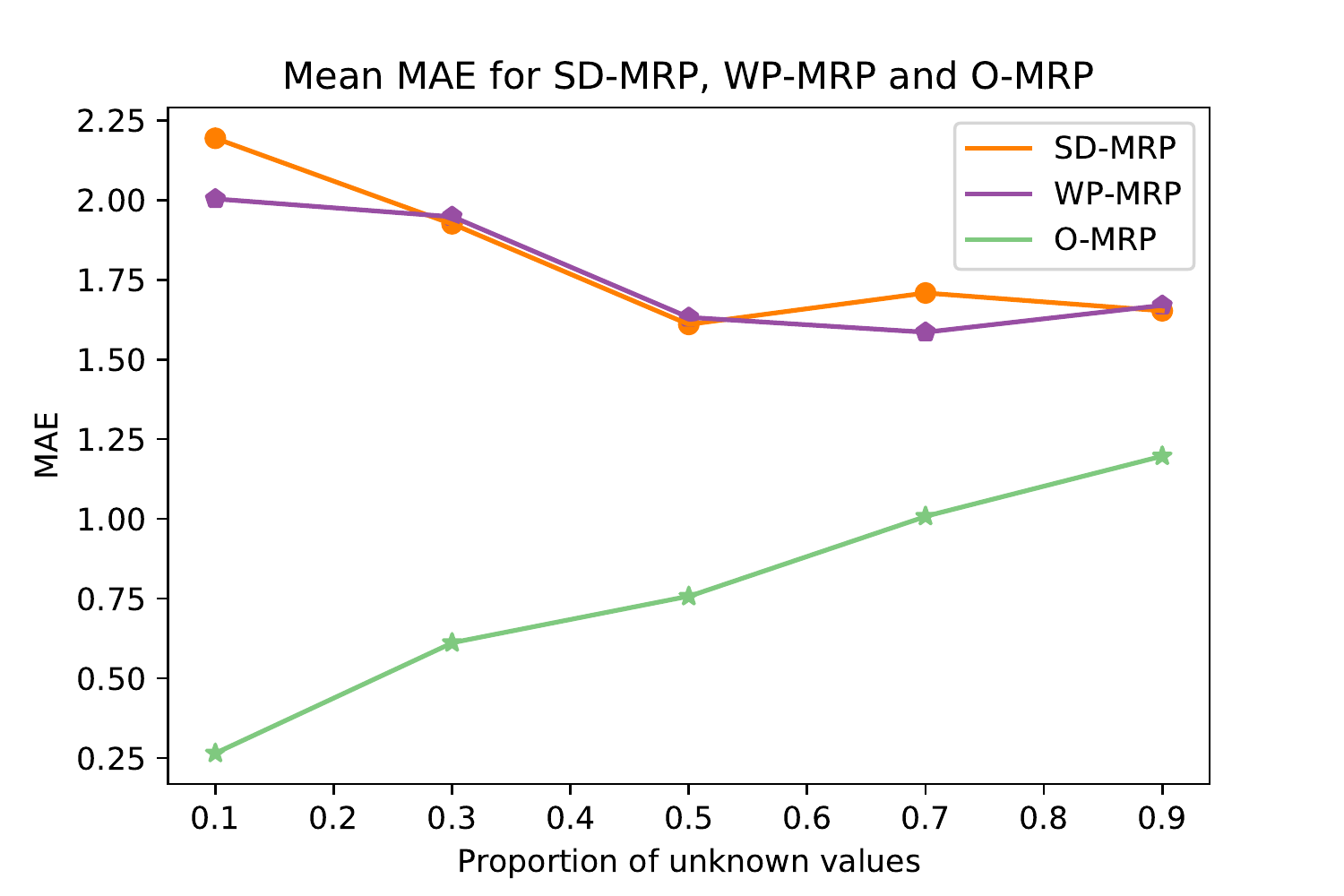}
  \caption{Mean MAE as a function of $p$ for COVID-19 interpolation trained on Daegu and tested on Daegu: MRP methods.}
\end{figure}\par 

\begin{figure}[H]
  \centering
  \includegraphics[width=0.55\linewidth]{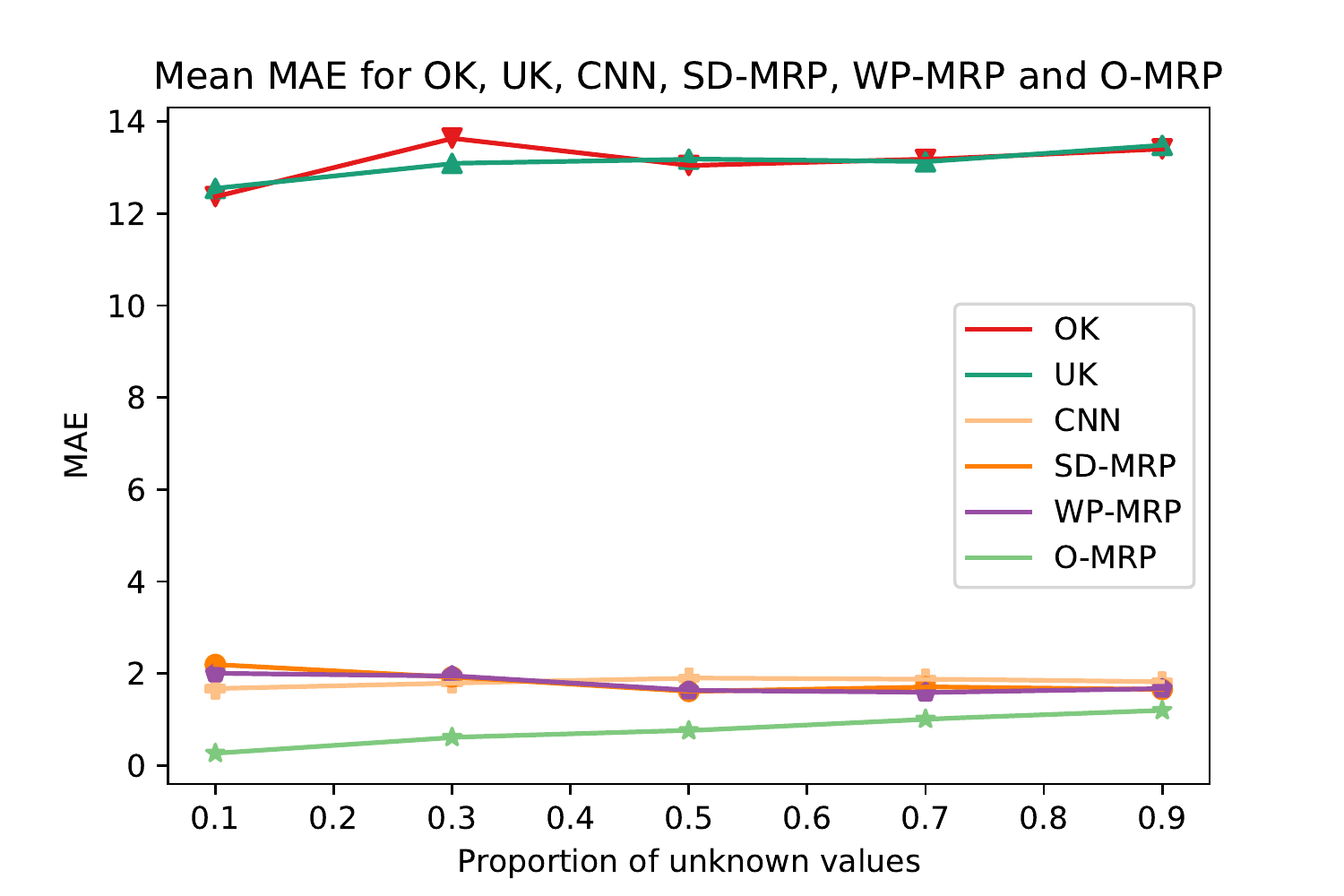}
  \caption{Mean MAE as a function of $p$ for COVID-19 interpolation trained on Daegu and tested on Daegu: most competitive methods.}
\end{figure}\par

\textbf{Training region: Seoul}

\begin{figure}[H]
  \centering
  \includegraphics[width=0.55\linewidth]{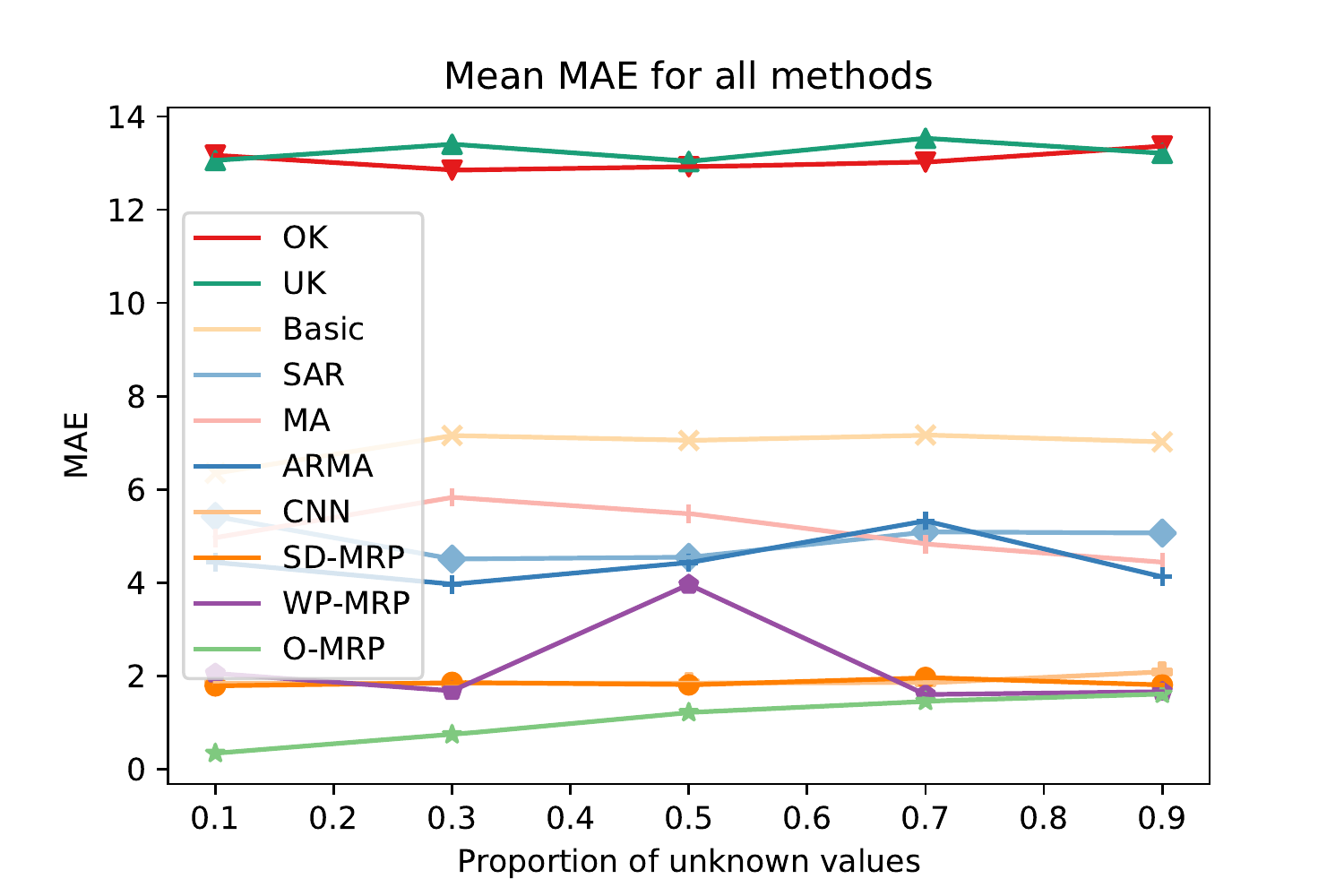}
  \caption{Mean MAE as a function of $p$ for COVID-19 interpolation trained on Seoul and tested on Daegu: all methods.}
\end{figure}\par 

\begin{figure}[H]
  \centering
  \includegraphics[width=0.55\linewidth]{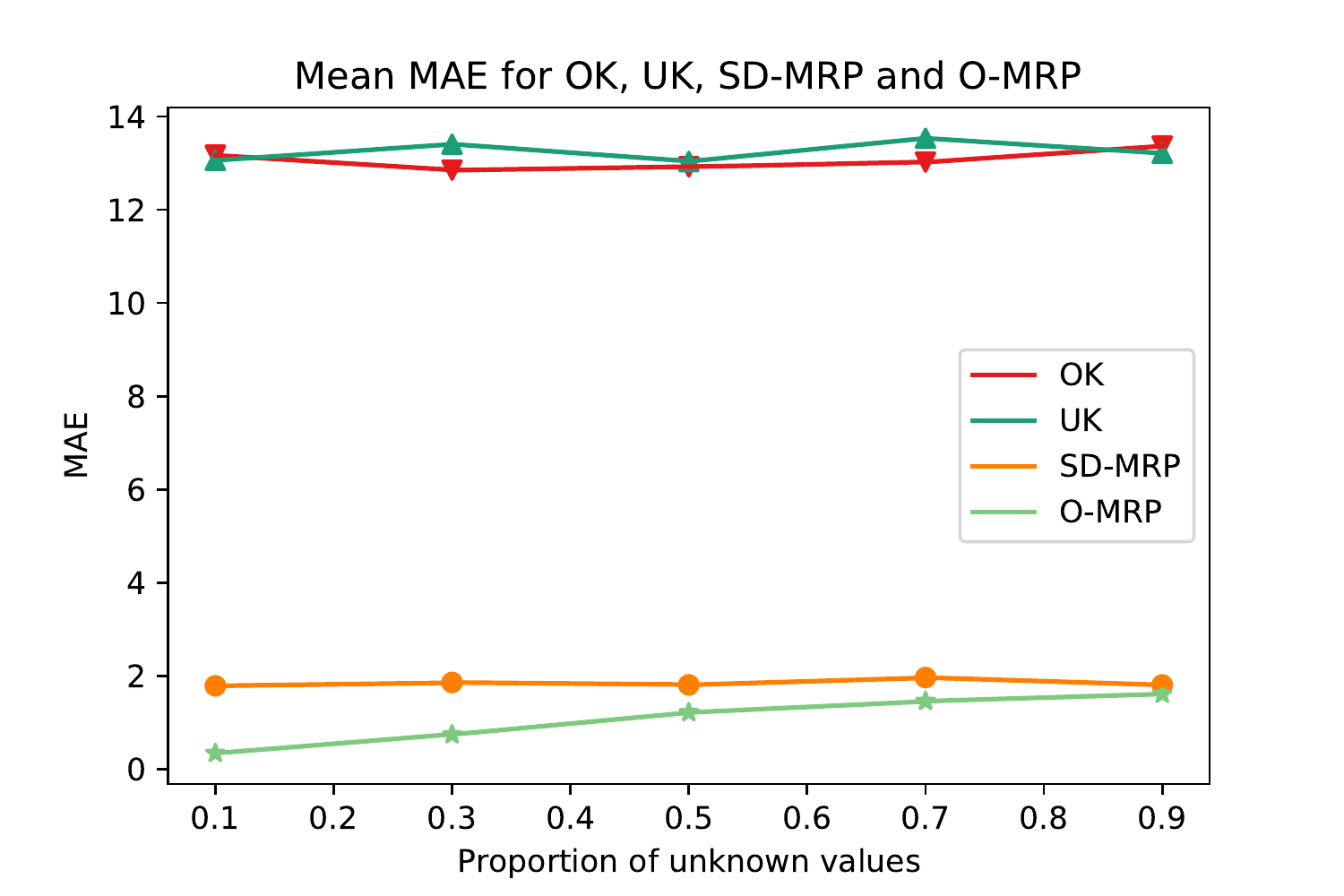}
  \caption{Mean MAE as a function of $p$ for COVID-19 interpolation trained on Seoul and tested on Daegu: purely spatial methods.}
\end{figure}\par 

\begin{figure}[H]
  \centering
  \includegraphics[width=0.55\linewidth]{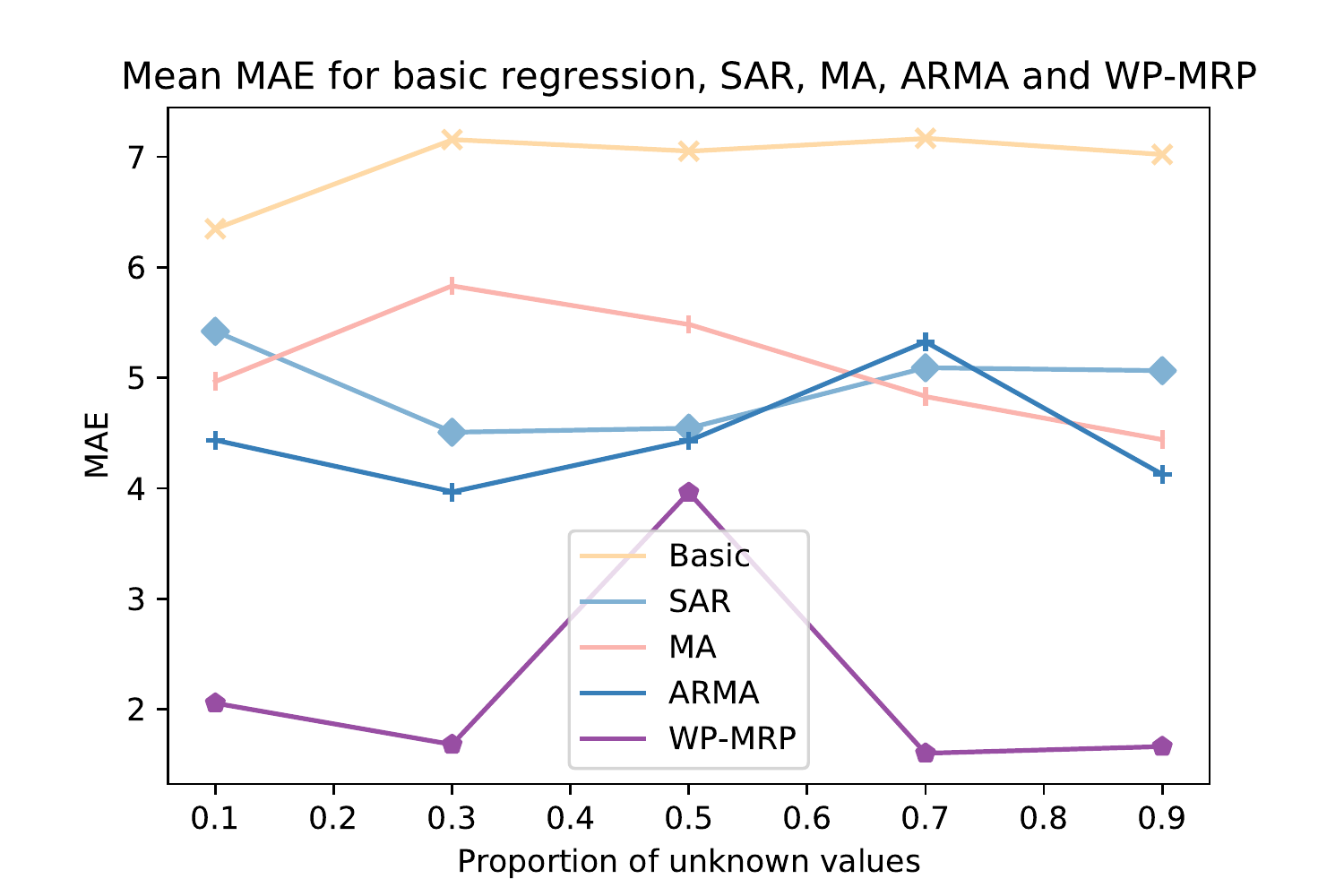}
  \caption{Mean MAE as a function of $p$ for COVID-19 interpolation trained on Seoul and tested on Daegu: (spatial) regression methods.}
\end{figure}\par 

\begin{figure}[H]
  \centering
  \includegraphics[width=0.55\linewidth]{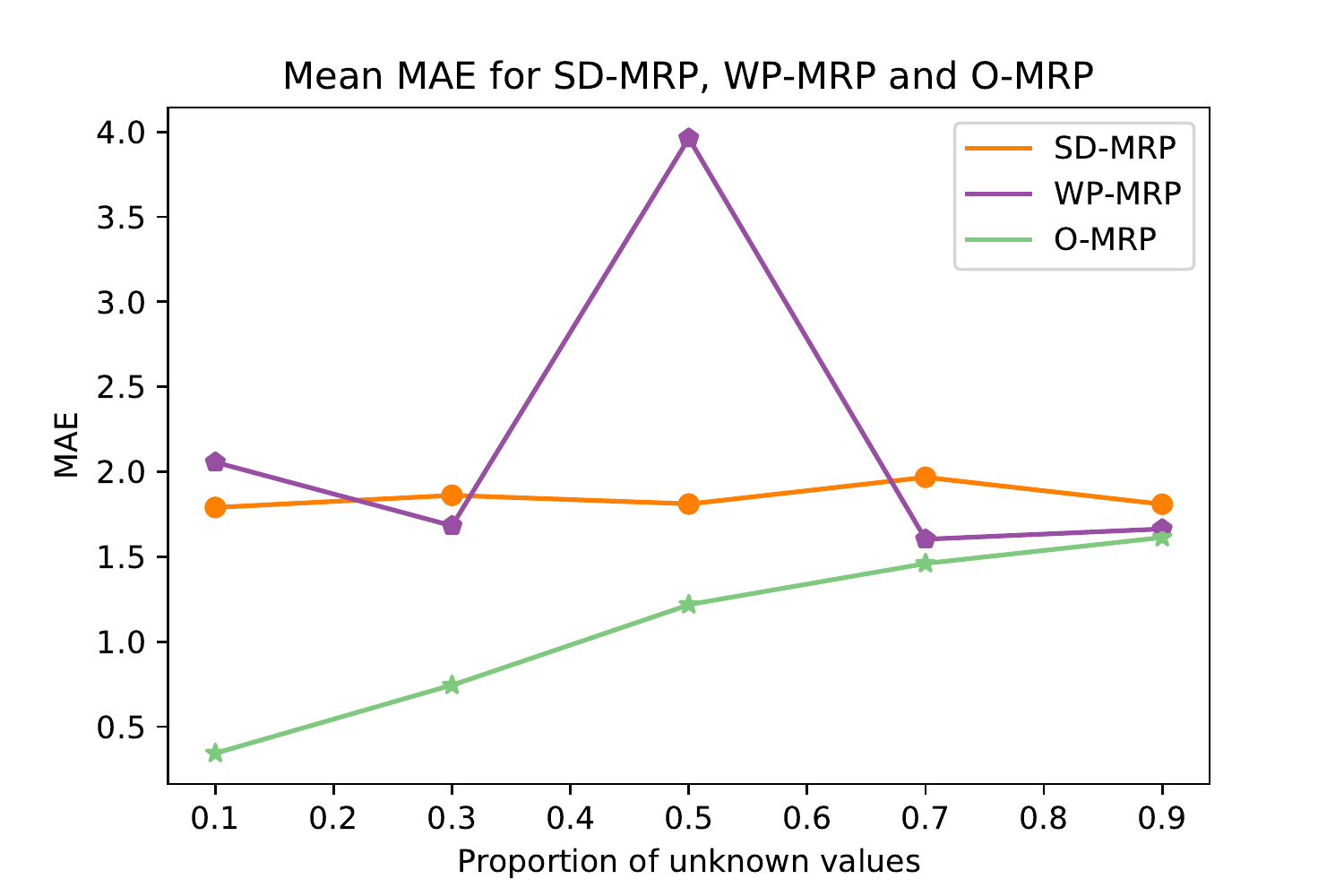}
  \caption{Mean MAE as a function of $p$ for COVID-19 interpolation trained on Seoul and tested on Daegu: MRP methods.}
\end{figure}\par 

\begin{figure}[H]
  \centering
  \includegraphics[width=0.55\linewidth]{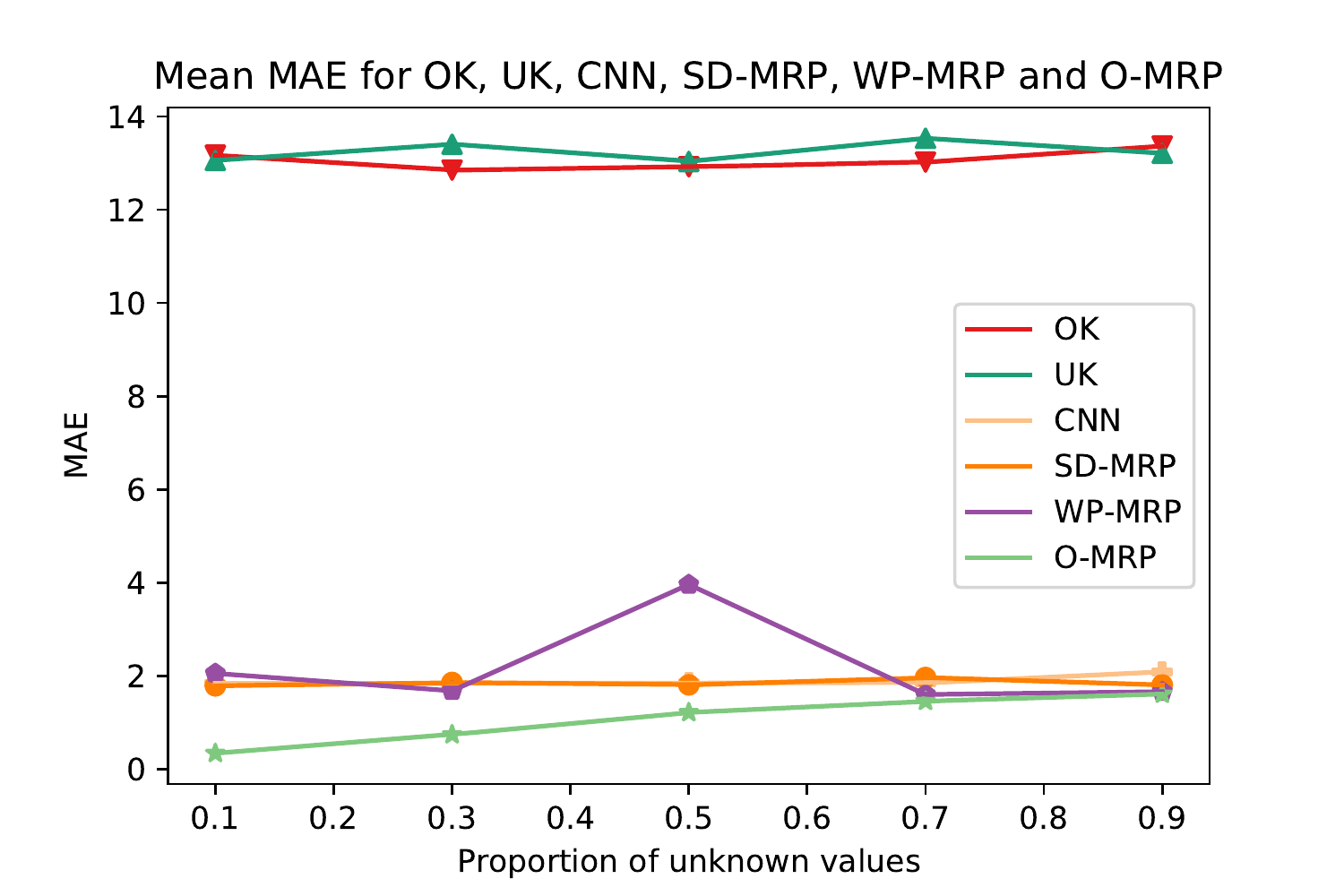}
  \caption{Mean MAE as a function of $p$ for COVID-19 interpolation trained on Seoul and tested on Daegu: most competitive methods.}
\end{figure}\par

\end{document}